\documentclass{article}

\usepackage{microtype}
\usepackage{graphicx}
\usepackage{booktabs} 

\usepackage[utf8]{inputenc}
\usepackage[T1]{fontenc}
\usepackage{booktabs}
\usepackage{multirow}
\usepackage{nccmath}

\usepackage{amsmath}
\usepackage{amsthm}
\usepackage{amssymb}
\usepackage{empheq}
\usepackage{xcolor, color, colortbl}
\usepackage{mdframed}
\usepackage{pifont}
\usepackage{enumitem}
\usepackage{graphicx} 
\usepackage{dsfont}
\usepackage[round]{natbib}

\usepackage{titlesec}

\usepackage[colorlinks=true,
    linkcolor=mydarkblue,
    citecolor=mydarkblue,
    filecolor=mydarkblue,
    urlcolor=mydarkblue,
    pdfview=FitH]{hyperref}

\usepackage[ruled,vlined]{algorithm2e}
\SetKwBlock{Repeat}{repeat}{}

\usepackage{nicefrac}
\usepackage{booktabs}
\usepackage{multirow}
\usepackage{rotating}
\usepackage{caption}
\usepackage{multicol}
\usepackage{lipsum}
\usepackage{caption}
\usepackage{subcaption}

\definecolor{mydarkblue}{rgb}{0,0.08,0.45}
\definecolor{mydarkgreen}{RGB}{39,130,67}

\definecolor{mydarkred}{RGB}{192,47,25}

\newcommand{\norm}[1]{\left\lVert#1\right\rVert}

\definecolor{myblue}{HTML}{D2E4FC}
\definecolor{Gray}{gray}{0.92}

\newmdtheoremenv{theo}{Theorem}
\newtheorem{theorem}{Theorem}

\newtheorem{lemma}{Lemma}
\newtheorem{definition}{Definition}

\newtheorem{innercustomgeneric}{\customgenericname}
\providecommand{\customgenericname}{}
\newcommand{\newcustomtheorem}[2]{%
  \newenvironment{#1}[1]
  {%
  \renewcommand\customgenericname{#2}%
  \renewcommand\theinnercustomgeneric{##1}%
  \innercustomgeneric
  }
  {\endinnercustomgeneric}
}

\newcustomtheorem{customtheorem}{Theorem}
\newcustomtheorem{customlemma}{Lemma}

\graphicspath{{./figures/}}

\usepackage[accepted, nohyperref]{icml2020}

\icmltitlerunning{A Dynamic Sampling Adaptive-SGD Method for Machine Learning}

\begin{document}

\twocolumn[
\icmltitle{A Dynamic Sampling Adaptive-SGD Method for Machine Learning}



\icmlsetsymbol{equal}{*}

\begin{icmlauthorlist}
\icmlauthor{Achraf Bahamou}{to}
\icmlauthor{Donald Goldfarb}{to}
\end{icmlauthorlist}

\icmlaffiliation{to}{Department of Industrial Engineering and Operations Research, Columbia University, NY, USA}

\icmlcorrespondingauthor{}{ab4689@columbia.edu}

\icmlkeywords{Machine Learning, ICML}

\vskip 0.3in
]



\printAffiliationsAndNotice{} 

\begin{abstract}
We propose a stochastic optimization method for minimizing loss functions,  expressed as an expected value, that adaptively controls the batch size used in the computation of gradient approximations and the step size used to move along such directions, eliminating the need for the user to tune the learning rate. The proposed method exploits local curvature information and ensures that search directions are descent directions with high probability using an acute-angle test and can be used as a method that has a global linear rate of convergence on self-concordant functions with high probability. Numerical experiments show that this method is able to choose the best learning rates and compares favorably to fine-tuned SGD for training logistic regression and DNNs. We also propose an adaptive version of ADAM that eliminates the need to tune the base learning rate and compares favorably to fine-tuned ADAM on training DNNs. In our DNN experiments, we rarely encountered negative curvature at the current point along the step direction in DNNs.
\end{abstract}

\section{Introduction}

The problem of interest is minimizing for $x \in \mathbb{R}^{d}$ functions of the following form :
\begin{equation}\label{eq:prob}
 F(x) = \int f(x; z,y) dP(z,y) = \mathbb{E}[f(x ; \xi)],
\end{equation}
common to problems in statistics and machine learning. For instance, in the empirical risk minimization framework, a model is learned from a set
$$\left\{\xi_{1} = (z_{1},y_{1}), \dots, (\xi_{m} = z_{m},y_{m})\right\},$$ of training data by minimizing an empirical loss function of the form
\begin{equation}\label{eq:empmean}
    F(x) = \frac{1}{m} \sum_{i=1}^{m} f\left(x, (z_{i},y_{i})\right) = \frac{1}{m} \sum_{i=1}^{m} F_i(x),
\end{equation}
where we define $F_i(x) = f(x;z^i,y^i)$, and $f$ is the composition of a prediction function (parametrized by  $x $) 
and a loss function, and $(z_i,y_i)$ are random input-output pairs with the uniform discrete probability distribution $P(z_i,y_i) = \frac{1}{m}$. 
An objective function of the form \eqref{eq:prob} is often impractical, as the distribution of $\xi$ is generally unavailable, making it infeasible to analytically compute $ \mathbb{E}[f(x ; \xi)]$. This can be resolved by replacing the expectation by the estimate \eqref{eq:empmean}. The strong law of large numbers implies that the sample mean in \eqref{eq:empmean} converges almost surely to \eqref{eq:prob} as the number of samples $m$ increases.
However, in practice, even problem \eqref{eq:empmean} is not a tractable for classical optimization algorithms, as the amount of data is usually extremely large. A better strategy when optimizing \eqref{eq:empmean} is to consider sub-samples of the data to reduce
the computational cost. This leads to stochastic algorithms
where the objective function changes at each iteration by
randomly selecting sub-samples.


\section{Related work}

Stochastic gradient descent SGD~\citep{vanillasgd}, using either single or mini-batch samples is widely-used; however, it offers several challenges in terms of convergence speed. Hence, the topic of learning rate tuning has been extensively investigated. Modified versions of SGD based on the use of momentum terms~\citep{Polyak64};~\citep{NESTEROV83} have been showed to speed-up the convergence for SGD on smooth convex functions. In DNNs, other popular variants of SGD scale the individual components of the stochastic gradient using past gradient observations in order to deal with variances in the magnitude in the stochastic gradient components (especially between layers). Among these are ADAGRAD  ~\citep{Duchi2011}, RMSPROP  ~\citep{Tieleman2012}, ADADELTA  ~\citep{zeiler2012}, and ADAM  ~\citep{Kingma2014AdamAM}, as well as the structured matrix scaling version SHAMPOO ~\citep{Gupta2018ShampooPS} of ADAGRAD. These methods are scale-invariant but do not avoid the need for prior tunning of the base learning rate.

 A large family of algorithms exploit a second-order approximation of the objective function to better capture its local curvature and avoid the manual choice of a learning rate. Stochastic L-BFGS methods have been recently studied in ~\citep{zhougaodon} and ~\citep{bollapragadaBFGS}. Although fast in theory (in terms of the number of iterations), they are more expensive in terms of iteration cost and memory requirement, thus they introduce a considerable overhead which makes the application of these approaches usually less attractive than first-order methods for modern deep learning architectures. Other Hessian approximation-based methods have been also proposed in ~\citep{hessianfree},~\citep{MartensG15},~\citep{tonga},~\citep{naturalg} that are based on the Gauss-Newton and natural gradient methods. However, pure SGD is still of interest as it often has better generalization properties ~\citep{sgdgen}.

\section{Our contributions}

- A new stochastic optimization method is proposed that uses an adaptive step size that depends on the local curvature of the objective function eliminating the need for the user to perform prior learning rate tuning or perform a line search.

- The adaptive step size is computed efficiently using a Hessian-vector product (without requiring computing the full Hessian ~\citep{Pearlmutter}) that has the same cost as a gradient computation. Hence, the resulting method can be applied to train to DNNs using backpropagation.

- A dynamic batch sampling strategy is proposed that increases the batch size progressively, using ideas from ~\citep{Bollapragada2017AdaptiveSS} but based upon a new "acute-angle" test. In contrast, the approach in \citep{zhougaodon} uses empirical processes concentration inequalities resulting in worst-case lower-bounds on batch sizes that increase rapidly to full batch.

- Our Adaptive framework can also be combined with other variants of SGD (Momentum, ADAM, ADADELTA..etc). We present an adaptive version of Momentum and ADAM and show numerically that the resulting methods are able to select excellent base learning rates.

- An interesting empirical observation in our DNN experiments was that, using our Adaptive framework on  convolutional and residual neural networks, we rarely encountered negative local curvature along the step direction.

\section{A Dynamic Sampling Adaptive-SGD Method}
\label{sec:progressive}

\subsection{Definitions and notation}
We propose an iterative method of the following form. At the k-th iteration, we draw $|S_k|$ i.i.d samples $S_k = \{ \xi_{1}, \dots, \xi_{|S_{k}|} \} $ and define the empirical objective function

\begin{equation*}
    F_{k}(x) = F_{S_{k}}(x)=\frac{1}{\left|S_{k}\right|} \sum_{i \in S_{k}} f\left(x, \xi_{i}\right) =\frac{1}{\left|S_{k}\right|} \sum_{i \in S_{k}} F_{i}(x).
\end{equation*}
An approximation to the gradient of $F$ can be obtained by sampling. Let $X_k$ denote the set of samples used to estimate the Hessian, and let 

\begin{equation*}
    g_{k} = g_{S_{k}}(x_{k}) = \nabla F_{S_{k}}\left(x_{k}\right)=\frac{1}{\left|S_{k}\right|} \sum_{i \in S_{k}} \nabla F_{i}\left(x_{k}\right).
\end{equation*}

An approximation to the Hessian of $F$ can also be obtained by sampling. At the iterate $x_k$, we define

\begin{equation*}
    G_{k} = G_{X_{k}}(x_{k}) = \nabla^2 F_{X_{k}}\left(x_{k}\right)=\frac{1}{\left|X_{k}\right|} \sum_{i \in X_{k}} \nabla^2 F_{i}\left(x_{k}\right).
\end{equation*}
A first-order method based on these approximations is then given by

\begin{equation} \label{genersgd}
    x_{k+1}=x_{k}-t_{k} \nabla F_{S_{k}}\left(x_{k}\right),
\end{equation}
where, in our approach, $t_{k}$ is the adaptive step size, and the sets of samples $S_k$ and $X_k$ that are used to estimate the gradient and Hessian adaptively grow as needed,  and are drawn independently. In the following sections,  we will describe how we choose the step size $t_{k}$ and the batch sizes $|S_{k}|$ and $|X_{k}|$. 
We use the following notation:
\begin{equation*}
   d_k = - g_k;\hskip0.5em \delta_k=\|d_k\|_{x_k}=\sqrt{g_k^{T} \nabla^2 F\left(x_{k}\right) g_k}
\end{equation*}

\begin{definition}
A convex function $f : \mathbb{R}^{n} \rightarrow \mathbb{R}$ is self-concordant if there
exists a constant $\kappa$ such that for every $x \in \mathbb{R}^{n}$ and every $h \in \mathbb{R}^{n}$, we have :
\begin{equation*}
    \left|\nabla^{3} f(x)[h, h, h]\right| \leq \kappa\left(\nabla^{2} f(x)[h, h]\right)^{3 / 2},
\end{equation*}
$f$ is standard self-concordant if the above is satisfied for $\kappa = 2$.
\end{definition}
Self-concordant functions were introduced by Nesterov and Nemirovski in the context of interior-point methods ~\citep{Nesterov1994InteriorpointPM}. Many problems in machine learning have self-concordant formulations: In ~\citep{zhangb15disco} and  ~\citep{Bach2009SelfconcordantAF}, it is shown that regularized regression, with either logistic loss or hinge loss, is self-concordant.

\subsection{Assumptions}

We first give the intuition behind the sampling strategies and step-size used in our method. The formal proof of the convergence of our algorithm will be given in the Appendix. Throughout this section, our theoretical analysis is based on the following technical assumptions on $F$:
\begin{enumerate}

  \item \textbf{Hessian regularity:} $ \exists M \geq m>0$ s.t $\forall x \in \mathbb{R}^{n}$:
$$
m I \preceq \nabla^{2} F(x) \preceq M I.
$$
 The condition number, denoted $\kappa,$ is given by $M / m$ .
  \item \textbf{Uniform Sampled Hessian regularity:} For the above defined $M>0$, for every sample set $S$ and 
$x \in \mathbb{R}^{n}$, we have :
$$
0 \preceq \nabla^{2} F_S(x) \preceq M I.
$$
\item \textbf{Gradient regularity:} We assume that the norm of the gradient of $F$ stays bounded above during the optimization procedure:
  $$\exists \gamma>0  \forall k \geq 0: \|\nabla F\left(x_k\right)\| \leq \gamma. $$
 \item \textbf{Function regularity:} $F$ is standard self-concordant.
 
\end{enumerate}

\subsection{Analysis}
Consider the iterative method for minimizing self-
concordant functions described in \eqref{genersgd} using the step-size
$t_{k}^*=\frac{\rho_{k}}{\left(\rho_{k}+\delta_{k}\right) \delta_{k}}$, where $\delta_k=\|d_k\|_{x_k}=\sqrt{g_k^{T} G(x_k) g_k}$ and $\rho_{k}=g_{k}^{T} g$. Methods of this type have been analyzed in ~\citep{TranDinh2015CompositeCM} and ~\citep{gaogoldself} in the deterministic setting. In the latter paper, the above choice of $t_{k}$ is shown to guarantees a decrease in the function value.

\begin{lemma} \label{selfconc}
(Lemma 4.1, ~\citet{gaogoldself})

For $F$ standard self-concordant, for all $0 \leq t<\frac{1}{\delta_k}$:
\begin{equation*}
\quad F(x_k+t d_k) \leq F(x_k)+t g(x_k)^{T} d_k-\delta_k t-\log (1-\delta_k t).
\end{equation*}
\end{lemma}

Let $\Delta (\delta_k, t) = (\delta_k + \rho_k)t + log(1-\delta_k t)$, the above statement is equivalent to :
\begin{equation}\label{eq:gaodec}
\quad F(x_k+t d_k) \leq F(x_k)-\Delta (\delta_k, t).
\end{equation}
If $t = t_{k}^*$ then $F\left(x_{k}-t_{k} g_{k}\right) \leq F\left(x_{k}\right)-\omega\left(\eta_{k}\right)$, where $\eta_{k}=\frac{\rho_{k}}{\delta_{k}}  \text { and } \omega(z)=z-\log (1+z)$.

Since we do not want to compute the full Hessian at $x_k$, we cannot compute $\delta_k$ and $t_{k}^*$ (which is the true value of the argmax of $\Delta(t)$). Instead, we use their respective estimates using a sub-sampled Hessian:
\begin{equation*}
  \hat{\delta}_k=\sqrt{g_k^{T} \nabla^2 F_{X_{k}}\left(x_{k}\right) g_k}; \hskip0.5em \hat{t}_k = \frac{\rho_{k}}{\left(\rho_{k}+\hat{\delta}_{k}\right) \hat{\delta}_{k}}.
\end{equation*}
Consequently the update step in practice will be : 

\begin{equation*}\label{eq:genersgd}
    x_{k+1}=x_{k}-\hat{t}_k \nabla F_{S_{k}}\left(x_{k}\right).
\end{equation*}

We will initially study the behaviour of the estimated values $g_k, \hat{\delta}_k$, using the framework described in ~\citep{Roosta-Khorasani2019}, that shows that, for gradient and Hessian sub-sampling, using random matrix concentration inequalities, one can sub-sample in a way that first and second-order information, i.e., curvature, are well estimated with probability of at least $1-p$, where $0 < p < 1$ can be chosen arbitrarily.

\begin{lemma}  \label{dhdprob}
(Lemma 2 in ~\citep{Roosta-Khorasani2019}
Under Assumptions 1. and 2., given any $ 0<\varepsilon, p<1 \text { and } x, d \in \mathbb{R}^{n}$, if $|X| \geq 16 \kappa^{2} \ln (2 n / p) / \varepsilon^{2},$ we have:
\begin{equation}\label{eqn:conchess}
    \mathbb{P}\left(\left\| \nabla^{2} F_X(x)-\nabla^{2} F(\mathbf{x})\right\| \leq \varepsilon m\right) \geq 1-p,
\end{equation}
where $\|.\|$ is the operator norm induced by the Euclidean norm.
\end{lemma}

\begin{lemma}  \label{gkprob}
(Lemma 3 in ~\citep{Roosta-Khorasani2019}
If for any $x \in \mathbb{R}^{n}$ and any \(0<\nu \left\|\nabla F_{S}(\mathbf{x})\right\| , \delta <1\) and $|S| \geq\left[\frac{G(\mathrm{x})}{\left.\left\|\nabla F_{S}(\mathrm{x})\right\| \nu\right)}\right]^{2}(1+\sqrt{8 \ln \frac{1}{\delta}})^{2}$, then
$$
\mathbb{P}\left(\left\|\nabla F(\mathbf{x})-\nabla F_{S}(x)\right\| \leq \nu\left\|\nabla F_{S}(\mathbf{x})\right\|\right) \geq 1-\delta,
$$
where \(\left\|\nabla F_{i}(\mathrm{x})\right\| \leq G(\mathrm{x}), \quad \forall i=1, \ldots, n\)
\end{lemma}

If we sample the Hessian at the rate described in Lemma ~\ref{dhdprob}, the following inequalities are satisfied with probability of at least $1-p$
\begin{equation*}
    \frac{\hat{\delta_k}}{\sqrt{1+\varepsilon}} \leq \delta_k \leq \frac{\hat{\delta_k}}{\sqrt{1-\varepsilon}} = \hat{\delta}_{k, \varepsilon}.
\end{equation*}

This allows us to estimate the curvature along an arbitrary direction $d$ by sub-sampling the local norm squared, $d^{T} \nabla^2 F \left(x\right) d$, with precision $\varepsilon$. This combined with Lemma \ref{selfconc} and the use of step size $t_{k, \varepsilon} = \frac{\rho_{k}}{\left(\rho_{k}+\hat{\delta}_{k, \varepsilon}\right) \hat{\delta}_{k, \varepsilon}}$, guarantees a decrease in $F(.)$ with probability of at least $1-p$ that
\begin{equation*}
     F(x+t_{k, \varepsilon} d_k)  \leq F(x) -  \frac{1-\varepsilon}{2M(1+\Gamma)} (\frac{g_k^T g}{\|g_k\|})^2.
\end{equation*}

Using the gradient sampling rate in Lemma \ref{gkprob}, we can control the stochasticity of the right most term above to obtain
\begin{theorem}
    Let $0 <\nu\gamma, \varepsilon, p, \delta < 1$. Suppose that $F$ satisfies the Assumptions 1-4 and that we have an efficient way to compute $G(x)$ where \(\left\|\nabla F_{i}(\mathrm{x})\right\| \leq G(\mathrm{x}), \quad \forall i=1, \ldots, n\). Let $\{x_k, X_k, S_k\}$ be the set of iterates, and sample Hessian and gradient batches generated by taking the step $t_{k, \varepsilon}$ at iteration $k$ starting from any $x_0$, where $|S_{k}|, |X_{k}|$ are chosen such that Lemma \ref{gkprob} and Lemma \ref{dhdprob} are satisfied at each iteration. Then, for any $k$, with probability $(1-p)(1-\delta)$ we have :
	\begin{equation*}
	    (F(x_{k+1}) - F(x^*)) \leq \rho (F(x_k) - F(x^*)),
	\end{equation*}
     where $\rho=  1 - \frac{m(1-\nu)^2(1-\varepsilon)}{2(1+\nu^2) M (1+\frac{\gamma}{\sqrt{m}})}.$
	\label{theorem1}
\end{theorem}

As pointed out in ~\citep{Roosta-Khorasani2019}, for many generalized linear model (that are also
self-concordant), the upper bounds $G(x)$ and $\gamma$ are known. However, computing these values for other settings, (e.g. DNNs) can be a challenging task. On the other hand, the above method often leads to a large batch-size since we are requiring very strong guarantees on the quality of the estimates. More specifically, we require the estimated curvature along {\it any} direction (not just the descent direction) to be close to the true curvature, as well as the sampled gradient to be close (in norm) to the full gradient which is overkill to controlling the error in the term $(\frac{g_k^T g}{\|g_k\|})^2$.

In the following sections, we will refine the batch-size strategy to be less aggressive and we will provide heuristics to achieve such conditions.

\subsection{Batch-size strategy}

\subsection{Gradient sub-sampling}

We propose to build upon the strategy introduced by ~\citep{Bollapragada2017AdaptiveSS} in the context of first-order methods. Their \textit{inner product test} determines a sample size such that the search direction is a descent direction with high probability. To control the term $(\frac{g_k^T g}{\|g_k\|})^2$, we propose instead, to choose a sample size $|S_k|$ that satisfies the following new test that we call the \textit{acute-angle test}:

\begin{equation}\label{eq:exactest}
\mathbb{E}\big( \| \frac{g_k}{\|g_k\|} - 
\frac{g_k^{T} g}{\|g\|\|g\|^{2}} g\|^{2} \big)  \leq p \nu^2.
\end{equation}

\begin{figure} 
\centering
\includegraphics[scale=0.2]{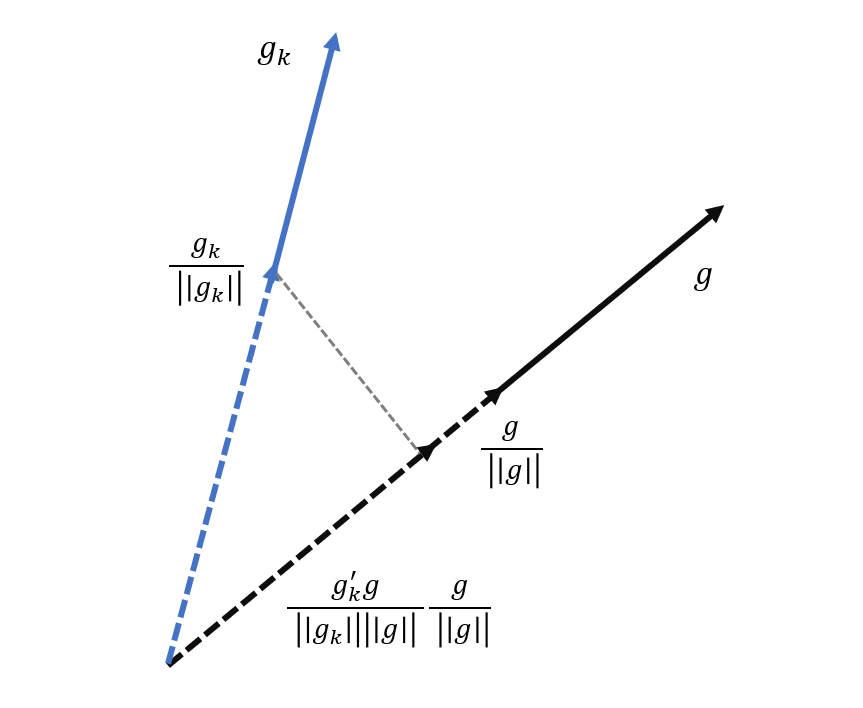}
\caption{projection of the normalized sampled gradient on the normalized true gradient}
\label{fig:gradfig}
\end{figure}

\begin{figure}[htp]
\centering
  \begin{subfigure}[b]{0.33\linewidth}
    \includegraphics[scale=0.3]{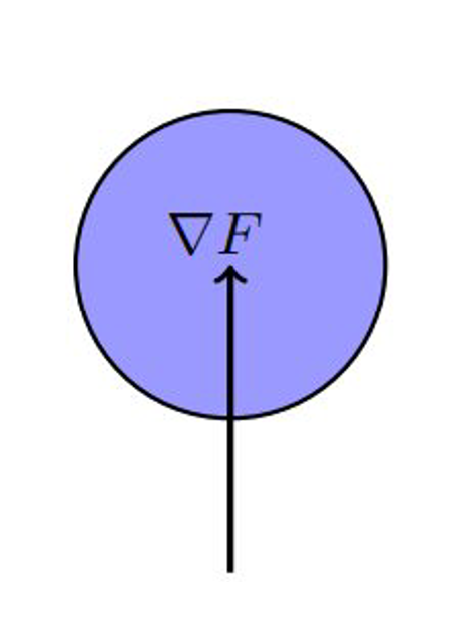}
    \caption{Norm test}
    \label{fig:figure1}
  \end{subfigure}%
  \hfill
  \begin{subfigure}[b]{0.33\linewidth}
    \includegraphics[scale=0.3]{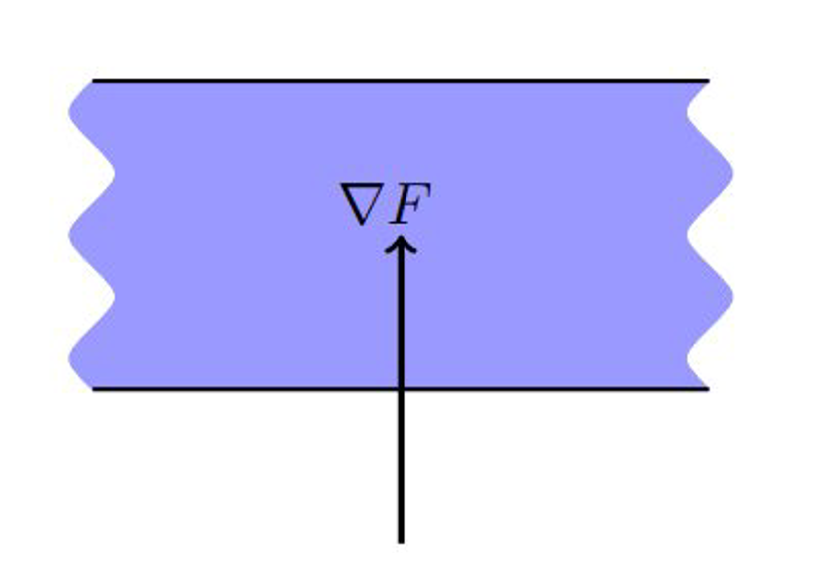}
    \caption{Inner-product}
    \label{fig:figure2}
  \end{subfigure}%
  \hfill
  \begin{subfigure}[b]{0.33\linewidth}
    \includegraphics[scale=0.3]{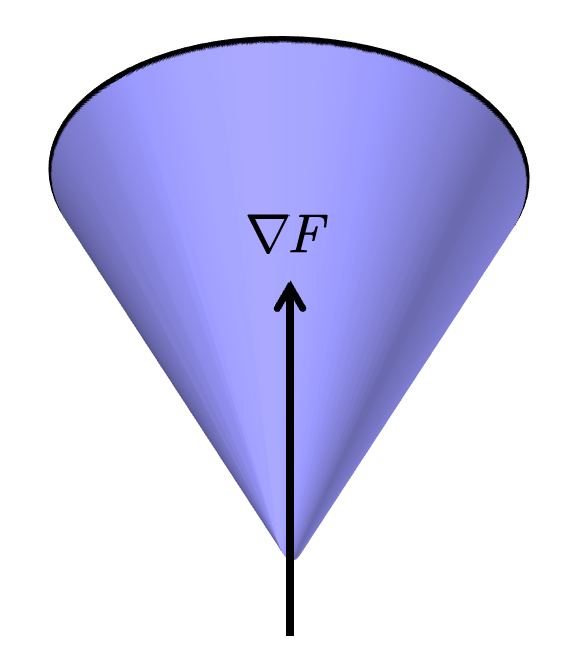}
    \caption{Acute-angle}
    \label{fig:figure3}
  \end{subfigure}
  \caption{Given a gradient $\nabla F$, the shaded areas denote the set
of vectors $g$ satisfying (a) the norm test \citep{Byrd2012SampleSS}; (b) the inner product test \citep{Bollapragada2017AdaptiveSS}; (c) our acute-angle test \eqref{eq:exactest}}
\label{fig:tests}
\end{figure}

Geometrically, this ensures that the normalized sampled  gradient is close to its projection on the true normalized gradient (see Figures~\ref{fig:gradfig}, \ref{fig:tests}). 

The left-hand side is difficult to compute and the computation of $\nabla F(x_k)$ can be prohibitively expensive but can be approximated using a running average $g_{avg}$ of the sampled gradient. Therefore we propose the following approximate version of the test
(using $g_k = \nabla F_{S_{k}}\left(x_{k}\right)$, $g_{i,k} = \nabla F_{i}\left(x_{k}\right)$ for $i \in S_k$):
\begin{align}\label{eq:approxtest}
\frac{1}{\left|S_{k}\right|} \frac{\sum_{i \in S_{k}}\norm{\frac{g_{i,k} }{\norm{g_{k}}} - \frac{g_{i,k}^{T} g_{avg}}{ \norm{g_{k}} \norm{g_{avg}}} \frac{g_{avg}}{\norm{g_{avg}}}}^2}{\left|S_{k}\right|-1} \leq p\nu^{2}.
\end{align}

We refer to~\eqref{eq:exactest} as the \textit{exact acute-angle test} and ~\eqref{eq:approxtest} as the \textit{approximate acute-angle test}. This approximate test will serve as an update criteria for the sample size. In other words, if the condition above is not satisfied we take as a new batch size:

\begin{align}
\left|S^g_{k+1}\right| = \frac{\sum_{i \in S_{k}}\norm{\frac{g_{i,k} }{\norm{g_{k}}} - \frac{g_{i,k}^{T} g_{avg}}{ \norm{g_{k}} \norm{g_{avg}}} \frac{g_{avg}}{\norm{g_{avg}}}}^2}{(\left|S_{k}\right|-1)p\nu^{2}}.\label{gupdate}
\end{align}

We can show, using Markov's inequality, that if ~\eqref{eq:exactest} is satisfied, then with probability at least $1-p$, we have:
\begin{align*}
    \left(\frac{(\nabla F_{S_k}(x_k)^{T} \nabla F(x_k))^2}{ \norm{\nabla F_{S_k}(x_k)}^2}\right)  \geq (1-\nu^2) \norm{\nabla F(x_k)}^2.
\end{align*}

\subsection{Hessian sub-sampling}
Computing the Hessian sub-sampling size $|X|$ described in Lemma \ref{dhdprob} requires $|X| \geq 16 \kappa^{2} \ln (2 n / p) / \varepsilon^{2}$ for given $\delta$, $\varepsilon$ can be difficult in practice to ensure in complex models such as DNNs where the condition number $\kappa$ is hard to compute and where the above bound could lead to very large batch-sizes. Instead of computing the the Hessian sub-sampling set such that the concentration inequality \eqref{eqn:conchess} holds, we use the sample set $S_k$ used to compute the stochastic gradient and we choose $|S_k|$ such that:
\begin{equation}\label{eqn:conchessP}
    \mathbb{P}((\hat{\delta_k}^2 - \delta_k^2)^2 \leq \varepsilon^2  \delta_k^4) \geq 1-p.
\end{equation}
 Inequality \eqref{eqn:conchessP} is less restrictive than \eqref{eqn:conchess} where the precision on the curvature is required for any arbitrary direction. Using Markov's inequality and similar approximation ideas introduced by \citep{Bollapragada2017AdaptiveSS}, we can derive an approximate test version and update formula for $|S_k|$ such that \eqref{eqn:conchessP} is satisfied:
 \begin{equation}
    \mathbb{E}((\hat{\delta_k}^2 - \delta_k^2)^2) \leq p\varepsilon^2  \delta_k^4 \label{dhdtest},
\end{equation}
 \begin{equation}
|S^h_{k+1}| =  \frac{1}{\varepsilon^{2} (\left|S_{k}\right|-1)} \frac{\sum_{i \in S_{k}}(\hat{\delta}_{i,k}^2 - \hat{\delta}_{k}^2)^2}{p \hat{\delta}_{k}^4},
\label{hupdate}
\end{equation}
with $\hat{\delta}_{i,k}^2 = g_k \nabla_i^2 F(x_k) g_k$ for $i \in S_k$ and the expectation is conditioned on $g_k$. The intuitive interpretation of this technique is that we penalize the step size with a factor $\sqrt{1-\varepsilon}$ such that $\varepsilon$ encodes the information on how accurate the sub-sampled Hessian preserves the curvature information along $g_k$ compared to the full Hessian.

This approximate approach to update the batch-size is given in the Algorithm \ref{alg:sampleupdate}.
\begin{algorithm}[h]
    \KwIn{Current vector $x_k$, current sample size $|S_k|$, constant  $\nu> 0$.}
    \KwResult{Returns $S_{k+1}$ that satisfies gradient and Hessian tests.}
	$S_k \leftarrow$ Randomly sample $|S_k|$ samples \\
    $N_{k+1} = \max(|S^g_{k+1}|, |S^h_{k+1}|)$ as in \eqref{gupdate} and \eqref{hupdate} \\
    \If{$N_{k+1} > |S_k|$}{
        Sample additional $N_{k+1} - |S_k|$ samples $A_{k+1}$\\
        $S_{k+1} \leftarrow S_{k} U A_{k+1}$ 
    }
    \Else{
        $S_{k+1} \leftarrow S_{k}$
    }
    \caption{Batch-size update}\label{alg:sampleupdate}
\end{algorithm}

\subsection{Dynamic Sampling Adaptive-SGD}

Based on the update criteria developed above we propose Algorithm \ref{alg:adasgd}.

\begin{algorithm}[h]
    \KwIn{Initial iterate $x_0$, initial gradient sample  $S_0$, and constants $1 >p, \varepsilon, \nu> 0$, max number of iterations $N$.}
    $k \leftarrow 0$ \\
    \While{$k < N$}{
        Choose samples $S_k$  using Algorithm~\ref{alg:sampleupdate} \\
		Compute $\hat{d}_k = -\nabla F_{S_k}(x_k)$ \\
		Compute $\rho_k = \nabla F_{S_{k}}\left(x_{k}\right)^{T} \nabla F\left(x_{k}\right)$ \\
		Compute $\hat{\delta}_k =\sqrt{\nabla F_{S_{k}}\left(x_{k}\right)^{T} \nabla^2 F_{S_{k}} (x_k) \nabla F_{S_{k}}\left(x_{k}\right)}$ \\
		Compute $t_{k, \varepsilon} = \frac{\rho_k}{\left(\rho_k+\hat{\delta}_{k, \varepsilon}\right) \hat{\delta}_{k, \varepsilon}} $ \\
		Compute new iterate: $x_{k+1} = x_k + t_{k, \varepsilon} \hat{d}_k$ \\
		Set $k \leftarrow k + 1$ 
    }
    \caption{Progressive sampling and Adaptive step-size SGD}\label{alg:adasgd}
\end{algorithm}

Practical aspects of the computations in Algorithm \ref{alg:adasgd} will be discussed in section \eqref{practical}.

\subsection{Convergence theorem:}
In this section, we state the main lemma and theorem which prove global convergence with high probability of Algorithm \ref{alg:adasgd}.
The acute-angle test, in conjunction with the Hessian sub-sampling allows the algorithm to make sufficient progress with high probability at every iteration if $p$ is chosen very small.\\

\begin{lemma}
	Suppose that $F$ satisfies Assumptions 1-4,
	Let $\{x_k\}$ be the iterates generated by the idealized Algorithm~\ref{alg:adasgd} where $|S_{k}|$ is chosen such that the (exact) acute-angle test~\eqref{eq:exactest} and Hessian sub-sampling test ~\eqref{dhdtest} are satisfied at each iteration for any given constants $0 <\nu, \varepsilon, p < 1$. Then, for all $k$, starting from any $x_0$, 
	\begin{align*}\label{c-sample-norm}
	    \frac{(\nabla F_{S_k}(x_k)^{T} \nabla F(x_k))^2}{ \norm{\nabla F_{S_k}(x_k)}^2}  \geq (1-\nu^2) \norm{\nabla F(x_k)}^2.
	\end{align*}
	with probability at least $1-p$. Moreover, with probability $(1-p)^2$, we have
	\begin{equation*}
        F(x_{k+1}) \leq F(x_{k}) - \frac{\alpha}{2} \|\nabla F(x_k)\|^2,
    \end{equation*}  
	where $\alpha = \frac{(1-\nu^2)(1-\varepsilon)}{M (1+\frac{\gamma}{\sqrt{m}})} > 0.$
	\label{mainlemma}
\end{lemma}

The proof of the above lemma follows the non-asymptotic probabilistic analysis of sub-sampled Newton methods in \citep{Roosta-Khorasani2019}, that allows for a small, yet non-zero, probability of occurrence of “bad events” in each iteration. Hence, the accumulative probability of occurrence of “good events” decreases with each iteration. Although the term “convergence” typically implies the asymptotic limit of an infinite sequence, we analyse here the non-asymptotic behavior of a finite number of random iterates and provide probabilistic results about their properties. Our main convergence theorem is:

\begin{theorem}
	Suppose that $F$ satisfies Assumptions 1-4,
	Let $\{x_k\}$ be the iterates generated by the idealized Algorithm~\ref{alg:adasgd} where $|S_{k}|$ is chosen such that the (exact) acute-angle test~\eqref{eq:exactest} and Hessian sub-sampling test ~\eqref{dhdtest} are satisfied at each iteration for any given constants $0 <\nu, \varepsilon, p < 1$. Then, for all $k$, starting from any $x_0$, with probability of at least $(1-p)^{2k}$, we have
	\begin{equation*} 
	    F(x_k) - F(x^*) \leq \rho^k (F(x_0) - F(x^*)),
	\end{equation*}
     where $\rho= 1 -m \alpha =  1 - \frac{m(1-\nu^2)(1-\varepsilon)}{M (1+\frac{\gamma}{\sqrt{m}})}.$
	\label{theorem2}
\end{theorem}

In order to obtain a non-zero probability of asymptotic convergence, we can decrease the probability p of failure at a rate faster that $1/k$ and the above result will hold with non-zero probability asymptotically at the expense of a more aggressive batch-size increasing strategy.

\section{Momentum and Scale-invariant versions: Ada-SGD with Momuntum, Ada-ADAM}

    A number of popular methods, especially in deep learning, choose per element update magnitudes based on past gradient observations or use momentum-accelerated directions. Our adaptive framework is general enough to combine with variants of SGD such as ADAM, SGD with momentum, ADADELTA and ADAGRAD, as well as second-order methods such as Block BFGS \citep{Gower2016StochasticBB}, \citep{Gao2016BlockBM}.

Since our framework can be applied to any direction $d_k$, we propose the Adaptive learning rate version of ADAM and SGD-with momentum in Appendix J.1 and J.2.

\section{Practical considertations}\label{practical}

\subsection{Hessian vector product}
In our framework, we need to compute the curvature along a direction $d_k$ given by : $\delta_k = \sqrt{d_k^T H_k d_k}$. Hence we need to efficiently compute the Hessian-vector product $H_k d_k$. Fortunately, for functions that can be computed using a computational graph (Logistic regression, DNNs, etc) there are automatic methods available for computing Hessian-vector products exactly~\citep{Pearlmutter}, which take about as much computation as a gradient evaluation. Hence, if $d_k$ is a mini-batch stochastic gradient, $H_k d_k$ can be computed with essentially the same effort as that needed to compute $d_k$. The method described in~\citep{Pearlmutter} is based on the differential operator:
$$\mathcal{R}\{F(w)\}=\left.(\partial / \partial r) F(w+r v)\right|_{r=0}.$$
Since $\mathcal{R}\left\{\nabla_{w} F\right\}=H v$ and $\mathcal{R}\{w\}=v$, to compute $H v$,~\citet{Pearlmutter} applies $\mathcal{R}$ to the back-propagation equations used to compute $\nabla_{w} F$.

\subsection{Individual gradients in Neural Network setting}

Unfortunately, the differentiation functionality provided by most software frameworks (Tensorflow, Pytorch, etc.) does not support computing gradients with respect to individual samples in a mini-batch making it expensive to apply the adaptive batch-size framework in that setting. On the other hand, it is usually impractical to substantially increase the batch-size due to memory-limitations. As a result, we only compute the adaptive learning rate part of our framework in the neural network setting at fixed milestone epochs for a small number of iterations (similar to learning rate schedulers that are widely used in practice) and we fix the batch-size between milestones, thus the problem of mini-batch computation does not arise.

\section{Numerical Experiments}
In this section, we present the results of numerical experiments comparing our Adaptive-SGD method to vanilla SGD with optimized learning rate, SGD with inner-product test,  SGD with augmented inner-product test, and SGD with norm test to demonstrate the ability of our framework to capture the "best learning rate" without prior tuning as well as the effectiveness and cost of our method against other progressive sampling methods. In all of our experiments, we approximate $\rho_k = g_k^T g$ by $\rho_k = g_k^{T} g_k^{avg}$ where $g_k^{avg}$ is a running average ($\beta = 0.9$) that approximates the full gradient. We fixed $p = 0.1, \nu = 0.1, \varepsilon = 0.01$;  i.e., we wanted a decrease guarantee to be true with probability $0.9$ and $\mathbb{E}(sin(\theta_k)^2) \leq 0.001$, where $\theta_k$ is the angle between the gradient and the sampled gradient at iteration $k$. We also decreased $p$ by  a factor 0.9 every 10 iterations to ensure that the cumulative probability does not converge to 0. These parameters worked well in our experiments; of course one could choose more restrictive parameters but they would result in a more aggressive batch-size increase. For more details on the experiments hyper-parameters (batch-size, DNN architectures..etc), please refer to Appendix J.

\subsection{Datasets}

For binary classification using logistic regression, we chose 6 data sets from LIBSVM \citep{libsvm} with a variety of dimensions and sample sizes, which are listed in details in Appendix J.3.

For our DNN experiments, we experimentally validated the ability of the adaptive framework Ada-SGD and Ada-ADAM to determine a good learning rate without prior fine tuning on a two-layer convolutional neural network with a ReLU activation function and batch normalization on the MNIST Dataset \citep{mnist}; ResNet18 \citep{resnetpaper}, VGG11 \cite{VGGNet}, and a custom ResNet-9 (DavidNet) on CIFAR10 Dataset \citep{cifar}. Please refer Appendix J for more details.

\subsection{Binary classification logistic regression}
We considered binary classification problems where the objective function is given by the logistic loss with $\ell_2$ regularization, with $\lambda = \frac{1}{N}$, $R(x)=\frac{1}{N} \sum_{i=1}^{N}\log(1 + \exp(-z^ix^Ty^i)) + \frac{\lambda}{2}\|x\|^2.$ 

\subsubsection{Comparison of different batch-increasing strategies}

\begin{figure}
  \includegraphics[width=80mm]{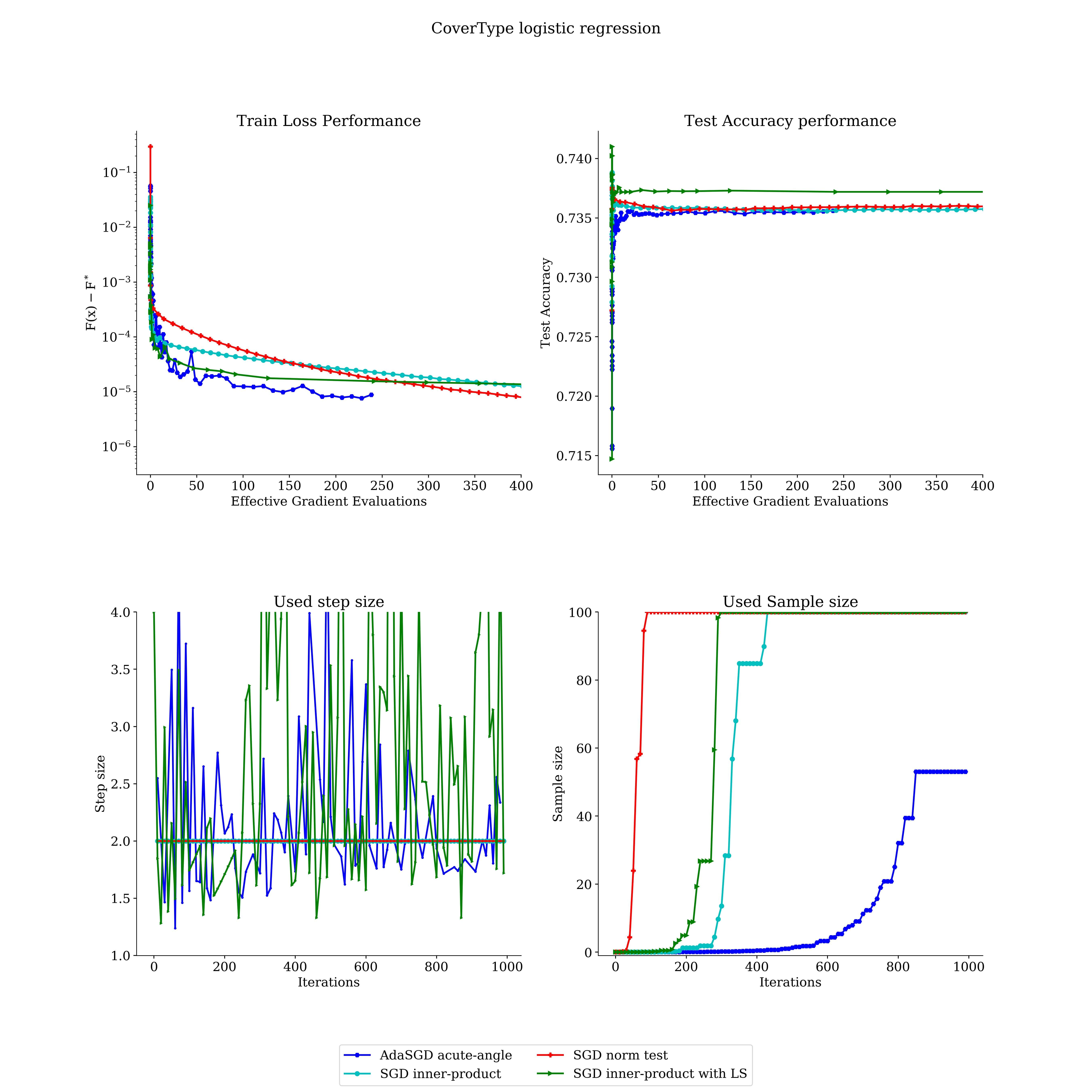} 
\caption{Numerical results for logistic regression on Covertype dataset}
\label{covtype}
\end{figure}

\begin{figure}
\begin{tabular}{cc}
  \includegraphics[width=41mm]{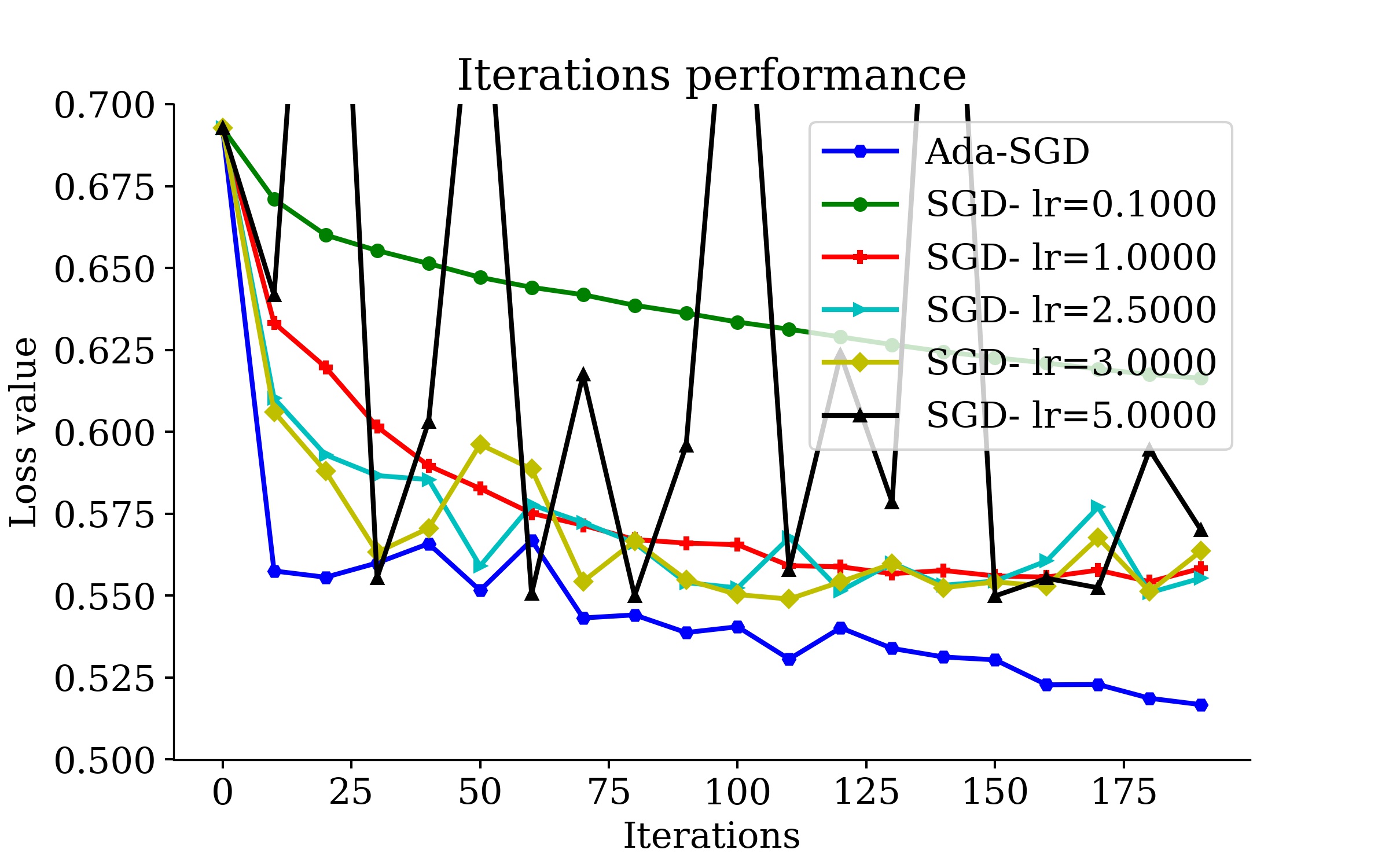} & \includegraphics[width=41mm]{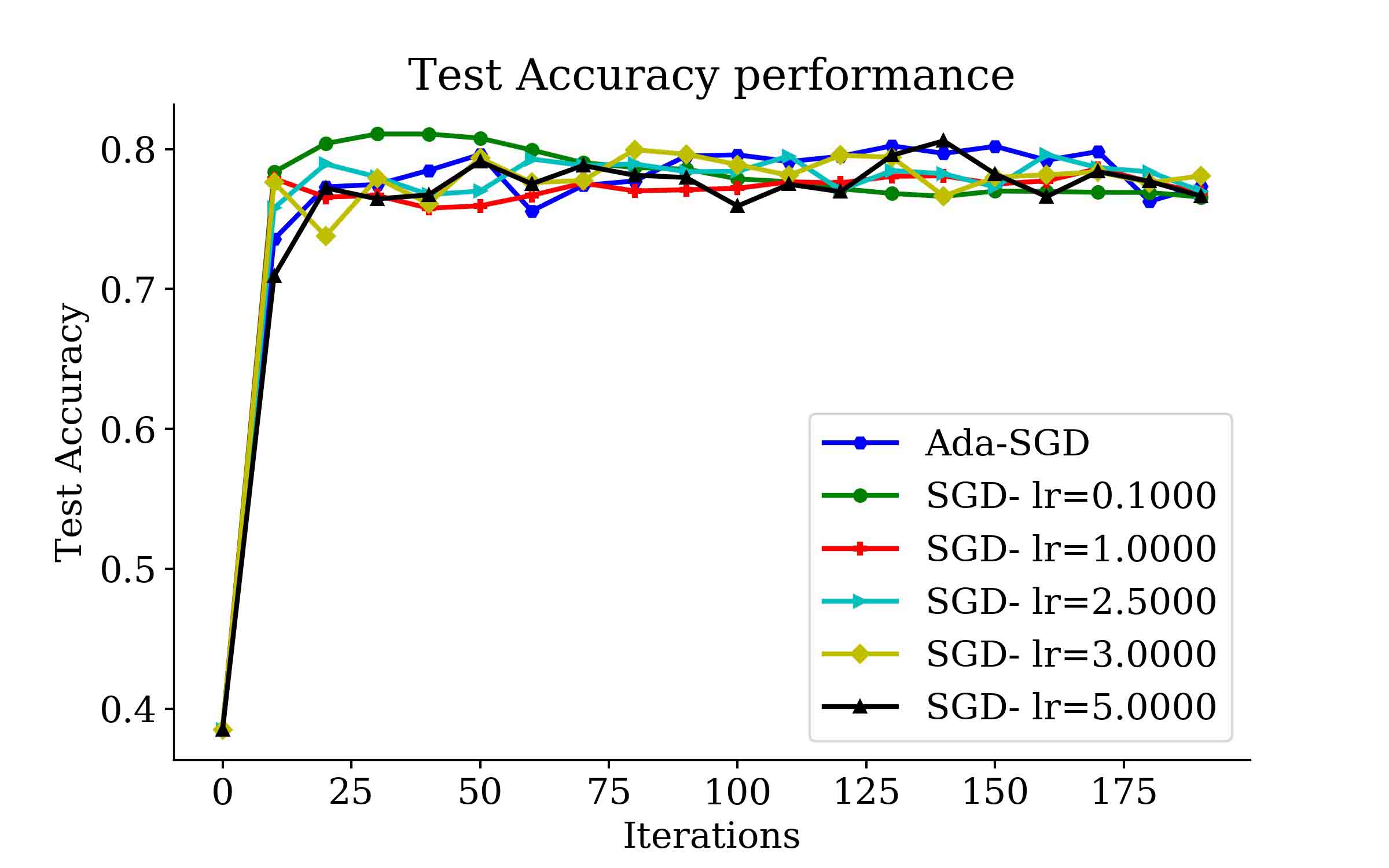} \\ \includegraphics[width=41mm]{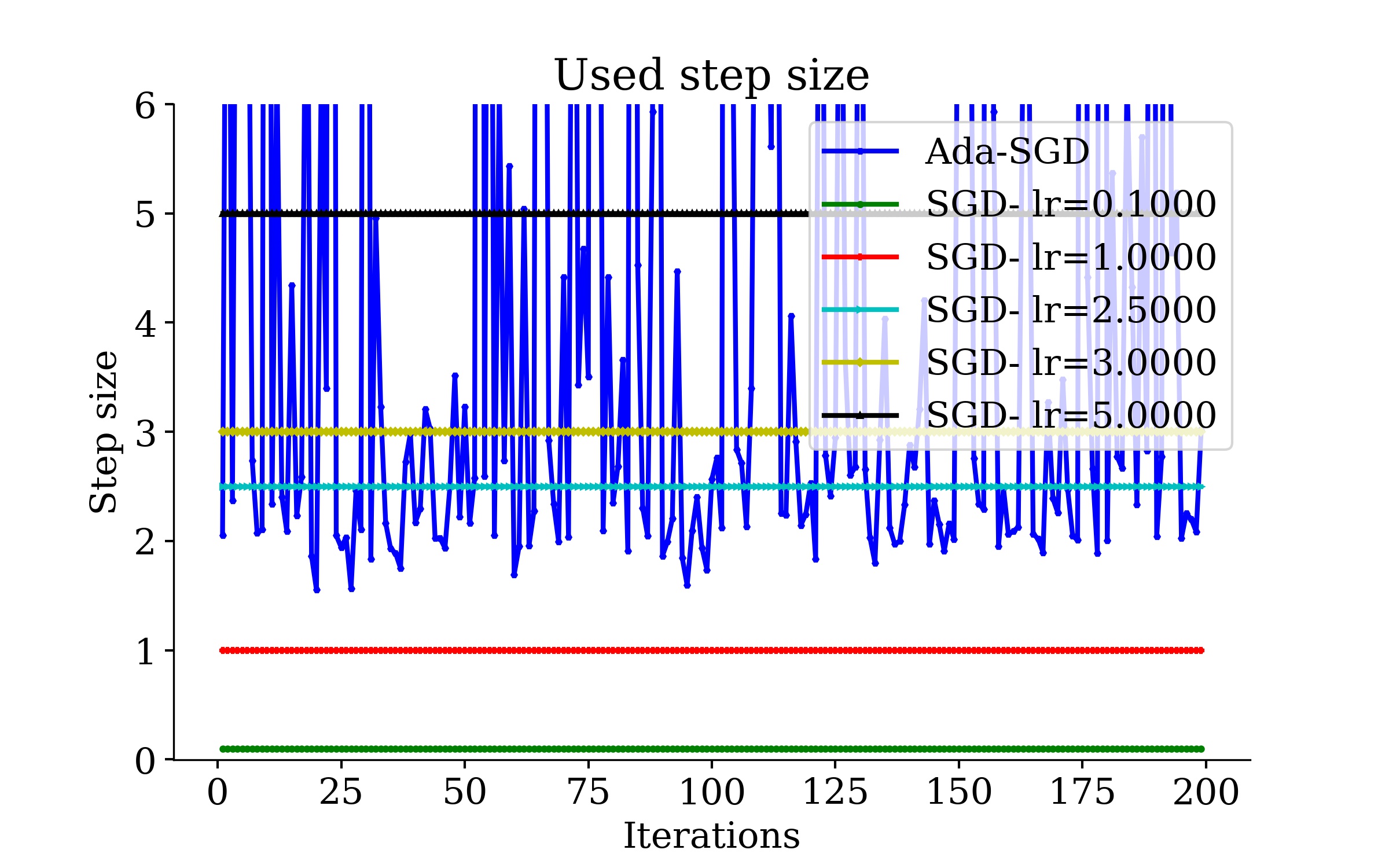} &
 \includegraphics[width=41mm]{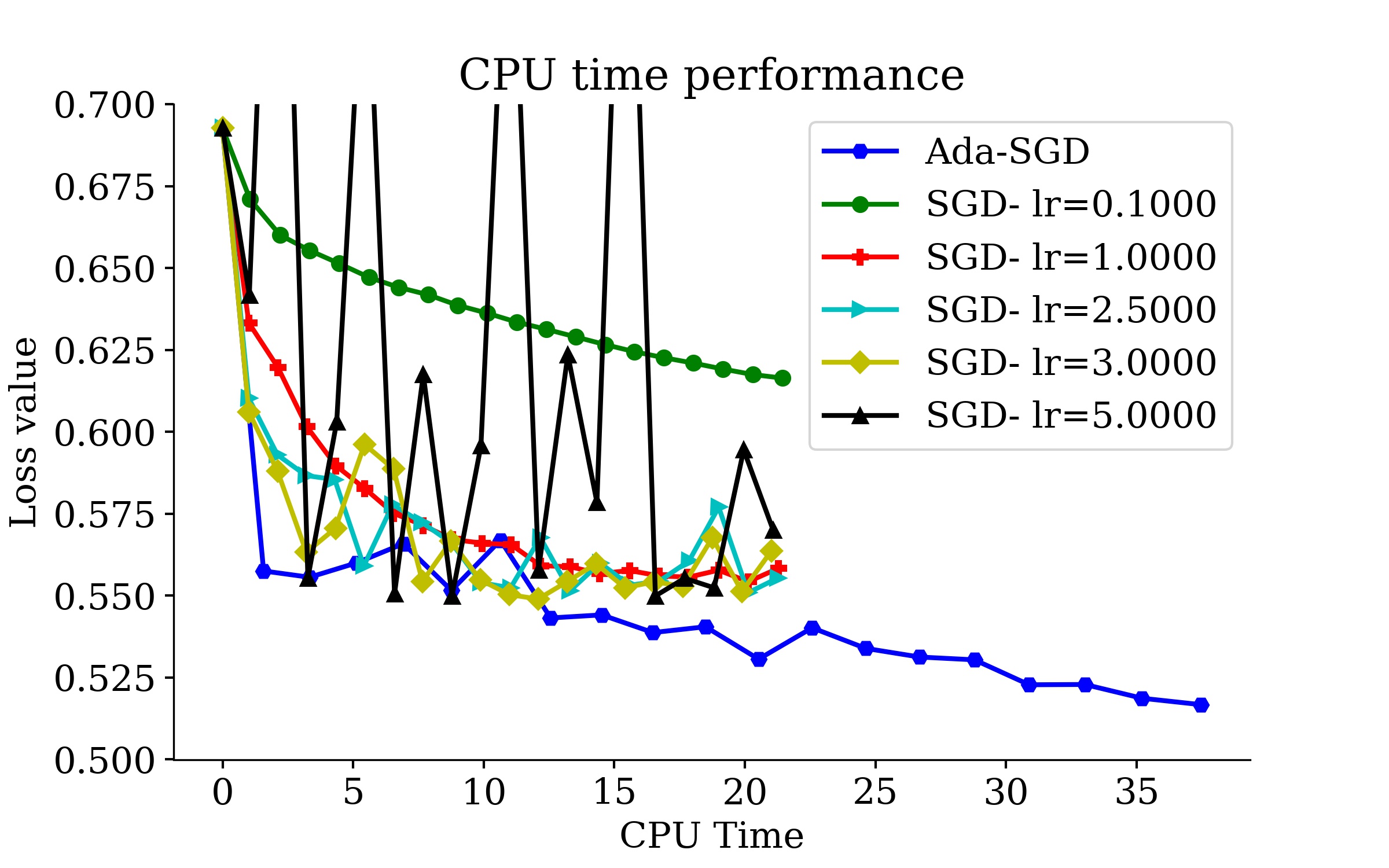} \\  \includegraphics[width=41mm]{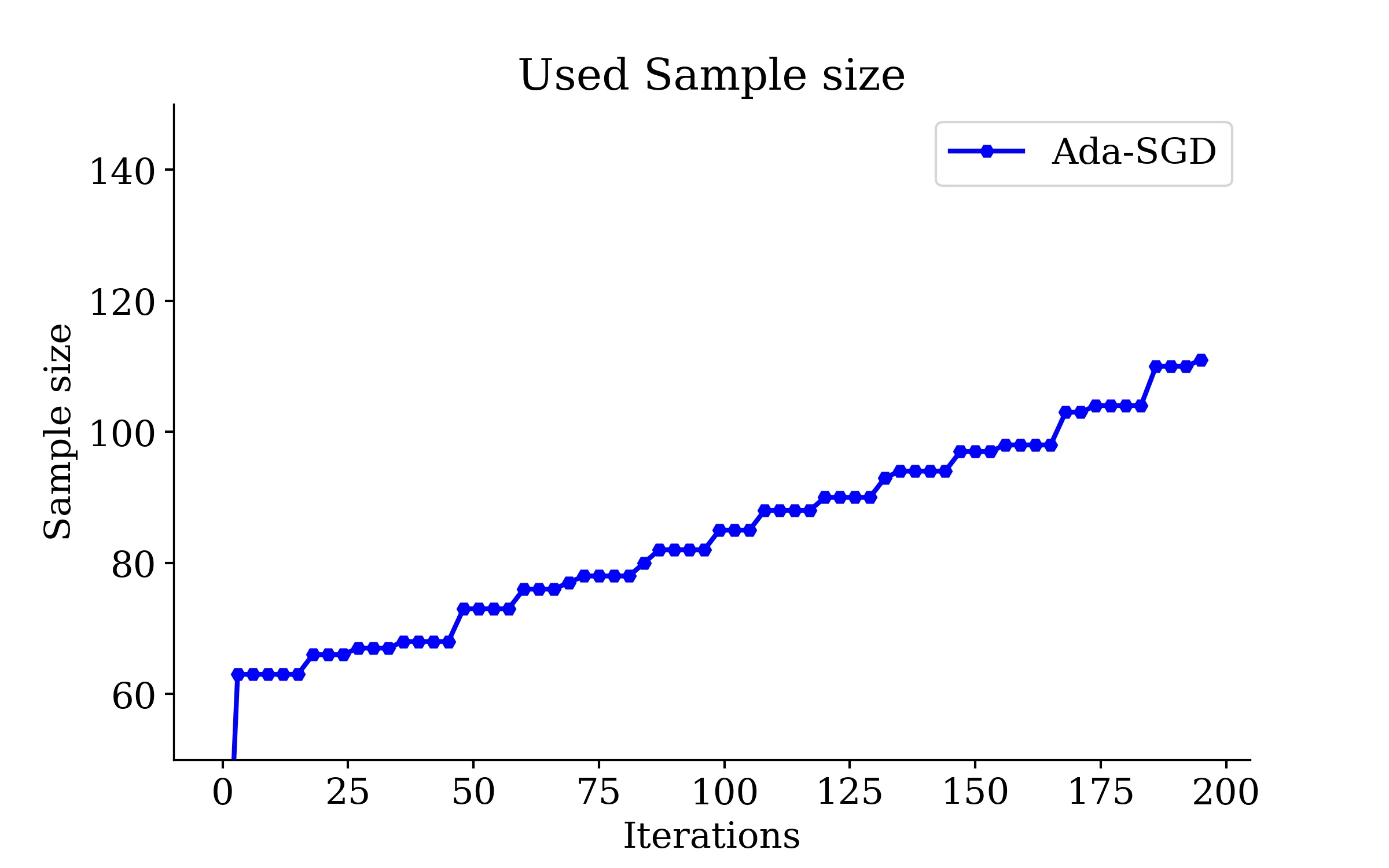} & 
 \includegraphics[width=41mm]{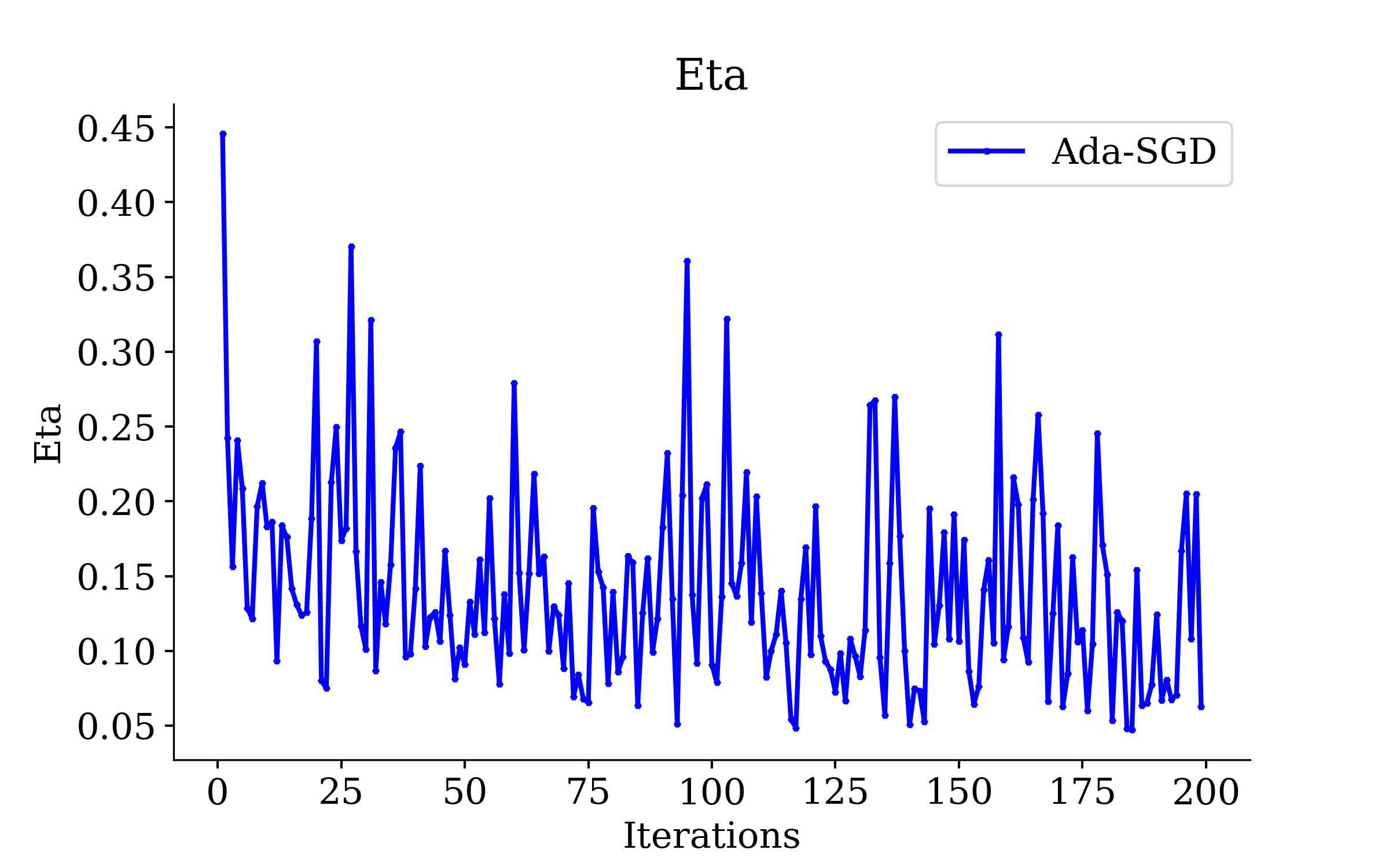}
\end{tabular}
\caption{Numerical results for logistic regression on webspam dataset}
\label{webspam}
\end{figure}

Numerical results on Covertype and Webspam datasets are reported in Figures  \ref{covtype}, \ref{webspam}. For results on other datasets mentioned in Table 1, please refer Appendix J.

We observe that the Ada-SGD successfully determines adequate learning rates and outperforms the vanilla SGD with optimized learning rate, SGD with inner-product test, SGD with augmented inner-product test, SGD with norm test, as shown in Figure \ref{covtype}. We also observe that the batch-size increases gradually and stays relatively small compared to the other methods.

In Figure \ref{webspam}, we observe that the Ada-SGD successfully determines adequate learning rates and outperforms the vanilla SGD with different learning rates. One interesting observation is that the learning rates determined by the Ada-SGD algorithm oscillate around the learning rate with value 2 which turned out to be the optimal learning rate for the vanilla SGD. The Eta sub-figure in Figure \ref{webspam} refers to $\eta_k = \frac{\rho_k}{\delta_k}$, which quantifies the decrease guarantee in our framework and should converge to zero since the objective function is bounded below. 

The above observations about the learning rate computed by 
our adaptive approach on logistic regression motivated us to introduce a hybrid version of our framework, where we run our framework, for a relatively small number of iterations in the beginning of the some milestone epochs, and then use the median of the adaptive learning rates as a constant learning rate until the next milestone (similar  to learning rate schedulers that are widely used in practice).

Another interesting observation is that the Adaptive-SGD sometimes takes large steps but manages to stay stable compared to vanilla SGD with a large step size. This is manly due to both the adaptive step computation based on local curvature and batch-increasing strategy that reduces the stochasticity of the gradient.

\subsection{Deep neural network results}

Although the loss function corresponding to neural network models is generally non-convex, we were motivated to apply our framework to training deep learning models based on the recent experimental observations in \citep{hessiannn}. This work numerically studied the spectral density of the Hessian of deep neural network loss functions throughout the optimization process and observed that the negative eigenvalues tend to disappear rather rapidly in the optimization process. In our case, we are only interested in the curvature along the gradient direction. Surprisingly, we found that negative curvature was rarely encountered at the current iterate along the step direction. Hence, when a negative value was encountered, we simply took the median of the step sizes over the most recent K steps as a heuristic. In our experiments we set K = 20.

\begin{figure}
\centering
\begin{tabular}{cc}
  \includegraphics[width=41mm]{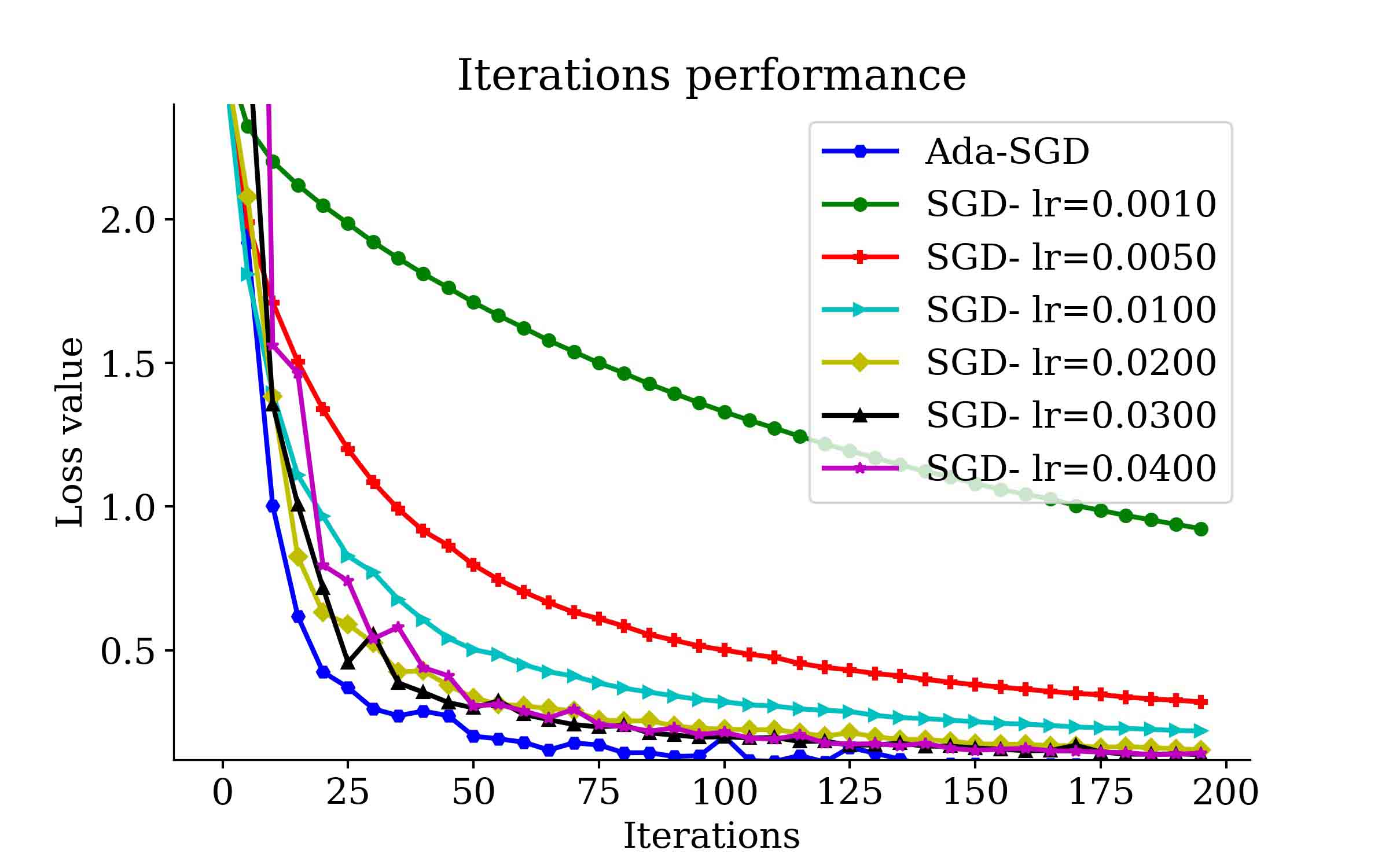} & \includegraphics[width=41mm]{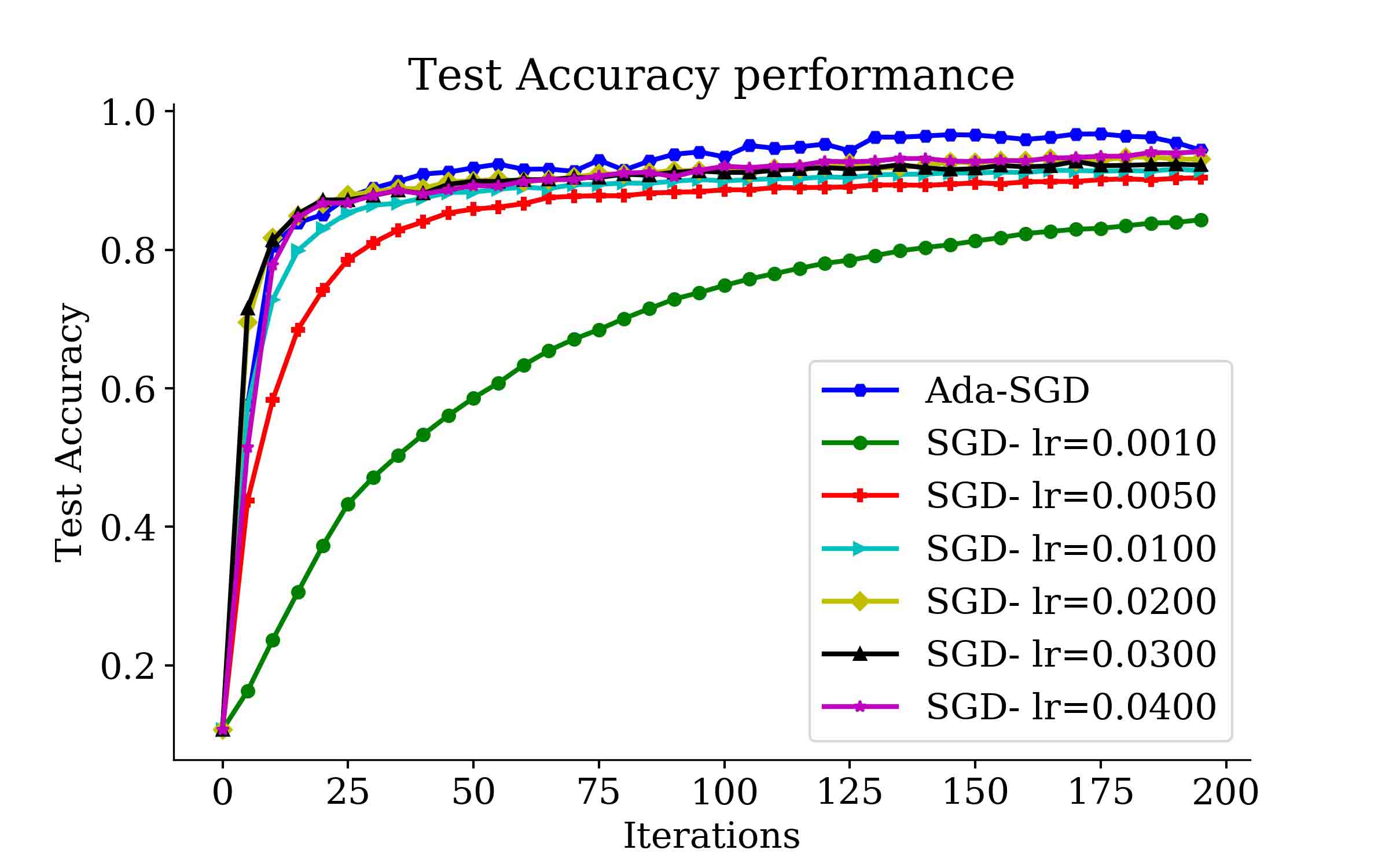} \\ \includegraphics[width=41mm]{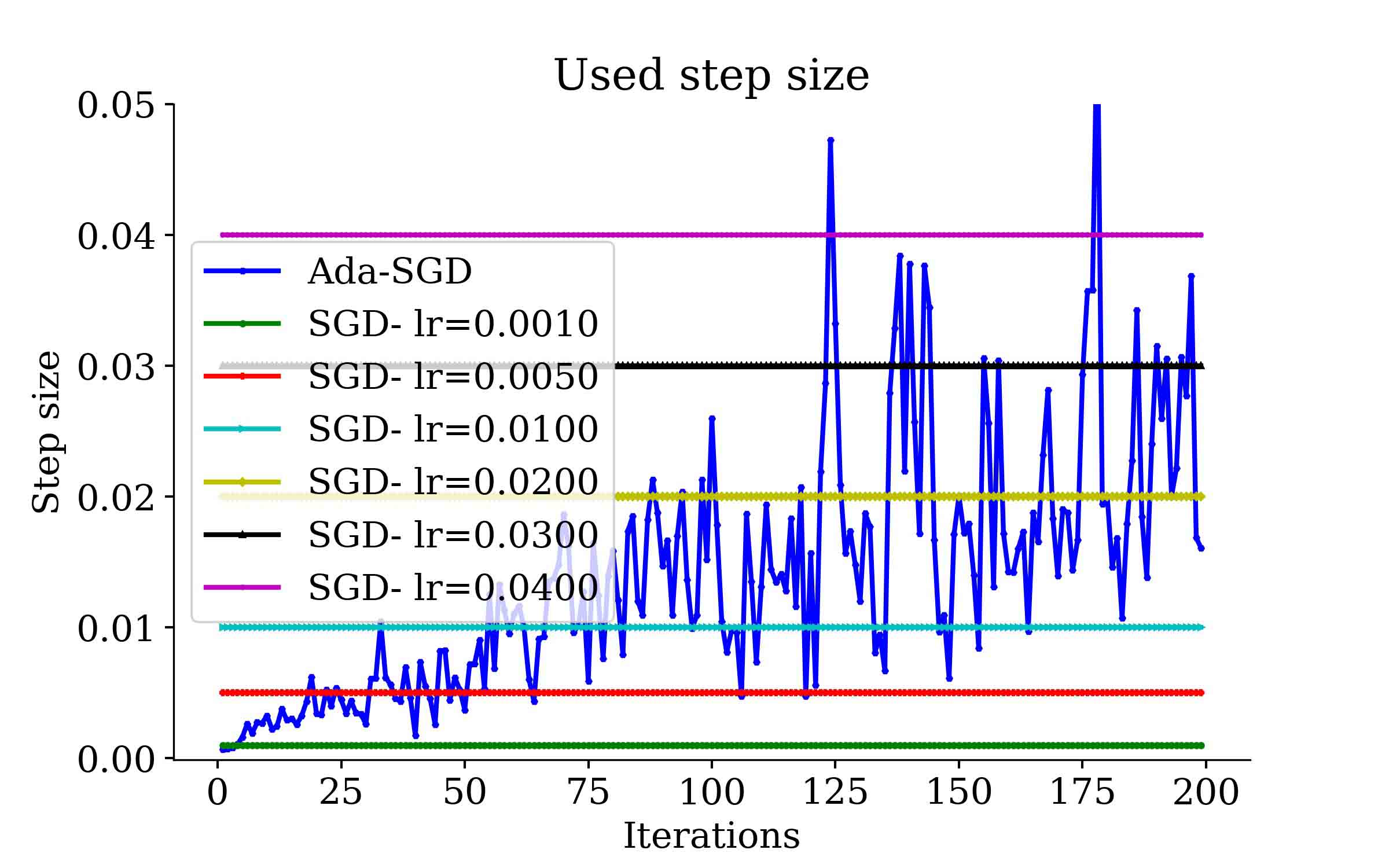} &
 \includegraphics[width=41mm]{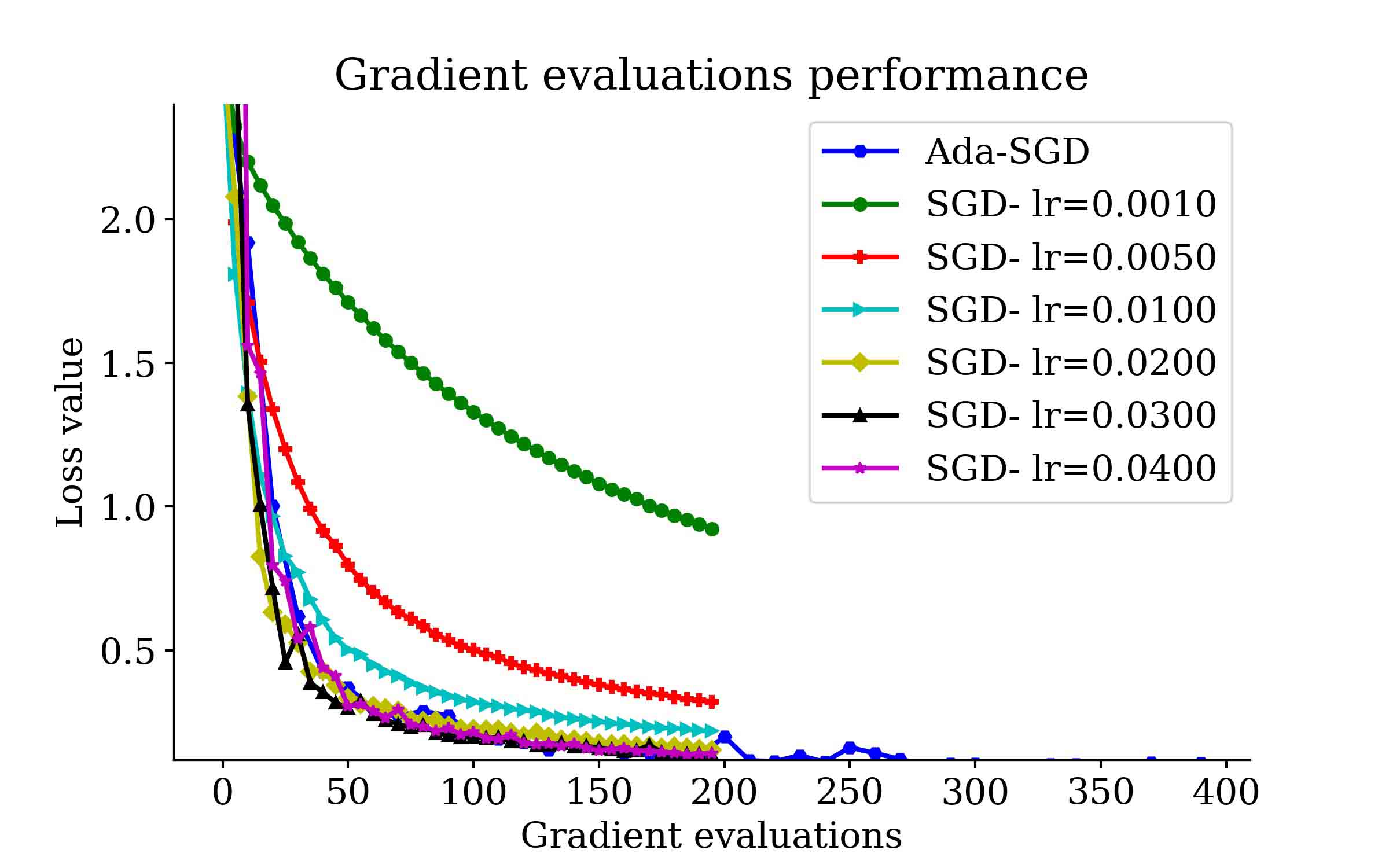} \\  \includegraphics[width=41mm]{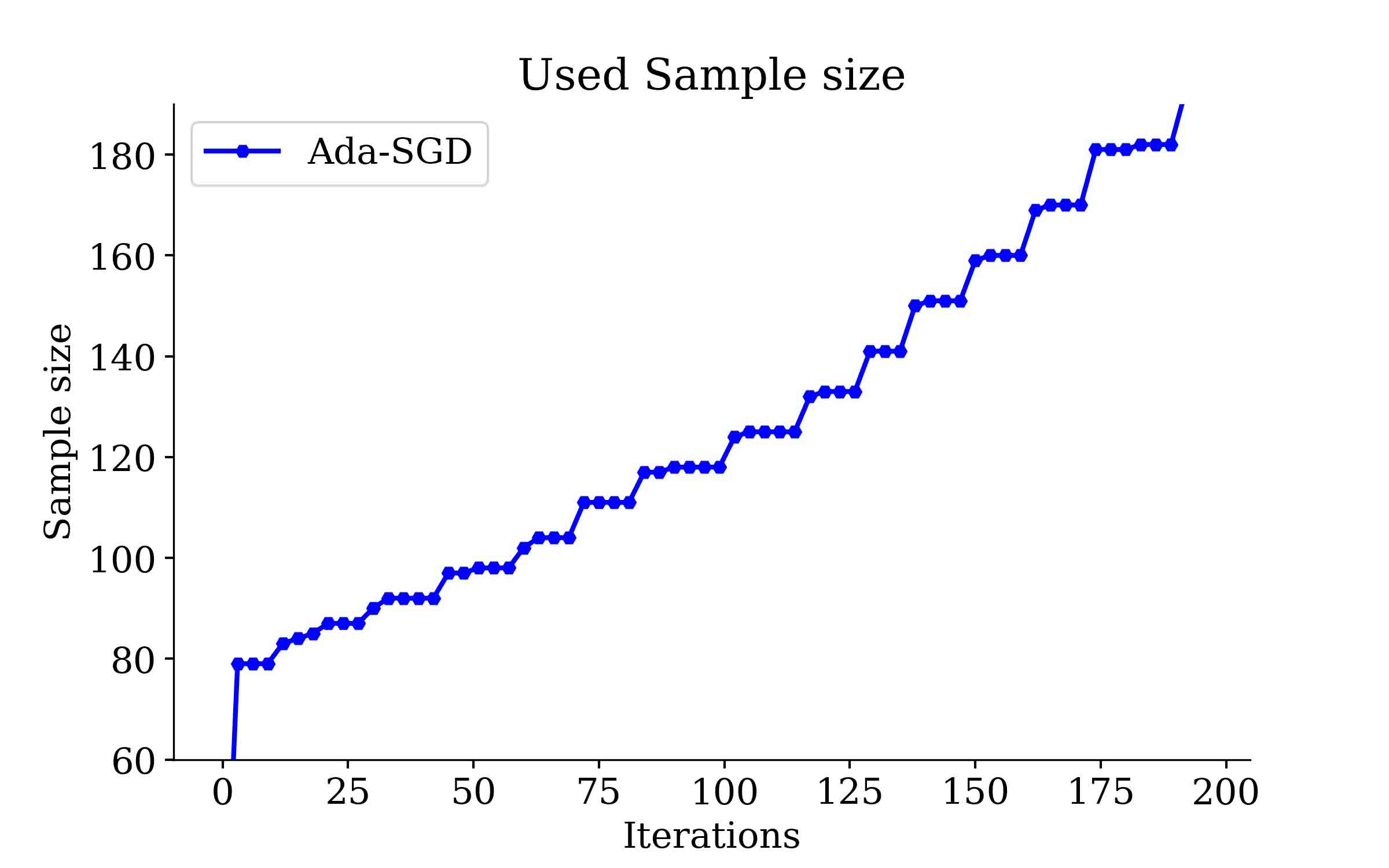} & 
 \includegraphics[width=41mm]{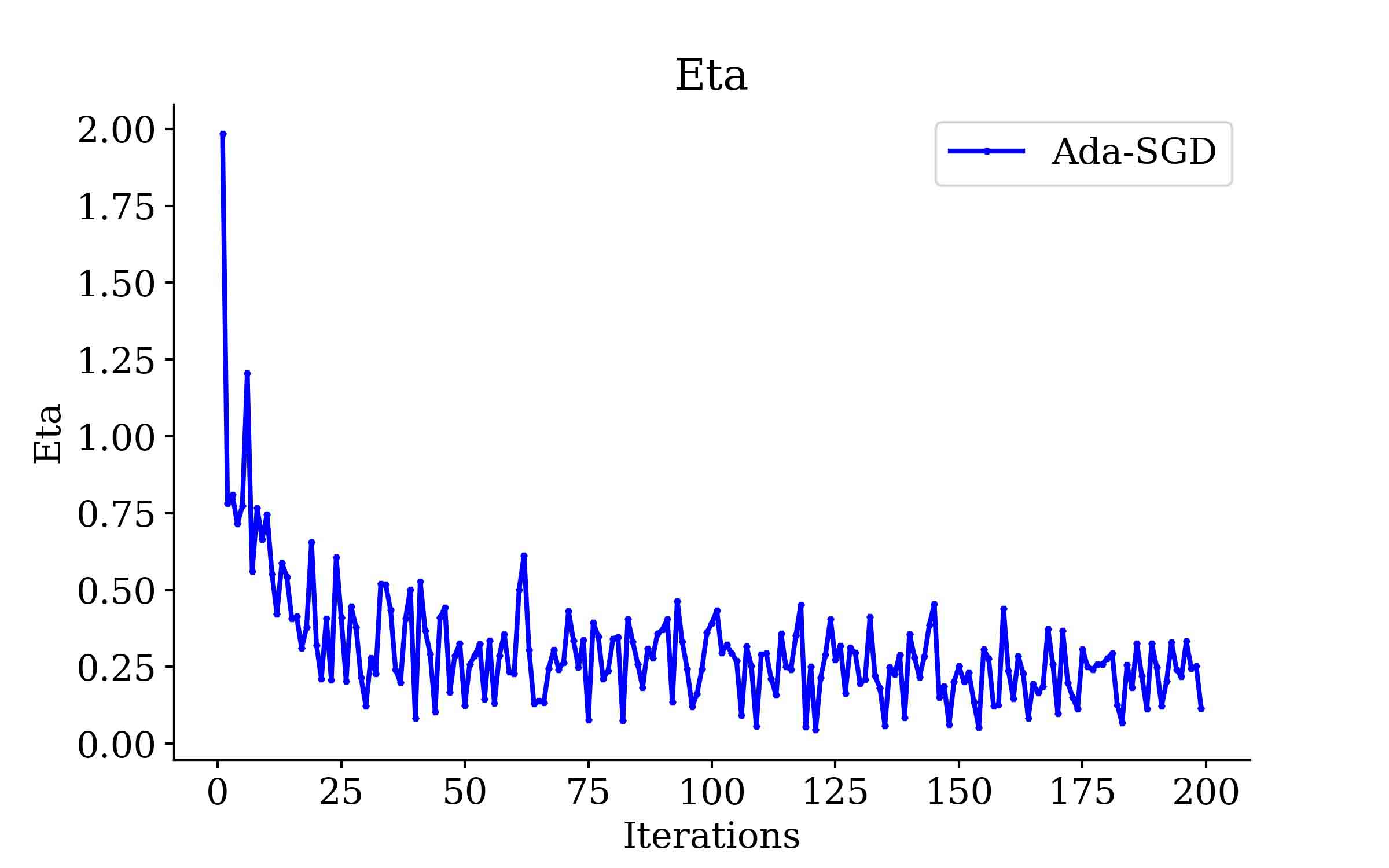}
\end{tabular}
\caption{Numerical results training CNN on MNIST dataset}
\label{mnist}
\end{figure}

\begin{figure}
  \includegraphics[width=90mm]{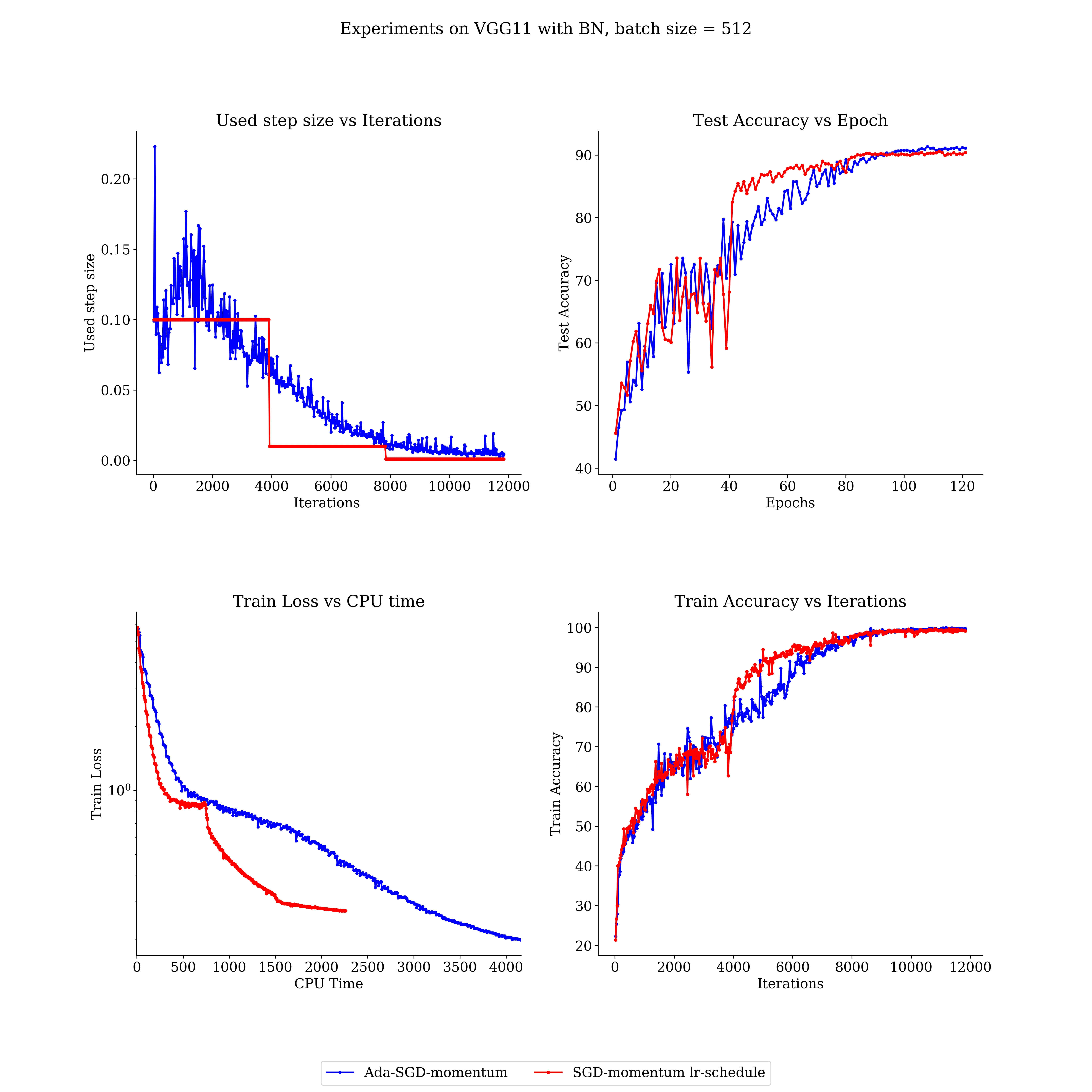} 
\caption{Numerical results on CIFAR10 using VGG11 model}
\label{vggcifar}
\end{figure}

The Adaptive-SGD method successfully determined near-optimal learning rates (see Figure \ref{mnist}) and outperformed vanilla SGD with different learning rates in the range of the chosen adaptive learning rates. The CNN we used was a two-layer convolutional neural network with ReLu activation function and batch-normalization as well as a fully connected layer at the end (Appendix J.4). The Adaptive-SGD with momentum also recovered near-optimal learning rates for VGG11 Figure \ref{vggcifar} and DavidNet architecture: According to DAWNBench\footnote{ \url{https://dawn.cs.stanford.edu/benchmark/CIFAR10/train.html}.}, DavidNet\footnote{ \url{https://myrtle.ai/how-to-train-your-resnet-4-architecture/}.} (a custom 9-layer Residual ConvNet proposed by David
C. Page.) is the fastest model for CIFAR-10 dataset (as of April 1st, 2019). We choose to experiment with this model because the author made a considerable effort to find a near-optimal piece-wise learning-rate schedule and we wanted to see whether our method mimics this schedule. Results are presented in Appendix (J.4.1).

\begin{figure}
  \includegraphics[width=90mm]{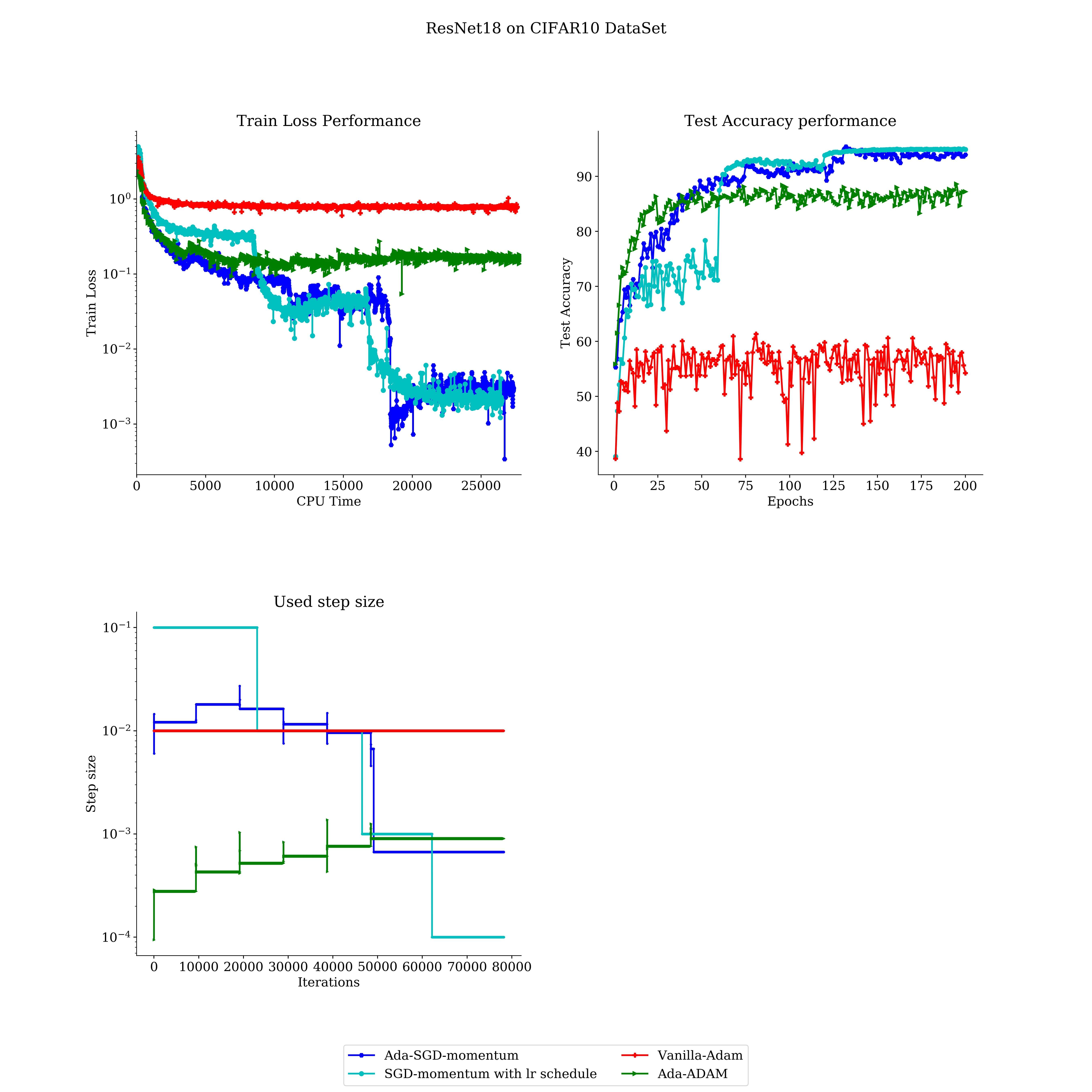} 
\caption{Numerical results on CIFAR10 using ResNet18 model}
\label{resnetcifar}
\end{figure}

However, as the number of parameters grows, the adaptive-step computation cost slows down the training. To fix this issue, we experimented with a large scale model using ResNet18, where we exploited the adaptive framework on SGD with momentum and ADAM to automatically tune the learning rate at the beginning of epochs 1, 25, 50, ... 150 (using first 20 mini-batch iterations to roughly estimate a good constant learning rate between milestones). This method can be seen as a substitute for the common approach of using a learning rate schedule to train large neural nets such where typically the user supplies an initial learning rate, milestone epochs and a decay lr-factor. We tested our framework against hyper-parameters used in \citep{resnetpaper} and our adaptive SGD with momentum achieved roughly the same performance without any hyper-parameters tuning. We believe this tool will be useful for selecting good learning rates. Results are presented in Figure \ref{resnetcifar}.

\section{Conclusion}

We presented an adaptive framework for stochastic optimization that we believe is a valuable tool for fast and practical learning rate tuning, especially in deep learning applications, and which can also be combined with popular variants of SGD such as ADAM, AdaGrad, etc. Studying theoretical convergence guarantees of our method in DNNs which generate non-convex loss functions, and the convergence or our adaptive framework with other variants of SGD suggest interesting avenues for future research.



\clearpage
\bibliography{paper}
\bibliographystyle{icml2020}

\newpage
\onecolumn

\titleformat{\section}{\Large\bfseries}{\thesection}{1em}{}
\titleformat{\subsection}{\large\bfseries}{\thesubsection}{1em}{}
\gdef\thesection{Appendix \Alph{section}}



{\centering{\LARGE\bfseries A Dynamic Sampling Adaptive-SGD Method for Machine Learning}

\vspace{1em}
\centering{{\LARGE\bfseries Supplementary material}}

}

\paragraph{Outline.} The supplementary material of this paper is organized as follows.
\begin{itemize}[leftmargin=*]
    \item \ref{apx:proofs} gives the proofs of the results in the Analysis section of the paper.
    \item \ref{apx:experiments} presents additional  numerical results.
\end{itemize}

\vspace{1em}

\section{Proofs}\label{apx:proofs}

\vskip 1em
\textbf{Notation:}

\begin{itemize}[label=-]
  \item $d_k = - g_k = - \nabla F_{S_{k}}\left(x_{k}\right)$.
  \item $\delta_k=\|d_k\|_{x_k}=\sqrt{g_k^{T} \nabla^2 F\left(x_{k}\right) g_k}$.
  \item $\hat{\delta}_k=\sqrt{g_k^{T} \nabla^2 F_{X_{k}}\left(x_{k}\right) g_k}$.
  \item $\rho_{k}=g_{k}^{T} g = \nabla F_{S_{k}}\left(x_{k}\right)^{T} \nabla F\left(x_{k}\right)$.
  \item $\eta_{k}=\frac{\rho_{k}}{\delta_{k}}$ ; $t_{k}^*=\frac{\rho_{k}}{\left(\rho_{k}+\delta_{k}\right) \delta_{k}}$.
  \item $\hat{\delta}_k=\sqrt{g_k^{T} \nabla^2 F_{X_{k}}\left(x_{k}\right) g_k}$.
  \item $\hat{t}_k = \frac{\rho_{k}}{\left(\rho_{k}+\hat{\delta}_{k}\right) \hat{\delta}_{k}}$.
\end{itemize}

\textbf{Theorem 1} 
    Let $0 <\nu\gamma, \varepsilon, p, \delta < 1$. Suppose that $F$ satisfies the Assumptions 1-4 and that we have an efficient way to compute $G(x)$ where \(\left\|\nabla F_{i}(\mathrm{x})\right\| \leq G(\mathrm{x}), \quad \forall i=1, \ldots, n\). Let $\{x_k, X_k, S_k\}$ be the set of iterates, and sample Hessian and gradient batches generated by taking the step $t_{k, \varepsilon}$ at iteration $k$ starting from any $x_0$, where $|S_{k}|, |X_{k}|$ are chosen such that Lemma \ref{gkprob} and Lemma \ref{dhdprob} are satisfied at each iteration. Then, for any $k$, with probability $(1-p)(1-\delta)$ we have :
	\begin{equation*} 
	    (F(x_{k+1}) - F(x^*)) \leq \rho (F(x_k) - F(x^*)),
	\end{equation*}
     where 
	\begin{equation*} 
	    \rho=  1 - \frac{m(1-\nu)^2(1-\varepsilon)}{2(1+\nu^2) M (1+\frac{\gamma}{\sqrt{m}})}.
	\end{equation*}

\begin{proof}
By Assumption 3, choosing $\nu \leq \frac{1}{\gamma}$ insures that the conditions for Lemma \ref{gkprob} are satisfied. Therefore, we have with probability $1-\delta$:

\begin{align*} 
\text{Lemma } \ref{gkprob}  &\implies \left\|g_{k}\right\|^{2}-g_{k}^{T} g=g_{k}^{T}\left(g_{k}-g\right) \leq\left\|g_{k}\right\|\left\|g-g_{k}\right\| \leq \nu\left\|g_{k}\right\|^{2}\\
&\implies g_{k}^{T} g \geq(1-\nu)\left\|g_{k}\right\|^{2}.
\end{align*}
Also:
\begin{align*} 
\eqref{gkprob} \implies \|g\|^{2}=\left\|g-g_{k}+g_{k}\right\|^{2} \leq 2\left\|g-g_{k}\right\|^{2}+2\left\|g_{k}\right\|^{2} \leq 2\left(1+\nu^{2}\right)\left\|g_{k}\right\|^{2}.
\end{align*}
Hence, with probability $1-\delta$:
\begin{align} 
\frac{\left(g_{k}^{T} g\right)^{2}}{\left\|g_{k}\right\|^{2}} \geq(1-\nu)^{2}\left\|g_{k}\right\|^{2} \geq \frac{(1-\nu)^{2}}{2\left(1+\nu^{2}\right)}\|g\|^{2} \label{lemma1part1}.
\end{align}
Let $\Delta (\delta_k, t) = (\delta_k + \rho_k)t + log(1-\delta_k t)$, using Lemma \eqref{selfconc}, we have:
\begin{equation}
\quad F(x_k+t d_k) \leq F(x_k)-\Delta (\delta_k, t).
\end{equation}
If we sample the Hessian at the rate described in Lemma \ref{dhdprob}, the following inequality is satisfied with probability $(1-p)$:
$$
 \inf_{d} \frac{|d^T(\nabla^{2} F_S(x)-\nabla^{2} F(\mathbf{x})) d|}{\|d\|^2} = \left\| \nabla^{2} F_S(x)-\nabla^{2} F(\mathbf{x})\right\| \leq \varepsilon m \leq \varepsilon \inf_{d} \frac{d^T \nabla^{2} F(\mathbf{x}) d}{\|d\|^2}.
$$
Taking $d = g_k$ results into the following inequalities being satisfied with  probability $(1-p)$:
\begin{equation}
    \frac{\hat{\delta_k}}{\sqrt{1+\varepsilon}} \leq \delta_k \leq \frac{\hat{\delta_k}}{\sqrt{1-\varepsilon}} = \hat{\delta}_{k, \varepsilon}.
\end{equation}
We have, for $t<\frac{1}{\delta}$:
$$
\frac{\partial \Delta}{\partial \delta}(\delta, t) = t-\frac{t}{1-\delta t} = \frac{\delta t^2}{1-\delta t} < 0.
$$
Using the monotonicity of $\Delta (., t)$, (14) and the choice of step size as $t_{k, \varepsilon} = \frac{\rho_{k}}{\left(\rho_{k}+\hat{\delta}_{k, \varepsilon}\right) \hat{\delta}_{k, \varepsilon}}$, we have:
\begin{equation}
\quad F(x_k+t_{k, \varepsilon} d_k) \leq F(x_k)-\Delta(\hat{\delta}_{k, \varepsilon}, t_{k, \varepsilon}).
\end{equation}
The choice of $t_{k, \varepsilon}$ is possible because we now prove that $t_{k, \varepsilon} < \frac{1}{\delta_k}$. Let $t(\delta) = \frac{\rho_k}{\delta (\delta+\rho_k)}$ clearly $t(.)$ is decreasing as a function of $\delta$. Using the inequality (14) we have
$$
t_{k, \varepsilon} = t(\hat{\delta}_{k, \varepsilon}) \leq t(\delta_k) = \frac{1}{\delta_k}-\frac{1}{\delta_k + \rho_k} \leq \frac{1}{\delta_k}.
$$

Now we obtain a decrease guarantee of the function with probability $(1-p)$
\begin{equation*}:
\quad F(x_k+t_{k, \varepsilon} d_k) \leq F(x_k)-\Delta(\hat{\delta}_{k, \varepsilon}, t_{k, \varepsilon}) = F(x_k)- \omega (\eta_{k, \varepsilon} ).
\end{equation*}

Now, using the Cauchy-Schwarz inequality, inequalities (14) and Assumption 1-3 we have:
\begin{equation*}
|\eta_{k, \varepsilon}| = |\frac{\rho_k}{\frac{\hat{\delta}_{k}}{\sqrt{1-\varepsilon}}}| \leq \frac{\|g\| \|g_k\|}{\delta_k} \leq \frac{\|g\| \|g_k\|}{\sqrt{g_k^T \nabla^2 F(x_k) g_k}} \leq \frac{\|g\|}{\sqrt{m}} \leq \frac{\gamma}{\sqrt{m}}.
\end{equation*}
By observing that $\omega(z) = z - log(1+z)$ satisfies $\omega(z) \geq \frac{1}{2}(1+\Gamma)^{-1} z^{2}$ for all $z \in[0, \Gamma]$, with probability $(1-p)$, we have:
 \begin{equation*}
     F(x_k+t_{k, \varepsilon}d_k)  \leq F(x_k) -  \frac{1-\varepsilon}{2(1+\frac{\gamma}{\sqrt{m}})} \hat{\eta}_{k}^2,
\end{equation*} 
with $\hat{\eta}_{k}=\frac{g_{k}^{T} g}{\hat{\delta}_{k}}$. Using Assumption 2, we have $\hat{\delta}_{k}^2 \leq M \|g_k\| ^2$. Therefore, we have:
 \begin{equation*}
     F(x_k+t_{k, \varepsilon} d_k)  \leq F(x_k) -  \frac{1-\varepsilon}{2 M (1+\frac{\gamma}{\sqrt{m}})} \frac{(g^T g_k)^2}{\|g_k\|^2}.
\end{equation*} 
Using \eqref{lemma1part1}, we have with probability $(1-p)(1-\delta)$:
 \begin{equation}
     F(x_k+t_{k, \varepsilon} d_k)  \leq F(x_k) -  \frac{(1-\nu)^2(1-\varepsilon)}{4(1+\nu^2) M (1+\frac{\gamma}{\sqrt{m}})} \|\nabla F(x_k)\|^2 \label{decreaselemma}.
\end{equation}
It is well known \citep{bertsekas2003convex} that for strongly convex functions:   
\[
\|\nabla F(x_k)\|^2 \geq 2m [F(x_k) - F(x^*)].
\]
 Substituting this in \eqref{decreaselemma} and subtracting $F(x^*)$ from both sides, we obtain with probability $(1-p)(1-\delta)$:
\begin{align*}
F(x_{k+1}) - F(x^*) &\leq  \rho (F(x_k) - F(x^*)), \nonumber
\end{align*}
with :
$$
\rho=  1 - \frac{m(1-\nu)^2(1-\varepsilon)}{2(1+\nu^2) M (1+\frac{\gamma}{\sqrt{m}})},
$$
which is the desired result.
\end{proof}

\textbf{Lemma 4.} Suppose that $F$ satisfies Assumptions 1-4,
	Let $\{x_k\}$ be the iterates generated by the idealized Algorithm~\ref{alg:adasgd} where $|S_{k}|$ is chosen such that the (exact) acute-angle test~\eqref{eq:exactest} and Hessian sub-sampling test ~\eqref{dhdtest} are satisfied at each iteration for any given constants $0 <\nu, \varepsilon, p < 1$. Then, for all $k$, starting from any $x_0$,  
	\begin{align*}\label{c-sample-norm}
	    \frac{(\nabla F_{S_k}(x_k)^{T} \nabla F(x_k))^2}{ \norm{\nabla F_{S_k}(x_k)}^2}  \geq (1-\nu^2) \norm{\nabla F(x_k)}^2.
	\end{align*}
	With probability at least $1-p$. Moreover, with probability $(1-p)^2$, we have
	\begin{equation*}
        F(x_{k+1}) \leq F(x_{k}) - \frac{\alpha}{2} \|\nabla F(x_k)\|^2.
    \end{equation*}  
	where
	\begin{align*} 
	    \alpha = \frac{(1-\nu^2)(1-\varepsilon)}{M (1+\frac{\gamma}{\sqrt{m}})} > 0.
	\end{align*}  

\begin{proof}
Since the (exact) acute-angle test~\eqref{eq:exactest} is satisfied, using Markov's inequality on the positive random variable $U_k := \| \frac{\nabla F_{S_{k}}\left(x_{k}\right)}{\|\nabla F_{S_{k}}\left(x_{k}\right)\|} - \frac{\nabla F_{S_{k}}\left(x_{k}\right)^{T} \nabla F\left(x_{k}\right)}{\|\nabla F_{S_{k}}\left(x_{k}\right)\|\left\|\nabla F\left(x_{k}\right)\right\|^{2}} \nabla F\left(x_{k}\right)\|^{2}$, we have $\eqref{eq:exactest} \implies \mathbb{E}(U_k) \leq p \nu^2$ and:
\begin{align*}
    \mathbb{P}(U_k \leq \nu ^2) = 1-\mathbb{P}(U_k > \nu ^2) \geq 1 - \frac{\mathbb{E}(U_k)}{\nu ^2} \geq 1- \frac{p \nu ^2}{\nu ^2} = 1-p.
\end{align*}
Therefore, the following inequality is satisfied with probability at least $1-p$:
\begin{align}
    \| \frac{\nabla F_{S_{k}}\left(x_{k}\right)}{\|\nabla F_{S_{k}}\left(x_{k}\right)\|} - \frac{\nabla F_{S_{k}}\left(x_{k}\right)^{T} \nabla F\left(x_{k}\right)}{\|\nabla F_{S_{k}}\left(x_{k}\right)\|\left\|\nabla F\left(x_{k}\right)\right\|^{2}} \nabla F\left(x_{k}\right)\|^{2} \leq \nu^2 \label{theorem2g}
\end{align}
\begin{align*} 
(\eqref{theorem2g})  &\implies 1 - 2\frac{(\nabla F_{S_k}(x_k)^{T} \nabla F(x_k))^2}{ \norm{\nabla F_{S_k}(x_k)}^2 \norm{\nabla F(x_k)}^2} + \frac{(\nabla F_{S_k}(x_k)^{T} \nabla F(x_k))^2}{ \norm{\nabla F_{S_k}(x_k)}^2 \norm{\nabla F(x_k)}^2}  \leq \nu^2 \\
&\implies 1 - \frac{(\nabla F_{S_k}(x_k)^{T} \nabla F(x_k))^2}{ \norm{\nabla F_{S_k}(x_k)}^2 \norm{\nabla F(x_k)}^2}  \leq \nu^2 \\
&\implies \frac{(\nabla F_{S_k}(x_k)^{T} \nabla F(x_k))^2}{ \norm{\nabla F_{S_k}(x_k)}^2} \geq (1-\nu^2) \norm{\nabla F(x_k)}^2.
\end{align*}
Therefore with probability at least $1-p$:
\begin{equation}
    \frac{\left(g_{k}^{T} g\right)^{2}}{\left\|g_{k}\right\|^{2}} \geq(1-\nu^{2})\|g\|^{2}. \label{lemma4part1}
\end{equation}

Since the Hessian sub-sampling test \eqref{dhdtest} is satisfied at each iteration, using Markov's inequality on the positive random variable $V_k := (\hat{\delta_k}^2 - \delta_k^2)^2$, we have $\eqref{dhdtest} \implies \mathbb{E}(V_k) \leq p\varepsilon^2  \delta_k^4$ and:
\begin{align*}
    \mathbb{P}(V_k \leq \varepsilon^2  \delta_k^4) = 1-\mathbb{P}(V_k > \varepsilon^2  \delta_k^4) \geq 1 - \frac{\mathbb{E}(V_k)}{\varepsilon^2  \delta_k^4} \geq 1- \frac{p \varepsilon^2  \delta_k^4}{\varepsilon^2  \delta_k^4} = 1-p.
\end{align*}
Therefore, the following inequality is satisfied with probability at least $1-p$:
\begin{equation}
    \frac{\hat{\delta_k}}{\sqrt{1+\varepsilon}} \leq \delta_k \leq \frac{\hat{\delta_k}}{\sqrt{1-\varepsilon}} = \hat{\delta}_{k, \varepsilon}.
\end{equation}

Following similar arguments to those used in the proof of Theorem \ref{theorem1}, we have with probability $(1-p)^2$:
\begin{equation}
    F(x_{k+1}) \leq F(x_{k}) - \frac{\alpha}{2} \|\nabla F(x_k)\|^2,\label{decreaselemma4}
\end{equation}  
where
\begin{align*} 
    \alpha = \frac{(1-\nu^2)(1-\varepsilon)}{M (1+\Gamma)} > 0.
\end{align*}
Which is the desired result.
\end{proof}

\textbf{Theorem 2.} Suppose that $F$ satisfies Assumptions 1-4,
	Let $\{x_k\}$ be the iterates generated by the idealized Algorithm~\ref{alg:adasgd} where $|S_{k}|$ is chosen such that the (exact) acute-angle test~\eqref{eq:exactest} and Hessian sub-sampling test ~\eqref{dhdtest} are satisfied at each iteration for any given constants $0 <\nu, \varepsilon, p < 1$. Then, for all $k$, starting from any $x_0$, with probability of at least $(1-p)^{2k}$, we have
	\begin{equation*}
	    F(x_k) - F(x^*) \leq \rho^k (F(x_0) - F(x^*)),
	\end{equation*}
     where 
	\begin{equation*} 
	    \rho= 1 -m \alpha =  1 - \frac{m(1-\nu^2)(1-\varepsilon)}{M (1+\frac{\gamma}{\sqrt{m}})}.
	\end{equation*}
	
\begin{proof}
It is well known \citep{bertsekas2003convex} that for strongly convex functions   
\[
\|\nabla F(x_k)\|^2 \geq 2m [F(x_k) - F(x^*)].
\]
 Substituting this into \eqref{decreaselemma4} and subtracting $F(x^*)$ from both sides, we obtain with probability $(1-p)^2$:
\begin{align*}
F(x_{k+1}) - F(x^*) &\leq  \rho (F(x_k) - F(x^*)), \nonumber 
\end{align*}
with :
\begin{equation*} 
    \rho= 1 -m \alpha =  1 - \frac{m(1-\nu^2)(1-\varepsilon)}{M (1+\frac{\gamma}{\sqrt{m}})},
\end{equation*}
from which the theorem follows by induction on $k$.
\end{proof}

\textbf{Decreasing p:} if we decrease $p = p_k$ to zero such that $\sum_{k = 1}^{\infty} p_k < \infty$, then the cumulative  probability in \ref{theorem2} will be:
$$
\prod_{k = 1}^{N} (1-p_k)^{2} = e^{ 2 \sum_{k = 1}^{N} log(1-p_k)}.
$$
As $p_k \xrightarrow{\infty} 0$, we have $-log(1-p_k) \stackrel{\infty}{\sim} p_k$ and $\sum_{k = 1}^{\infty} p_k < \infty$. Hence, by positiveness, $-\sum_{k = 1}^{N} log(1-p_k) < \infty$. Therefore, the cummulative probability $\prod_{k = 1}^{N} (1-p_k)^{2}$ converges to a non-zero probability. For example, for $p_k = \frac{1}{(k+1)^2}$, we get 
$$
\prod_{k = 1}^{N-1} (1-p_k)^{2} = \left(\prod_{k=2}^{N}\left(1-\frac{1}{k^2}\right)\right)^2 = \left(\prod_{k=2}^{N}\frac{k-1}{k}\prod_{k=2}^{N}\frac{k+1}{k}\right)^2=\left(\frac{N+1}{2N}\right)^2 \xrightarrow{\infty} 0.25.
$$

\section{Scale invariant versions and Numerical Results}\label{apx:experiments}

\subsection{Scale-invariant version: Ada-ADAM}\label{apx:adaadamapx}

The ADAM optimizer maintains moving averages of stochastic gradients and their element-wise squares:\\
\begin{align*} 
\tilde{m}_{k} &=\beta_{1} \tilde{m}_{k-1}+\left(1-\beta_{1}\right) g_{k}, \quad m_{k}=\frac{\tilde{m}_{k}}{1-\beta_{1}^{k+1}} \\ \tilde{v}_{k} &=\beta_{2}, \tilde{v}_{k-1}+\left(1-\beta_{2}\right) g_{k}^{2}, \quad v_{k}=\frac{\tilde{v}_{k}}{1-\beta_{2}^{k+1}},
\end{align*}

with $\beta_{1}, \beta_{2} \in(0,1)$, and updates $$x_{k+1}=x_{k}-\alpha \frac{m_{k}}{\sqrt{v_{k}}+\varepsilon'} = x_{k}+\alpha d_k.$$

Since our framework can be applied to any direction $d_k$, we propose the following Adaptive learning rate version of ADAM :
\begin{algorithm}[h]
    \KwIn{Initial iterate $x_0$, initial gradient sample  $S_0$, and constants $1 >p, \varepsilon, \nu> 0$, max number of iterations $N$, initial $\tilde{m}_0, \tilde{v}_0 = 0$}
    $k \leftarrow 0$ \\
    \While{$k < N$}{
        Choose samples $S_k$  using Algorithm~\ref{alg:sampleupdate} \\
        $\tilde{m}_{k} =\beta_{1} \tilde{m}_{k-1}+\left(1-\beta_{1}\right) g_{k},$ \hskip0.5em $m_{k}=\frac{\tilde{m}_{k}}{1-\beta_{1}^{k+1}}$\\
        $\tilde{v}_{k} =\beta_{2} \tilde{v}_{k-1}+\left(1-\beta_{2}\right) g_{k}^{2},$ \hskip0.5em    $v_{k}=\frac{\tilde{v}_{k}}{1-\beta_{2}^{k+1}}$\\
        $\hat{d}_{k}=-\frac{m_{k}}{\sqrt{v_{k}}+\varepsilon'}$\\
		Compute $\hat{\rho}_k = -\hat{d}_{k}^{T} \nabla F_{S_{k}}\left(x_{k}\right)$ \\
		Compute $\hat{\delta}_k =\sqrt{\hat{d}_{k}^{T} \nabla^2 F_X (x_k) \hat{d}_{k}}$ \\
		Compute $t_{k, \varepsilon} = \frac{\hat{\rho}_k}{\left(\hat{\rho}_k+\hat{\delta}_{k, \varepsilon}\right) \hat{\delta}_{k, \varepsilon}} $ \\
		Compute new iterate: $x_{k+1} = x_k + t_{k, \varepsilon} \hat{d}_k$ \\
		Set $k \leftarrow k + 1$ 
    }
    \caption{Ada-ADAM: Adaptive sampling and stepsize ADAM}\label{alg:adaadam}
\end{algorithm}

\subsection{Adaptive-SGD with momentum}\label{apx:adamom}

SGD with momentum optimizer maintains moving average of stochastic gradients:\\
\begin{equation*} 
v_{k} =\beta_{1} v_{k-1}+ g_{k}
\end{equation*}

with $\beta_{1} \in(0,1)$, and updates $$x_{k+1}=x_{k}-\alpha v_{k} = x_{k}-\alpha d_k.$$

Since our framework can be applied to any direction $d_k$, we propose the following Adaptive learning rate version of SGD with momentum :
\begin{algorithm}[H]
    \KwIn{Initial iterate $x_0$, initial gradient sample  $S_0$, and constants $1 >p, \varepsilon, \nu> 0$, max number of iterations $N$, initial $v_0 = 0$}
    $k \leftarrow 0$ \\
    \While{$k < N$}{
        Choose samples $S_k$  using Algorithm~\ref{alg:sampleupdate} \\
        $v_{k} =\beta_{1} v_{k-1}+ g_{k}$ \\
        $d_k = -v_{k}$ \\
		Compute $\hat{\rho}_k = -\hat{d}_{k}^{T} \nabla F_{S_{k}}\left(x_{k}\right)$ \\
		Compute $\hat{\delta}_k =\sqrt{\hat{d}_{k}^{T} \nabla^2 F_X (x_k) \hat{d}_{k}}$ \\
		Compute $t_{k, \varepsilon} = \frac{\hat{\rho}_k}{\left(\hat{\rho}_k+\hat{\delta}_{k, \varepsilon}\right) \hat{\delta}_{k, \varepsilon}} $ \\
		Compute new iterate: $x_{k+1} = x_k + t_{k, \varepsilon} \hat{d}_k$ \\
		Set $k \leftarrow k + 1$ 
    }
    \caption{Ada-SGD with momentum}\label{alg:adamom}
\end{algorithm}

\subsection{Binary classification logistic regression}

\subsubsection{Datasets}

For binary classification using logistic regression, we chose 6 data sets from LIBSVM \citep{libsvm} with a variety of dimensions and sample sizes, which are listed in Table \eqref{tab11}.
\begin{table}[h]
	\centering
	\label{datasets}
	\begin{tabular}{|c||c|c|c|}
		\hline
		Data Set  & Data Points $N$ & Dimension $n$  \\ \hline
		covertype   & 581012		   & 54	  \\
		webspam   & 350000          & 254 \\
		rcv1-train.binary & 47237 & 20242 \\
		real-sim & 20959 & 72309  \\
		a1a     & 30,956          & 123     \\    
		ionosphere     & 351         & 34    \\\hline
	\end{tabular}
	\caption{Characteristics of the binary datasets used in the experiments.}
	\label{tab11}
\end{table}

\subsubsection{Hyper-parameters and Training for Logistic regression}

 we performed grid search across the following hyper-parameters and values in the logistic regression setting:

\begin{center}
 \begin{tabular}{||c | c||} 
 \hline
 Hyperparameter & Values  \\ [0.5ex] 
 \hline\hline
 probability of test failure $p$ & \{0.05, 0.1, 0.15, 0.2, 0.5\}  \\ 
 \hline
 level of angle acuteness $\nu$ & \{0.05, 0.1, 0.15, 0.2, 0.5\}  \\
 \hline
 initial batch size & \{16\} \\
 \hline
 level of curvature precision $\epsilon$ & \{0.001, 0.005, 0.01, 0.02, 0.05, 0.1\} \\
 \hline
\end{tabular}
\end{center}

We also used a similar hyperparameter sweep for the CNN task on MNIST.

\subsubsection{Results:}

\begin{figure}
\begin{tabular}{ccc}
  \includegraphics[width=53mm]{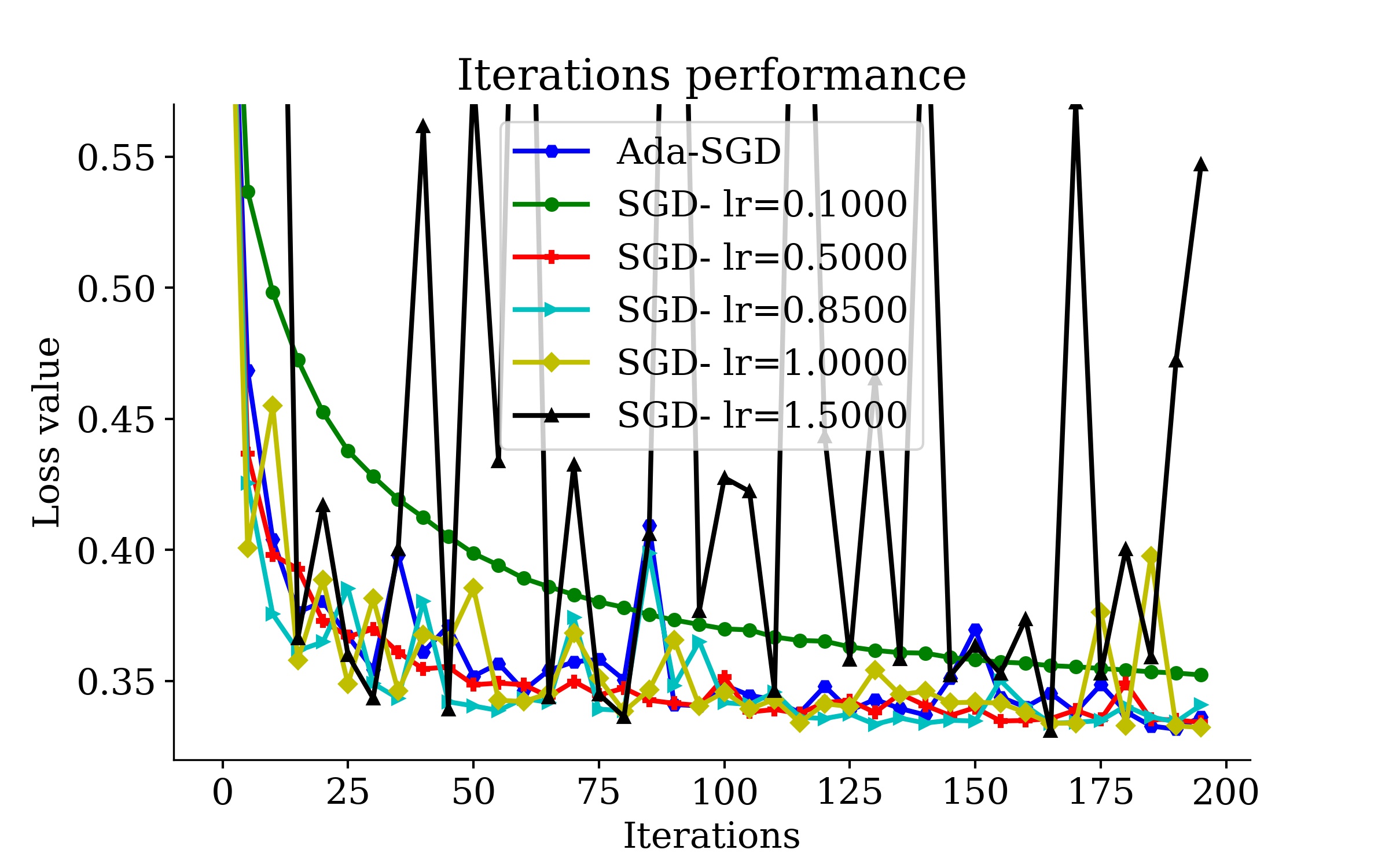} & \includegraphics[width=53mm]{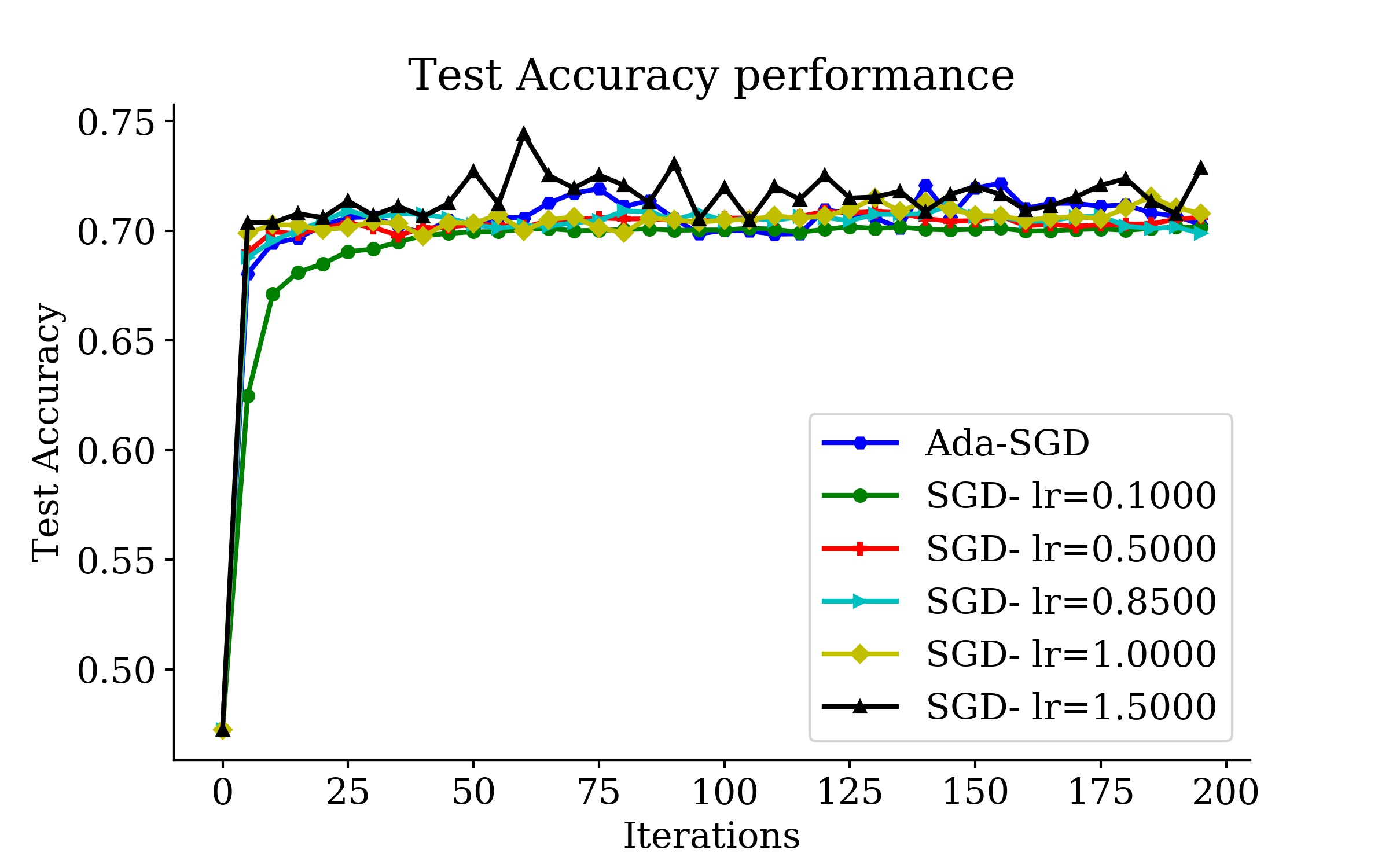} & \includegraphics[width=53mm]{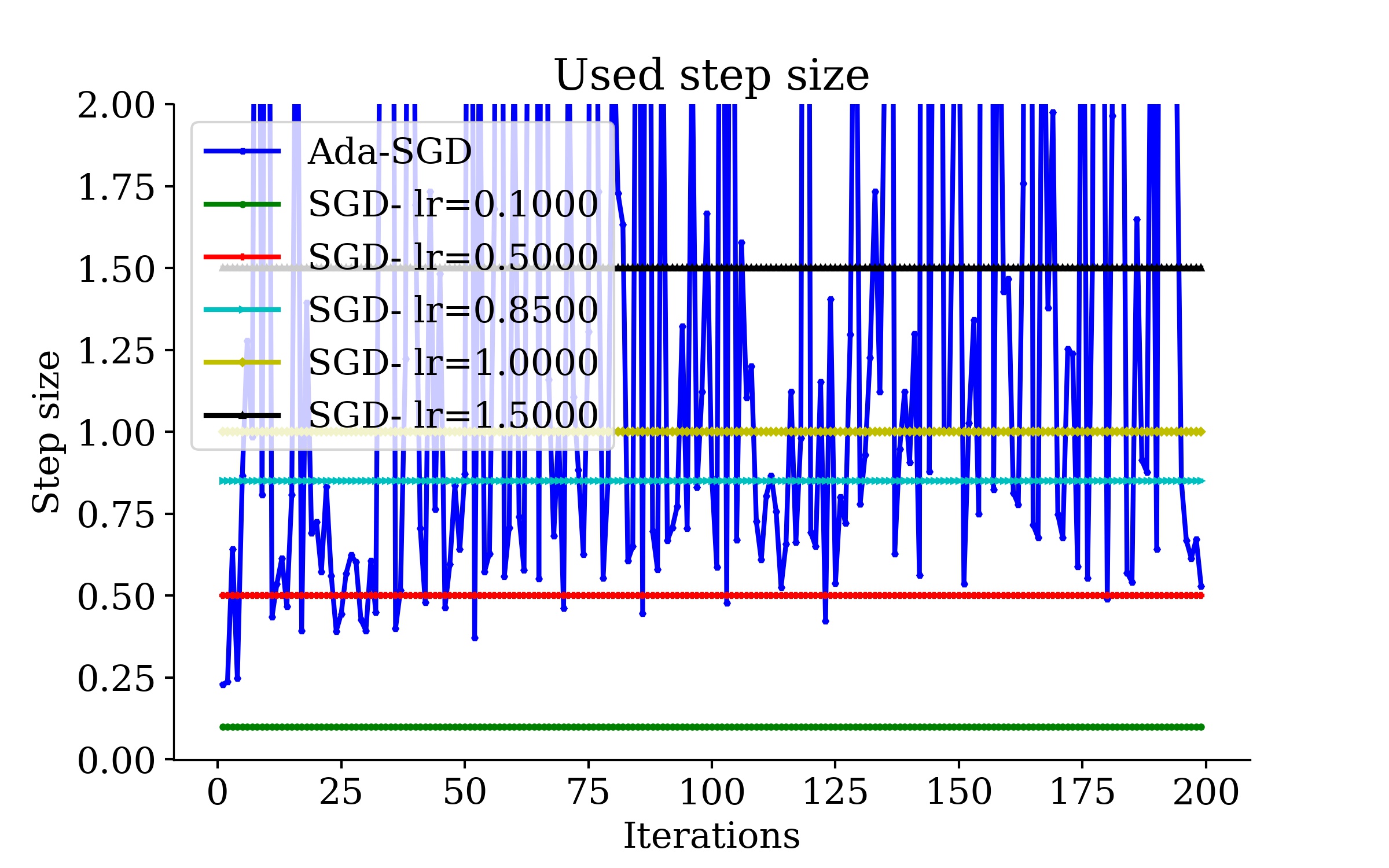} \\
 \includegraphics[width=53mm]{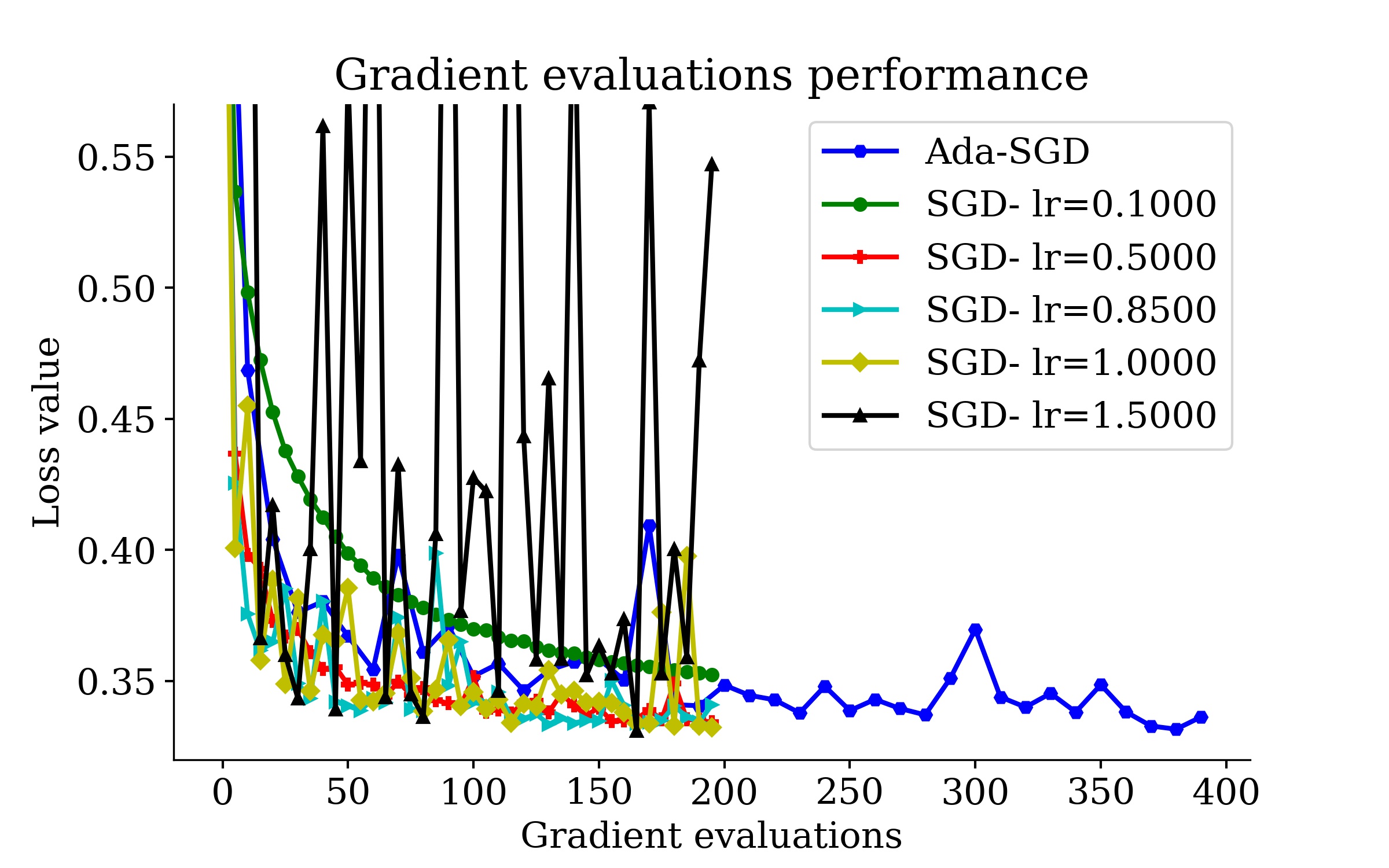} &   \includegraphics[width=53mm]{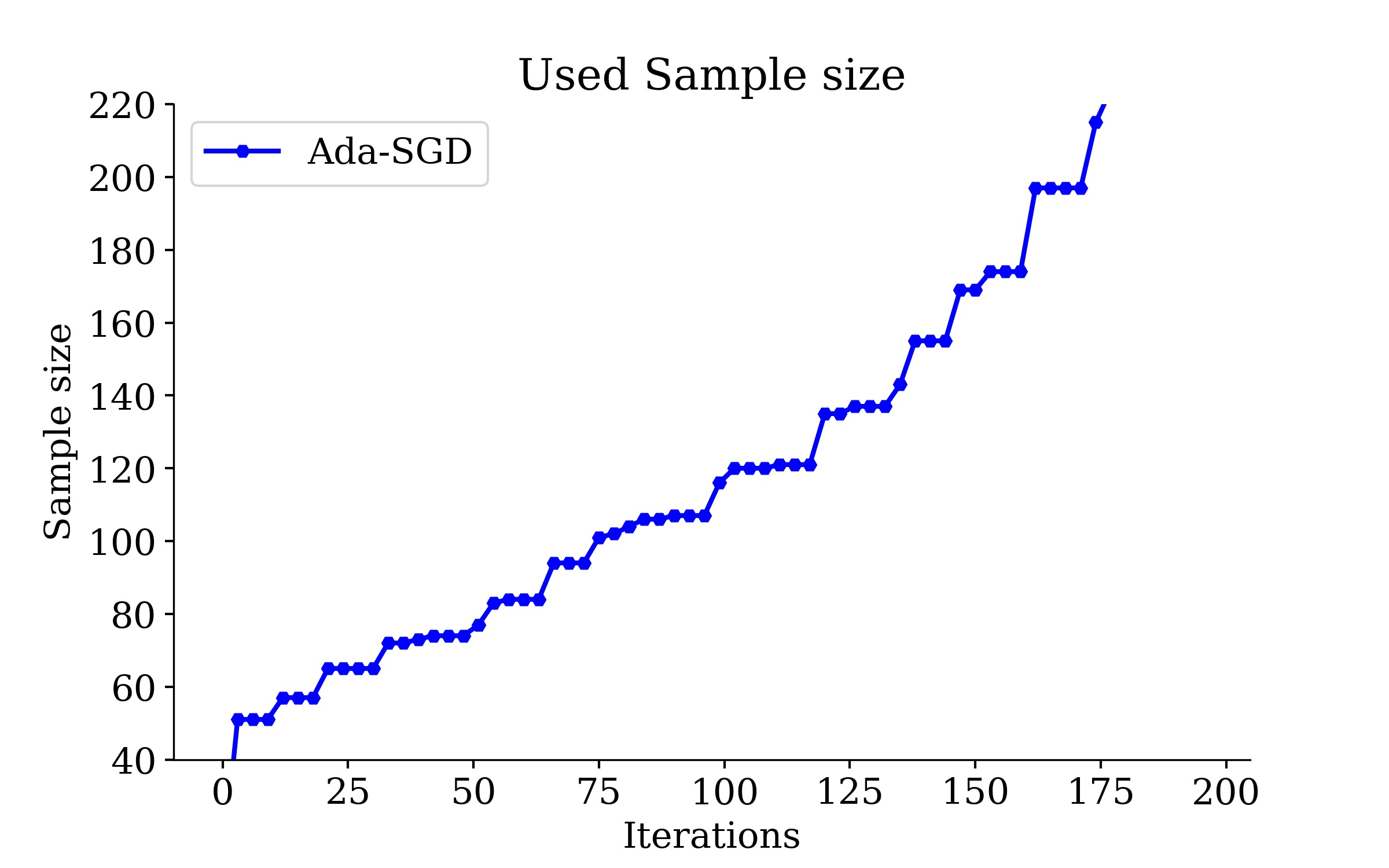} & 
 \includegraphics[width=53mm]{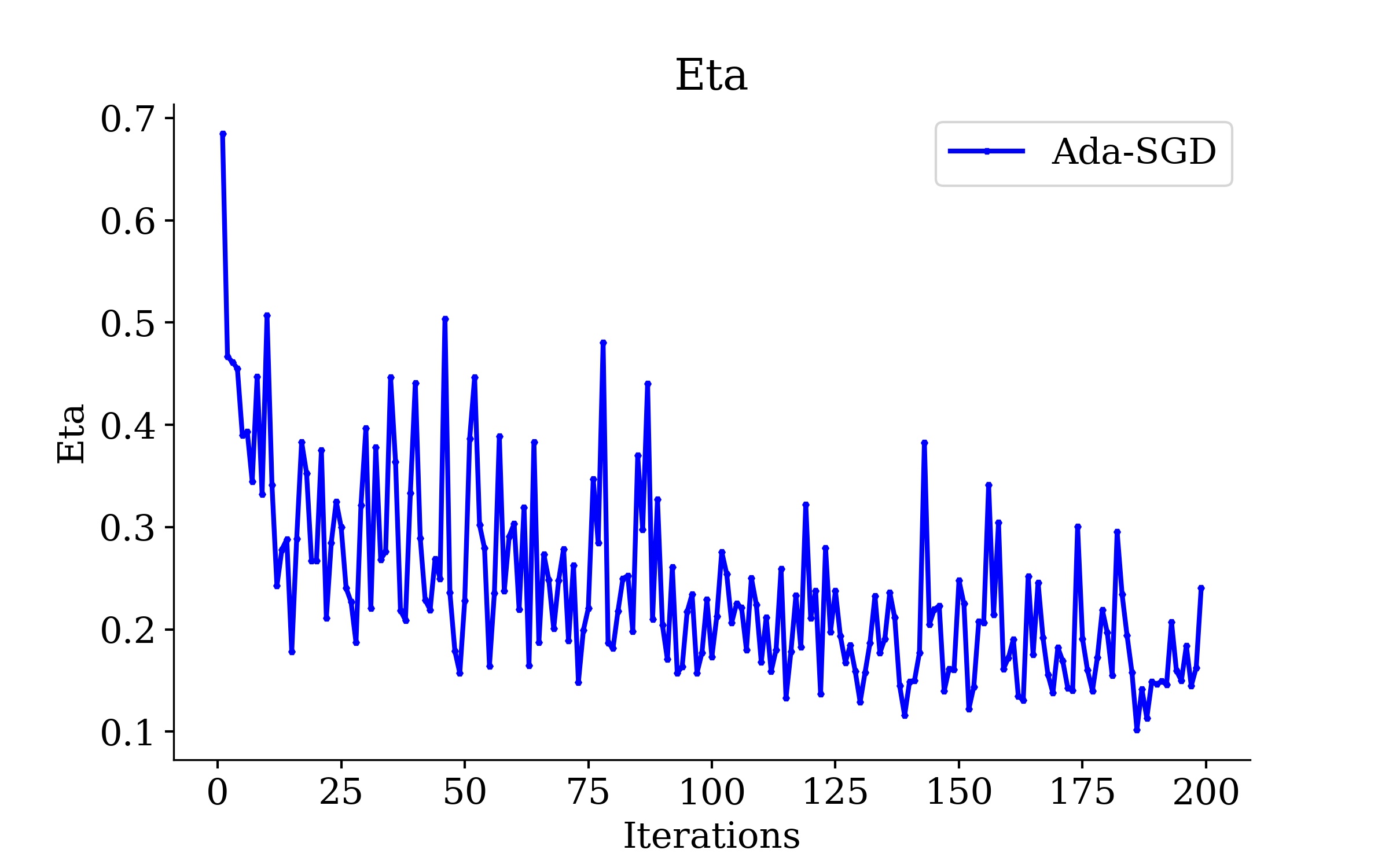} 
\end{tabular}
\caption{Numerical results logistic regression on a1a dataset}
\label{a1a}
\end{figure}

\begin{figure}
\begin{tabular}{ccc}
  \includegraphics[width=53mm]{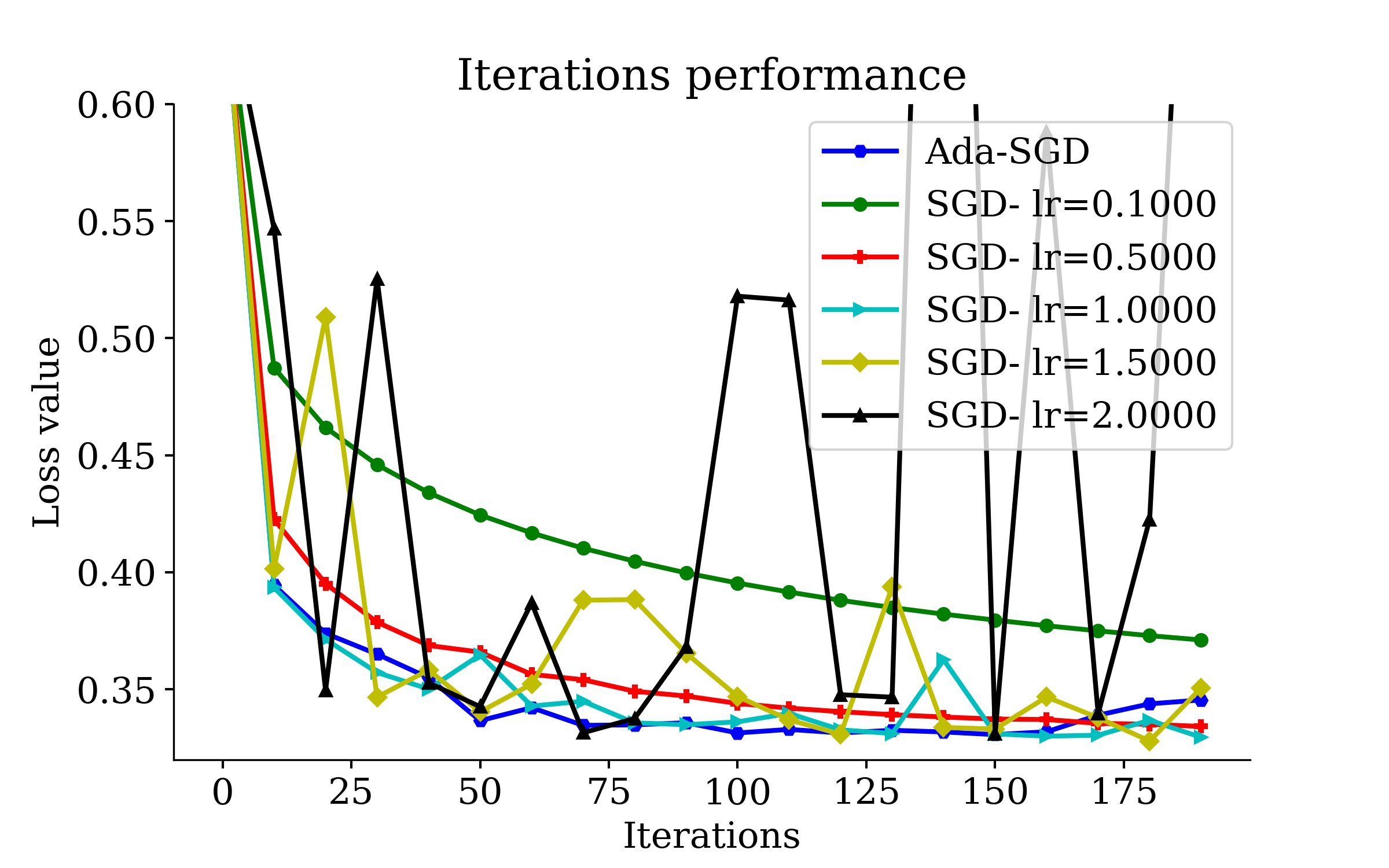} & \includegraphics[width=53mm]{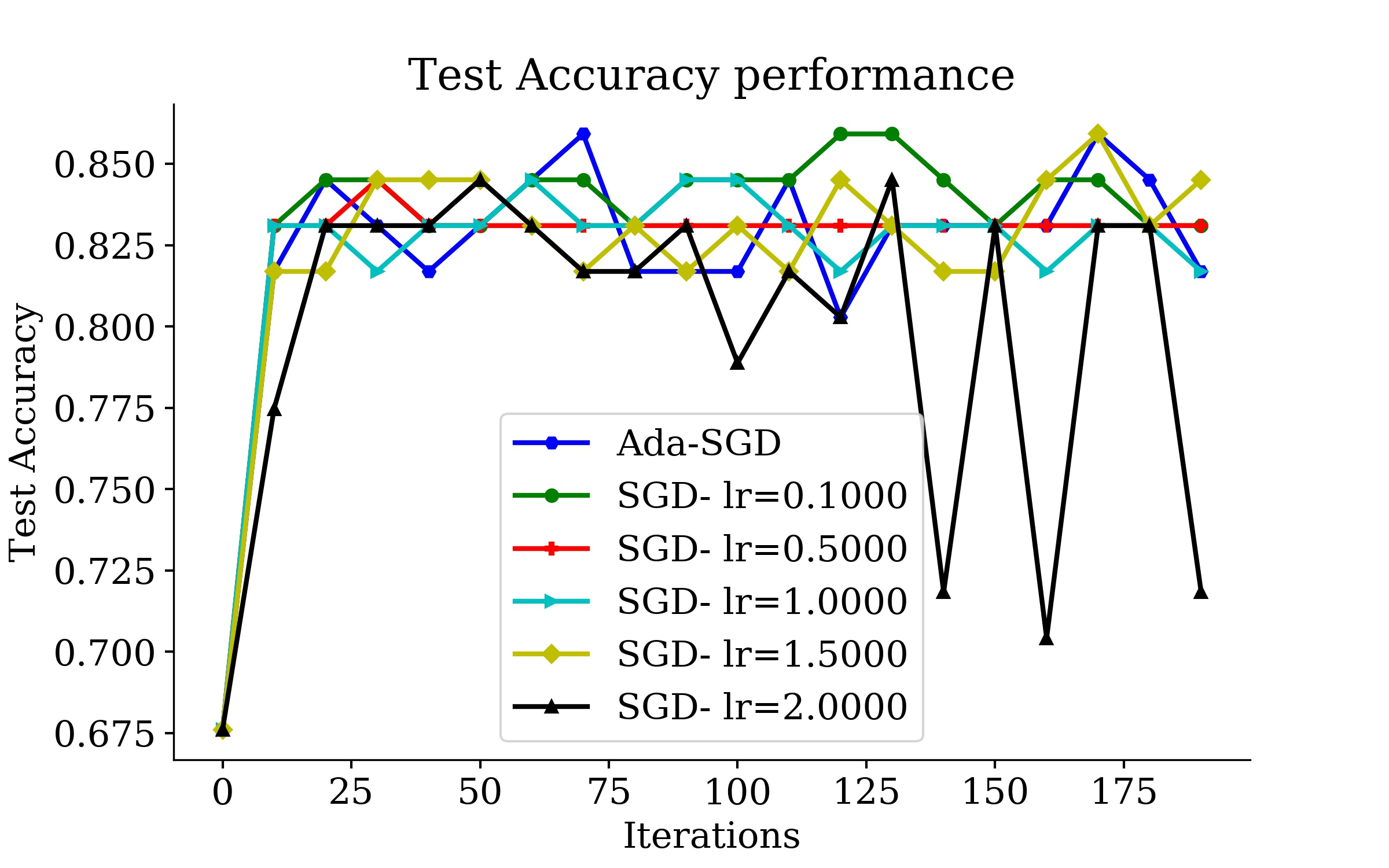} & \includegraphics[width=53mm]{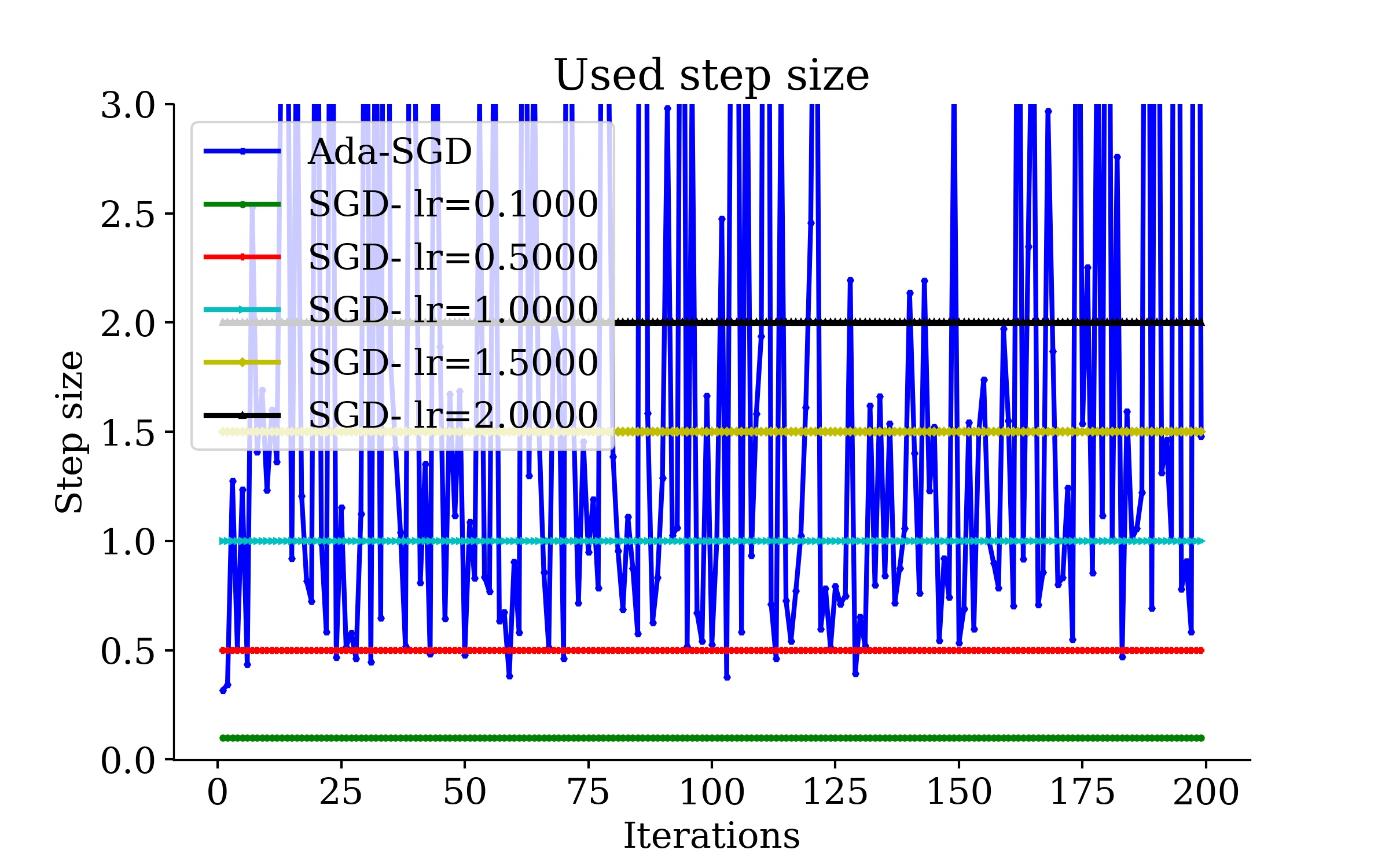} \\
 \includegraphics[width=53mm]{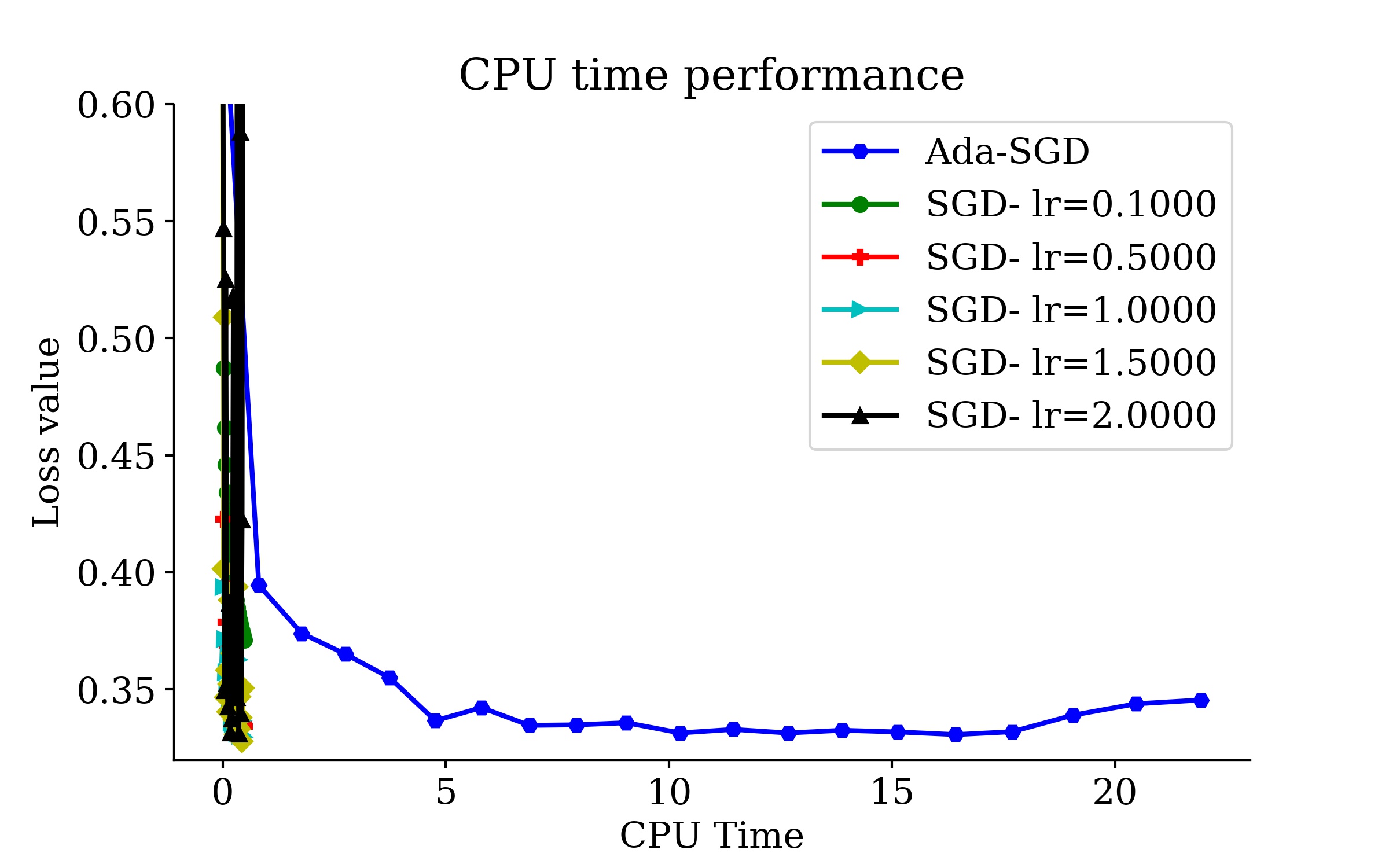} &   \includegraphics[width=53mm]{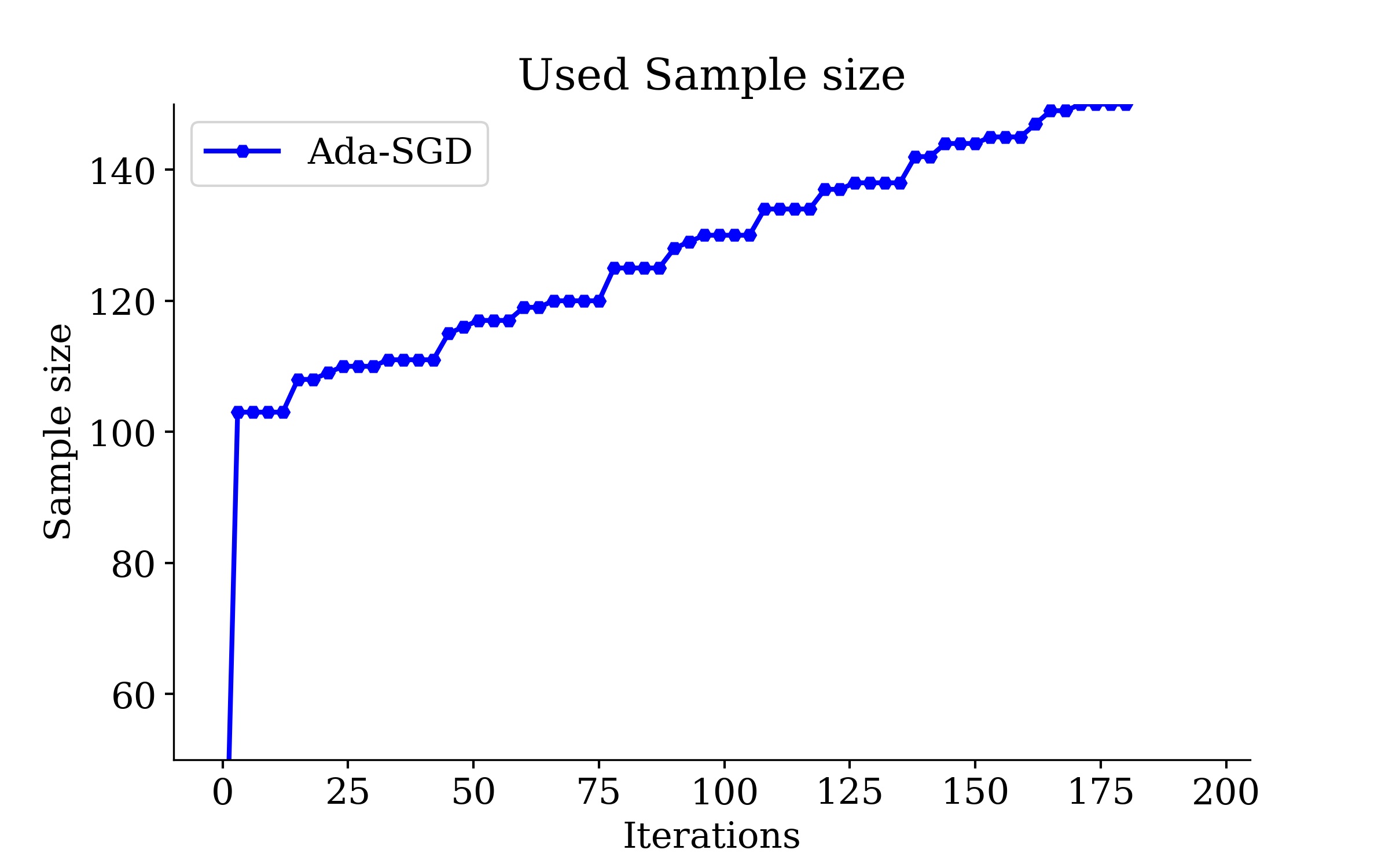} & 
 \includegraphics[width=53mm]{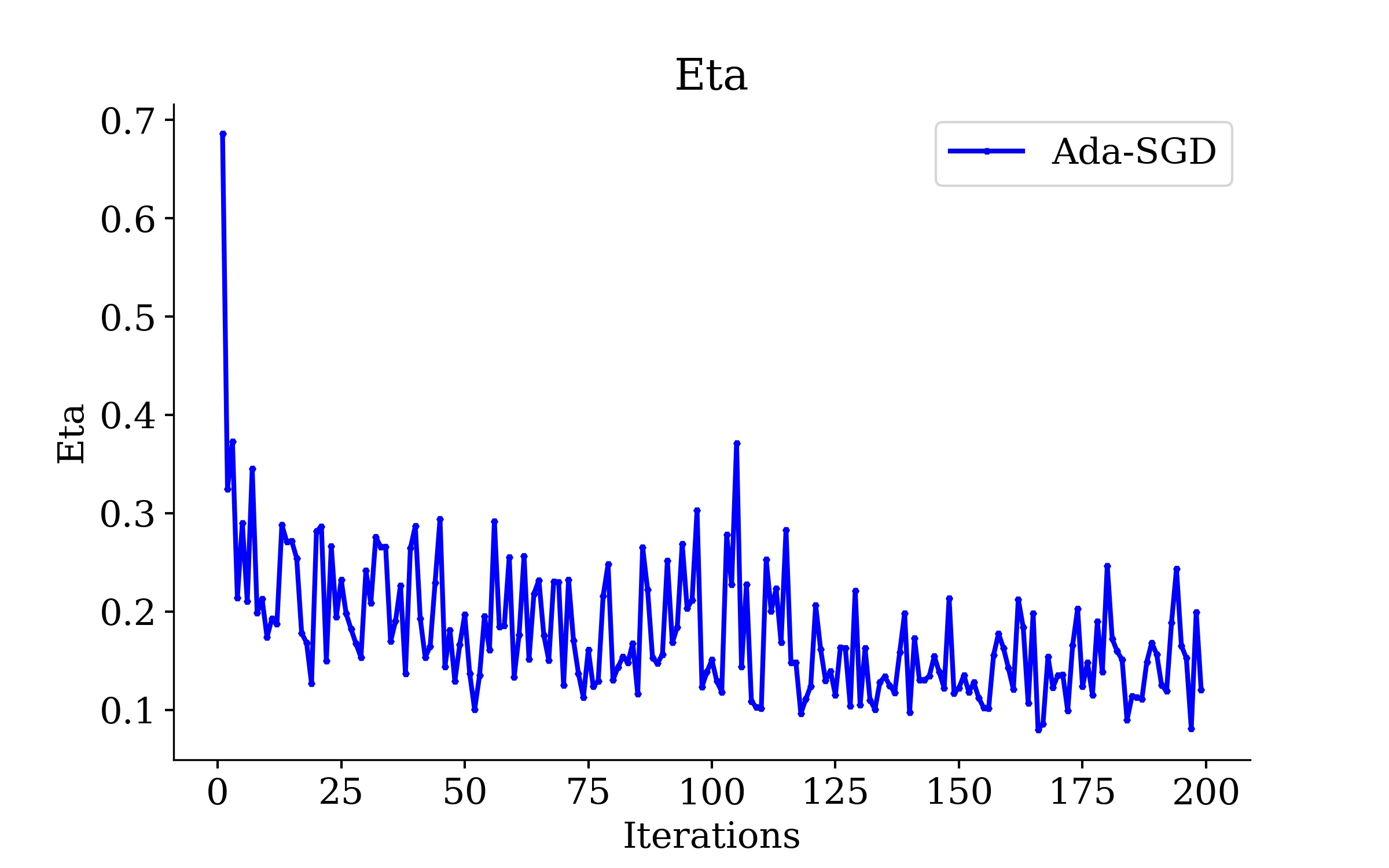} 
\end{tabular}
\caption{Numerical results logistic regression on inosphere dataset}
\label{inoshpere}
\end{figure}

\begin{figure}
\begin{tabular}{ccc}
  \includegraphics[width=53mm]{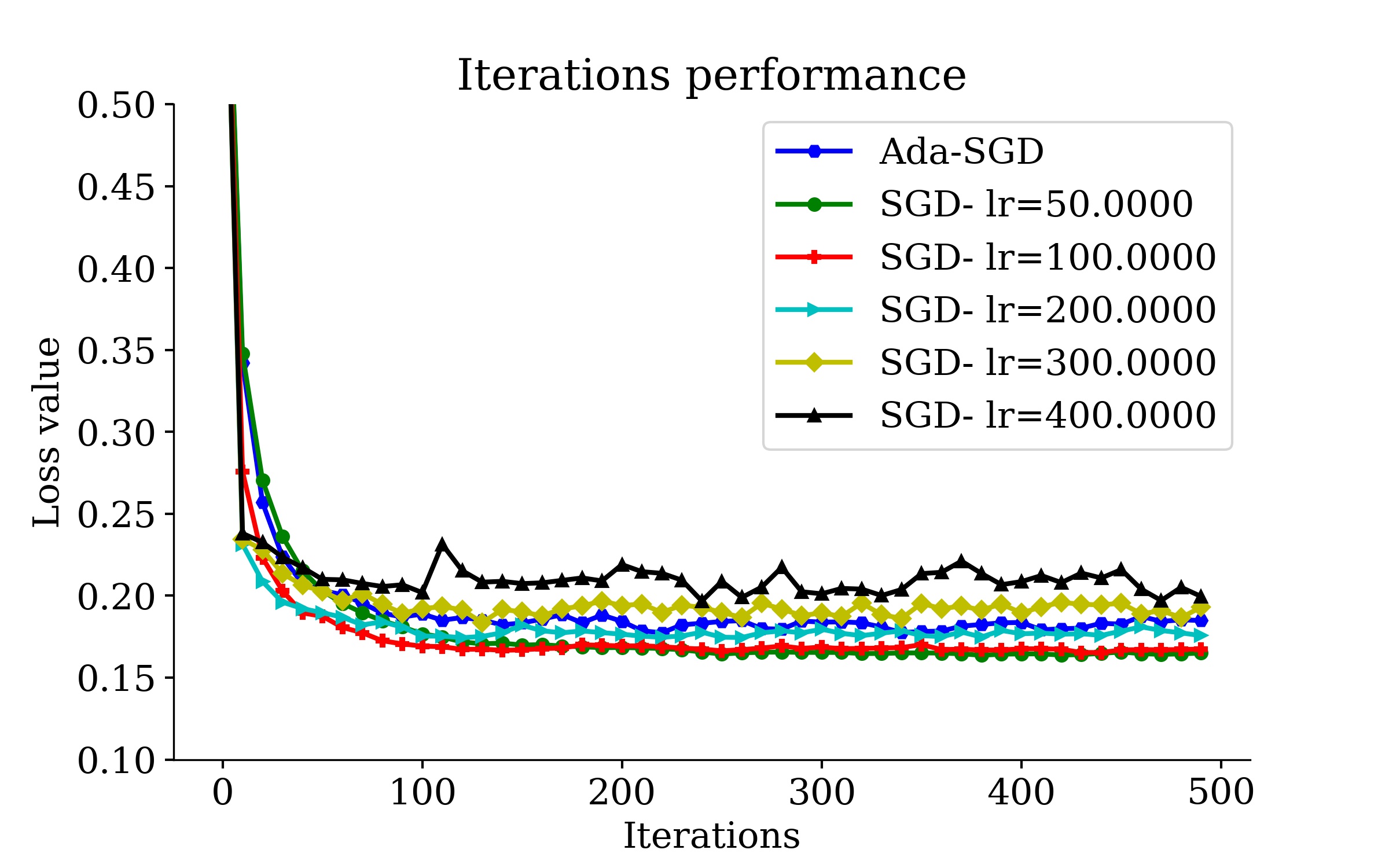} & \includegraphics[width=53mm]{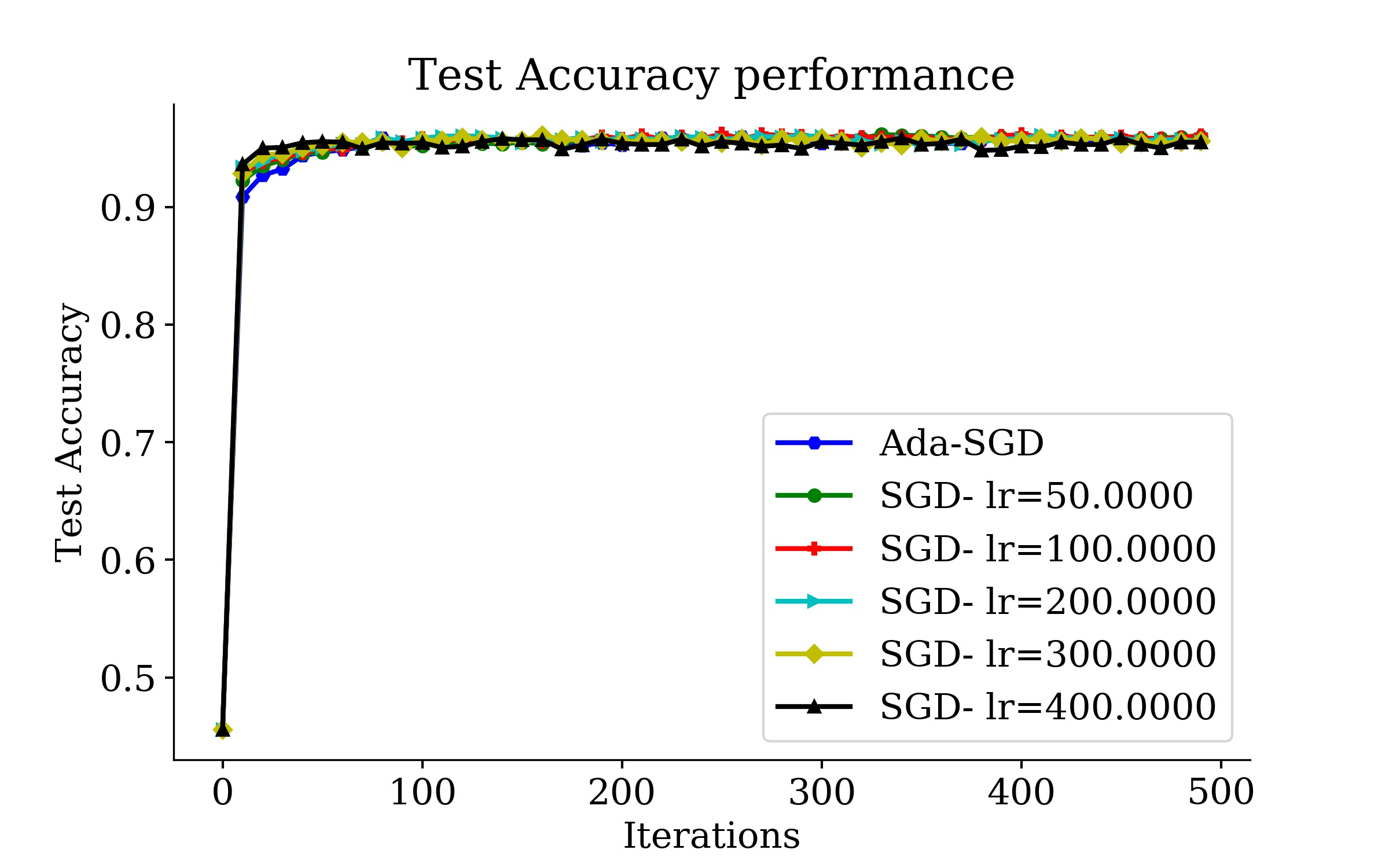} & \includegraphics[width=53mm]{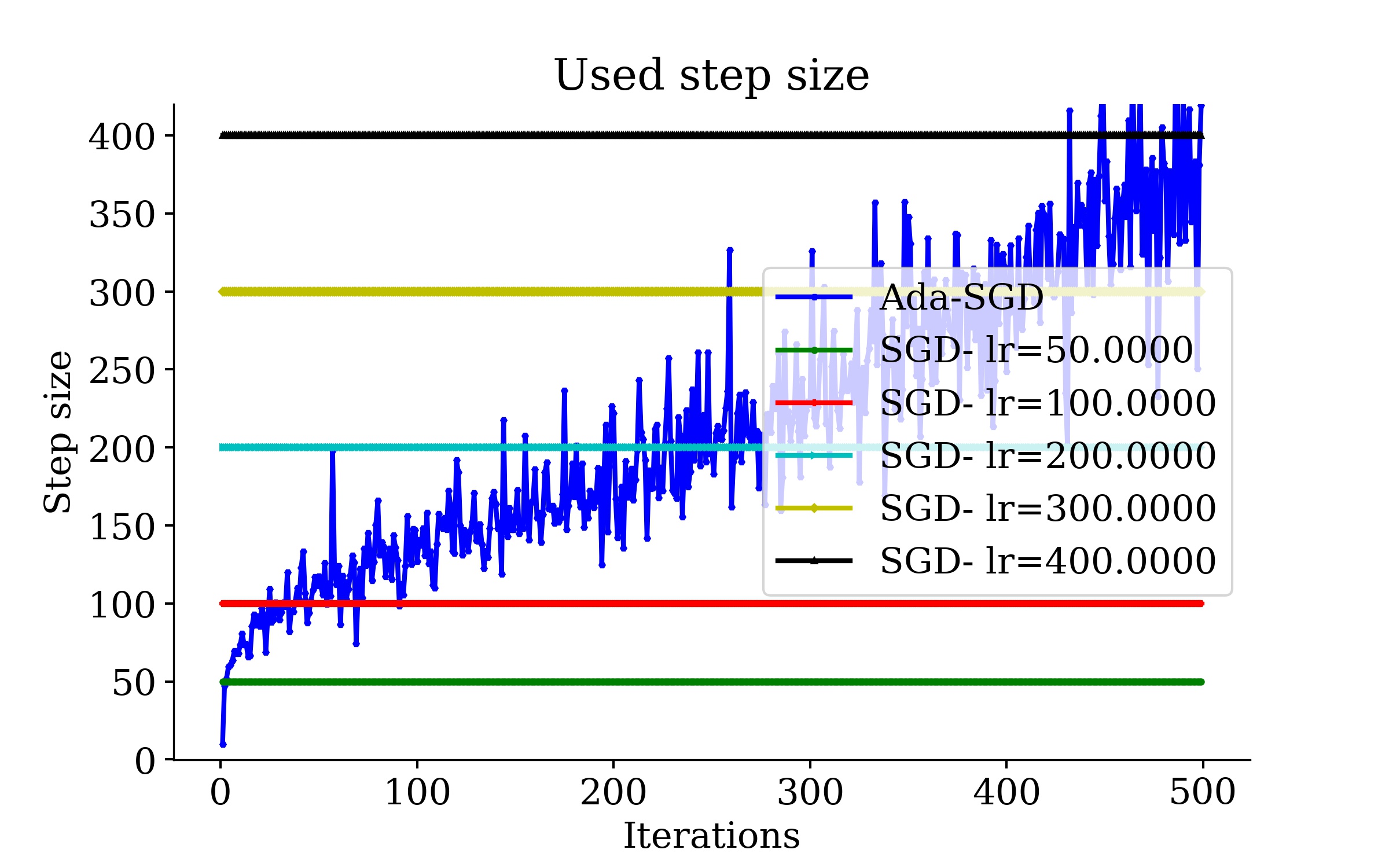} \\
 \includegraphics[width=53mm]{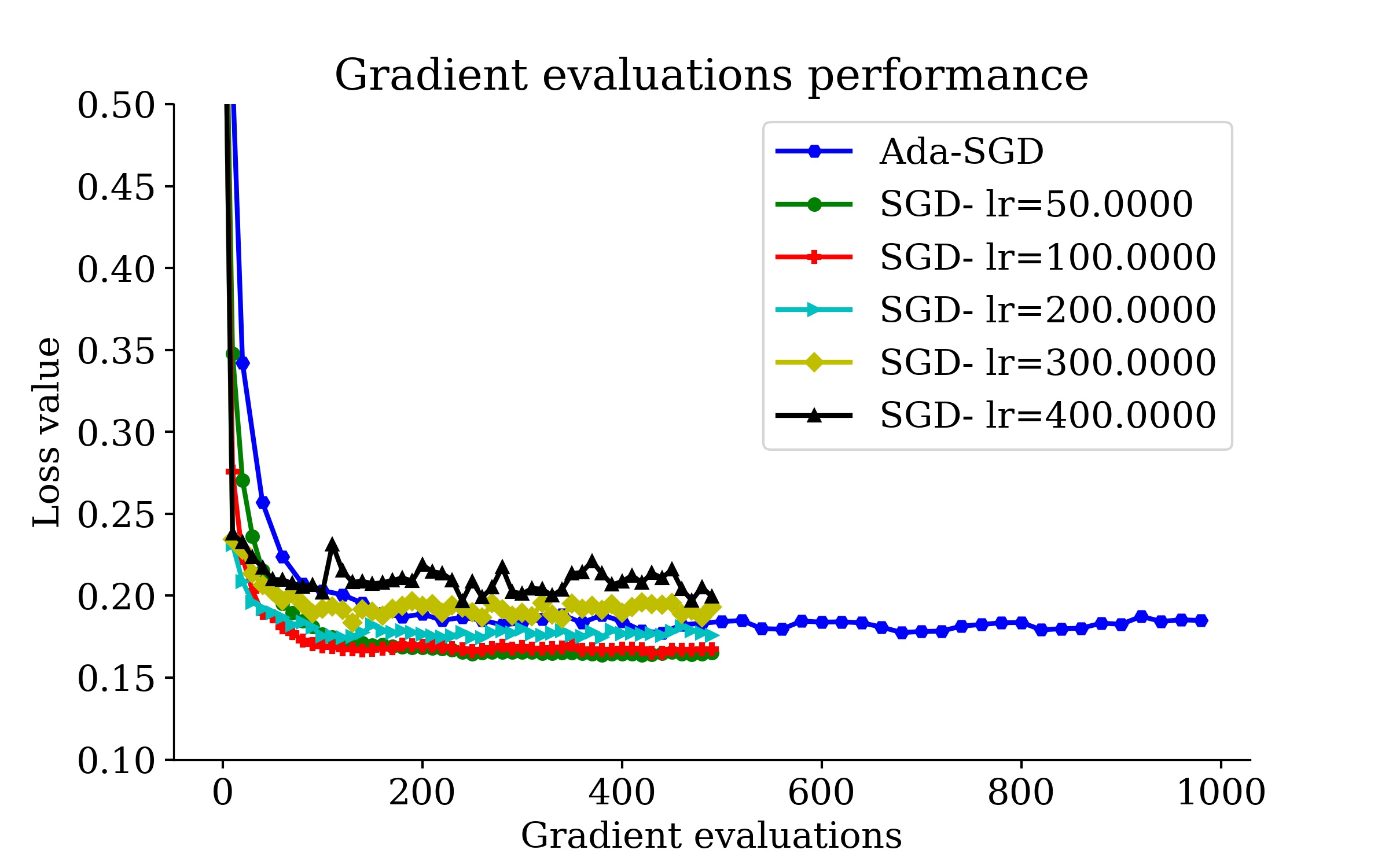} &   \includegraphics[width=53mm]{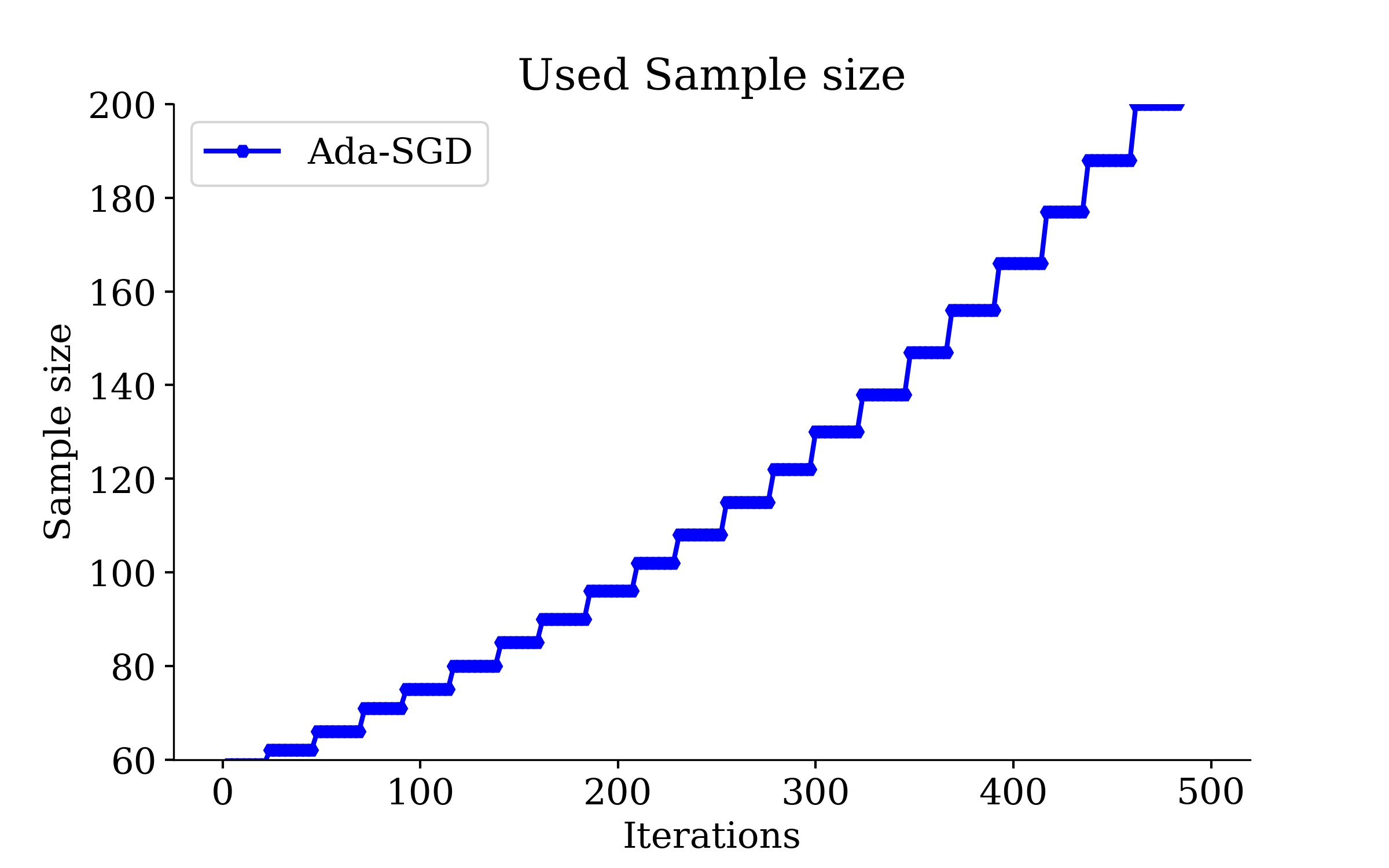} & 
 \includegraphics[width=53mm]{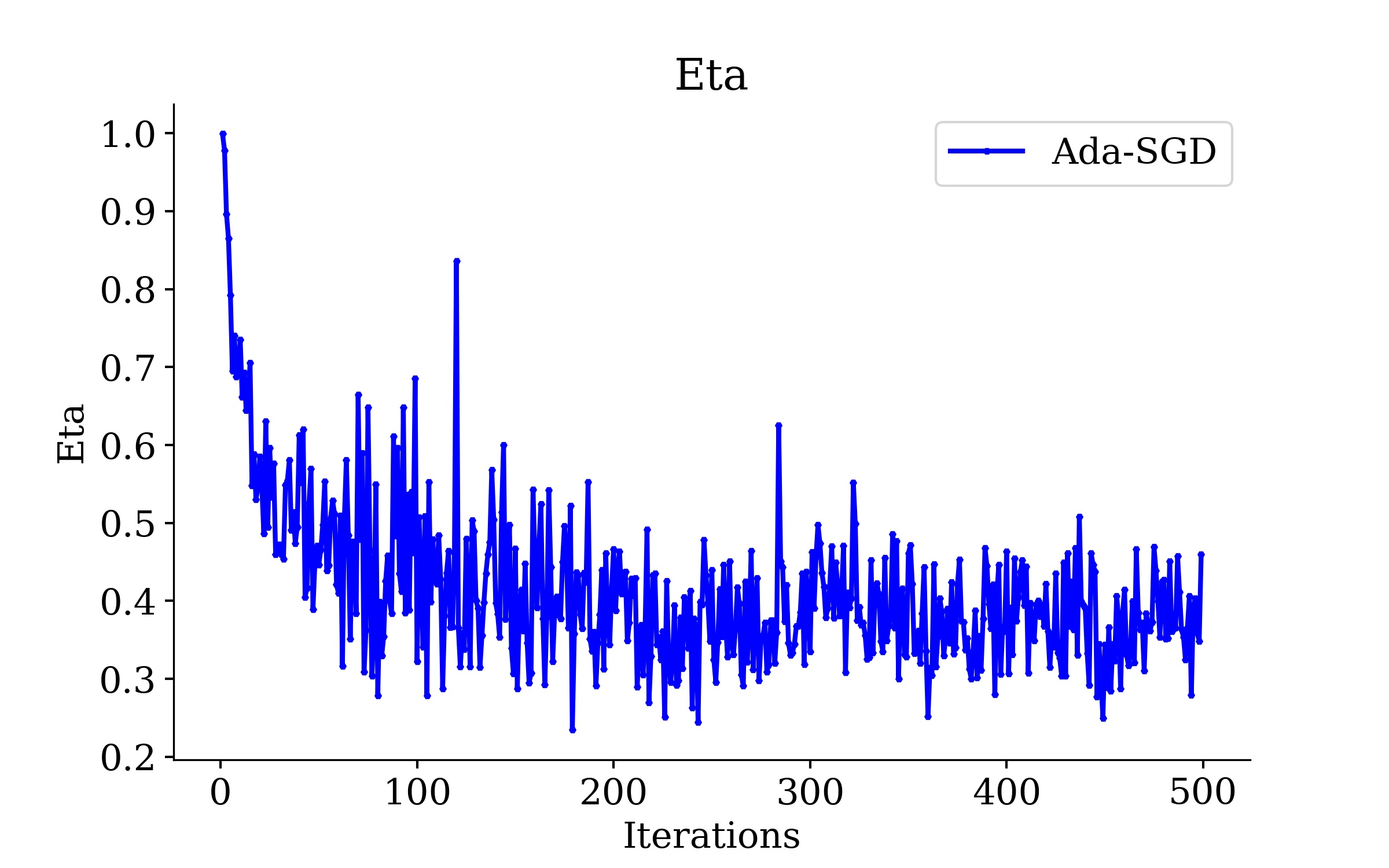} 
\end{tabular}
\caption{Numerical results logistic regression on rcv1 dataset}
\label{rcv1}
\end{figure}

\begin{figure}
\begin{tabular}{ccc}
  \includegraphics[width=53mm]{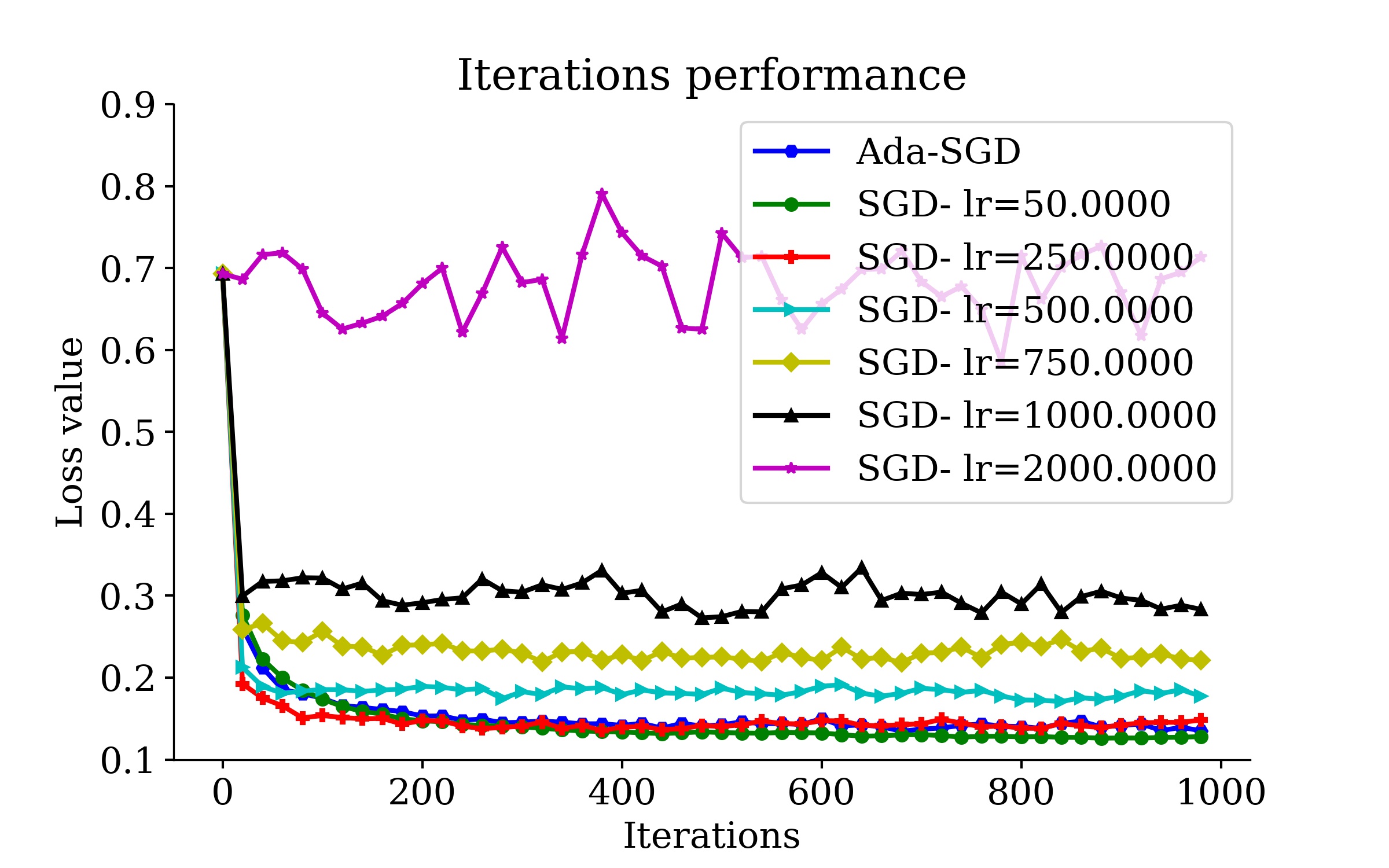} & \includegraphics[width=53mm]{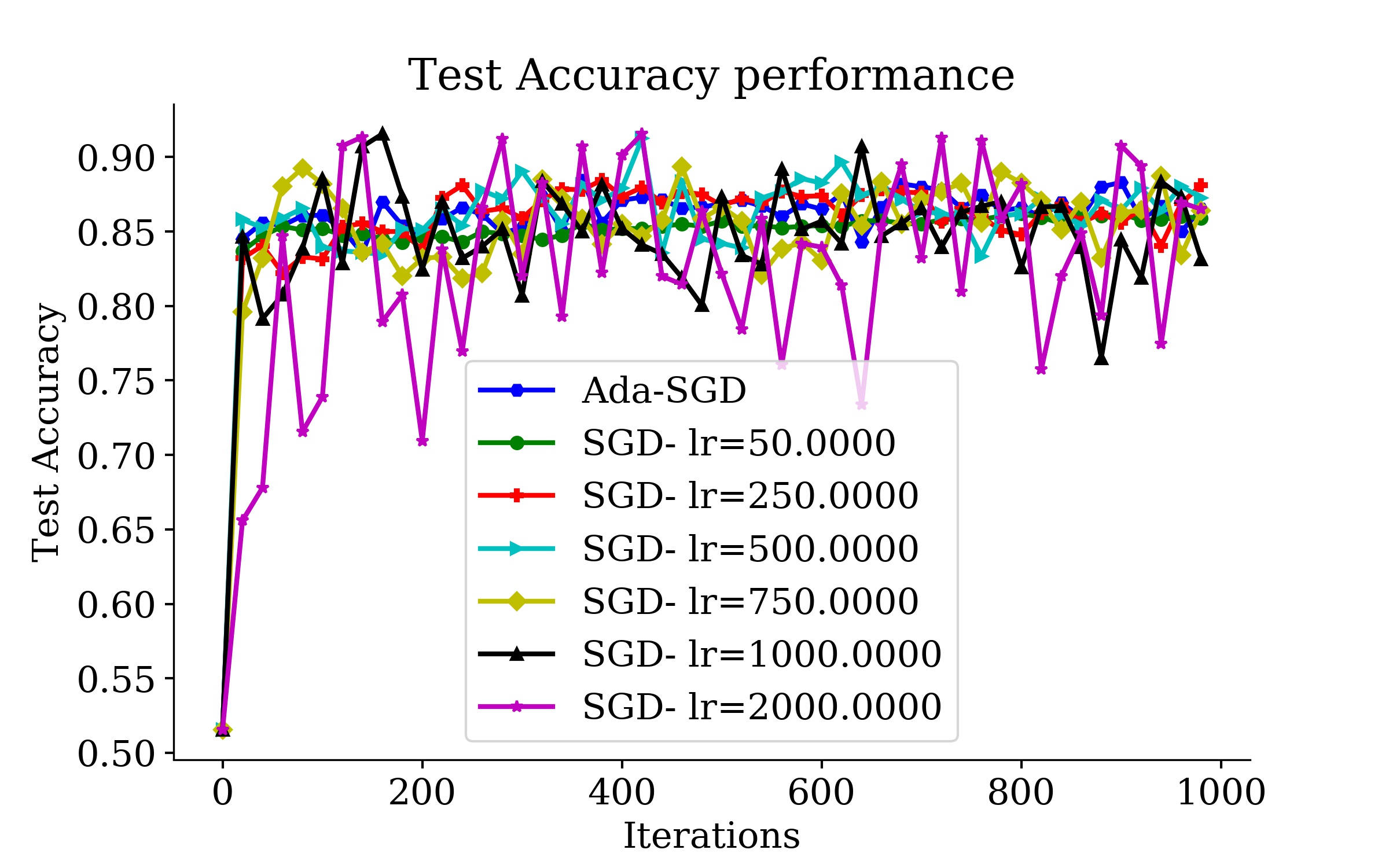} & \includegraphics[width=53mm]{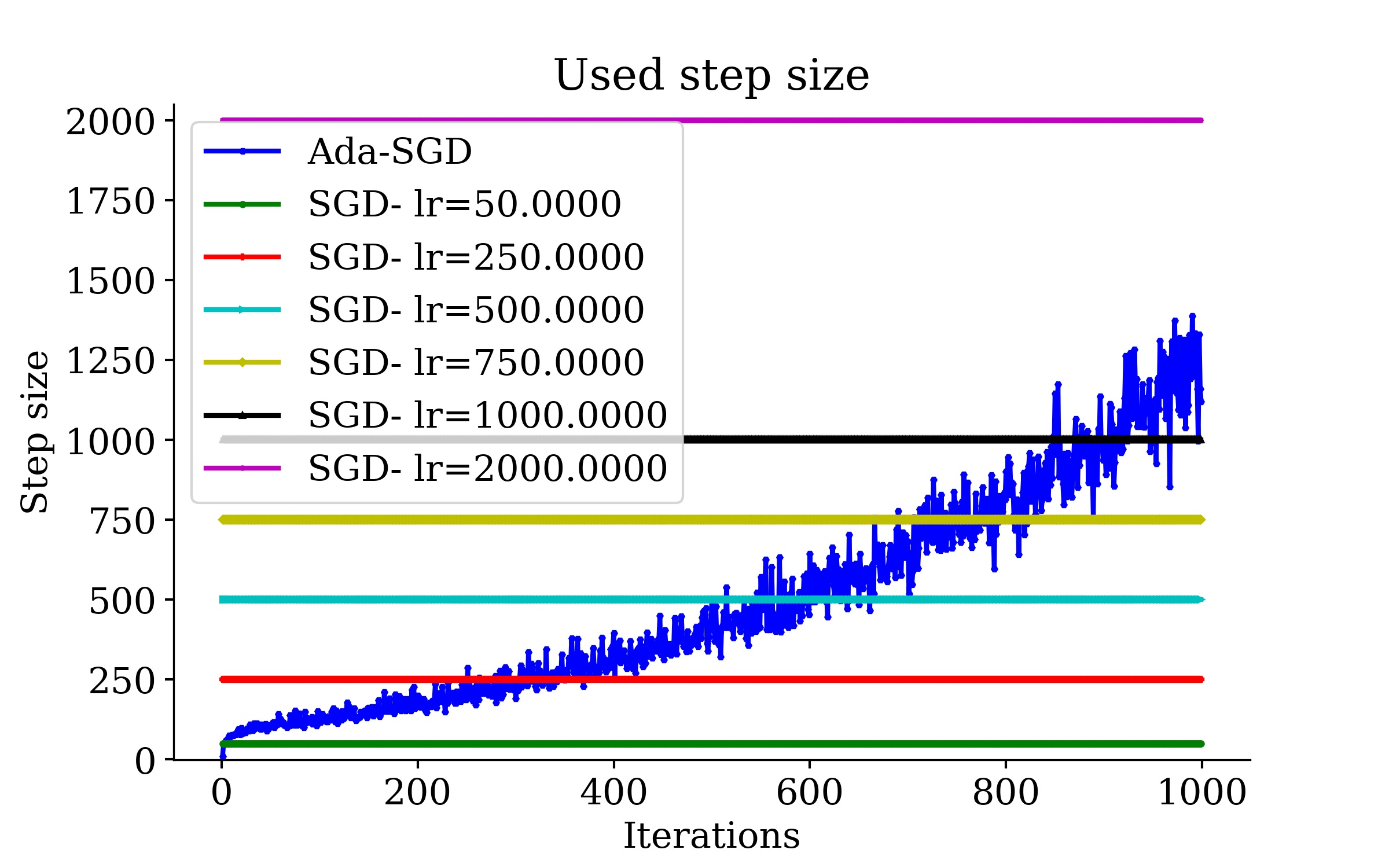} \\
 \includegraphics[width=53mm]{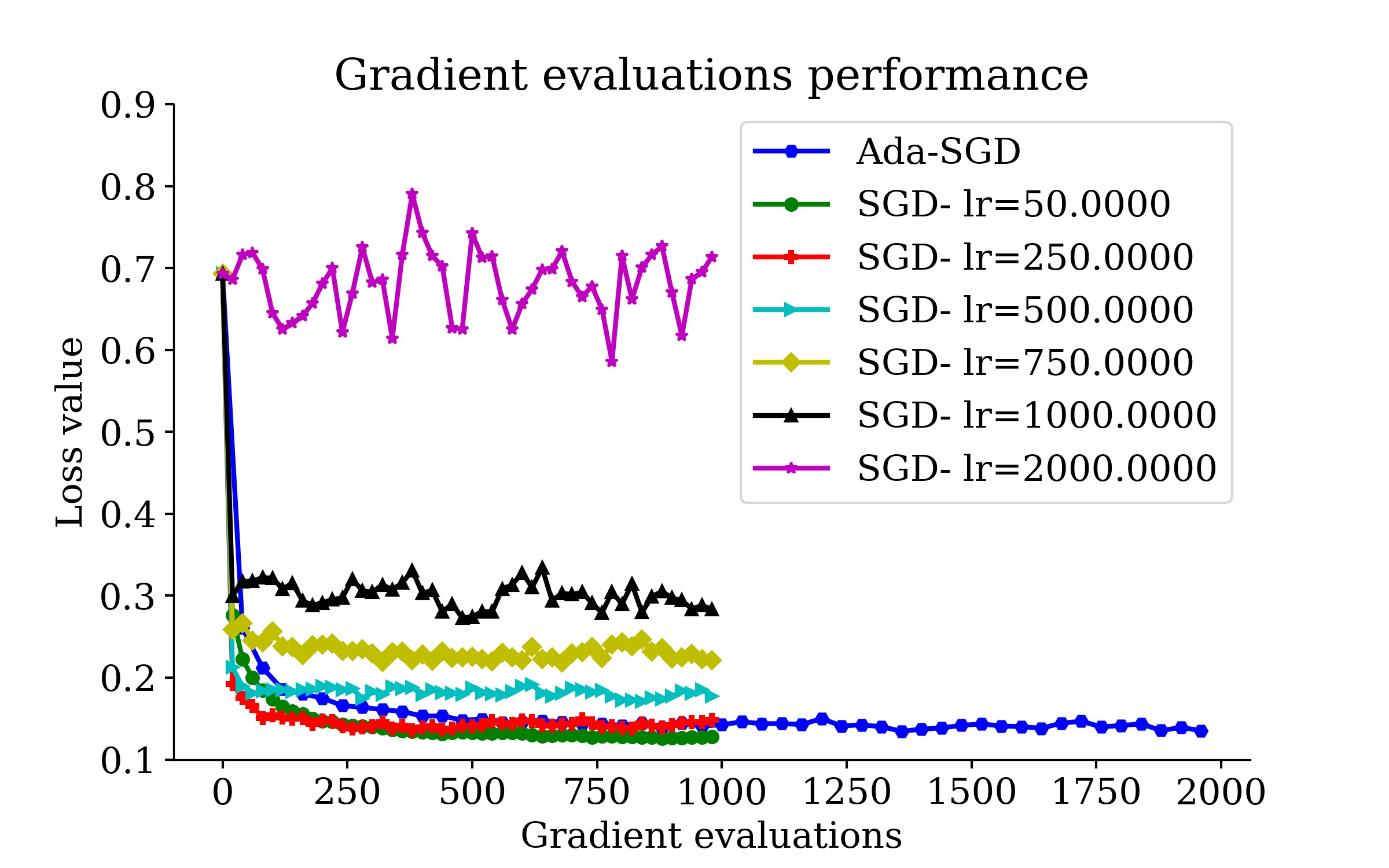} &   \includegraphics[width=53mm]{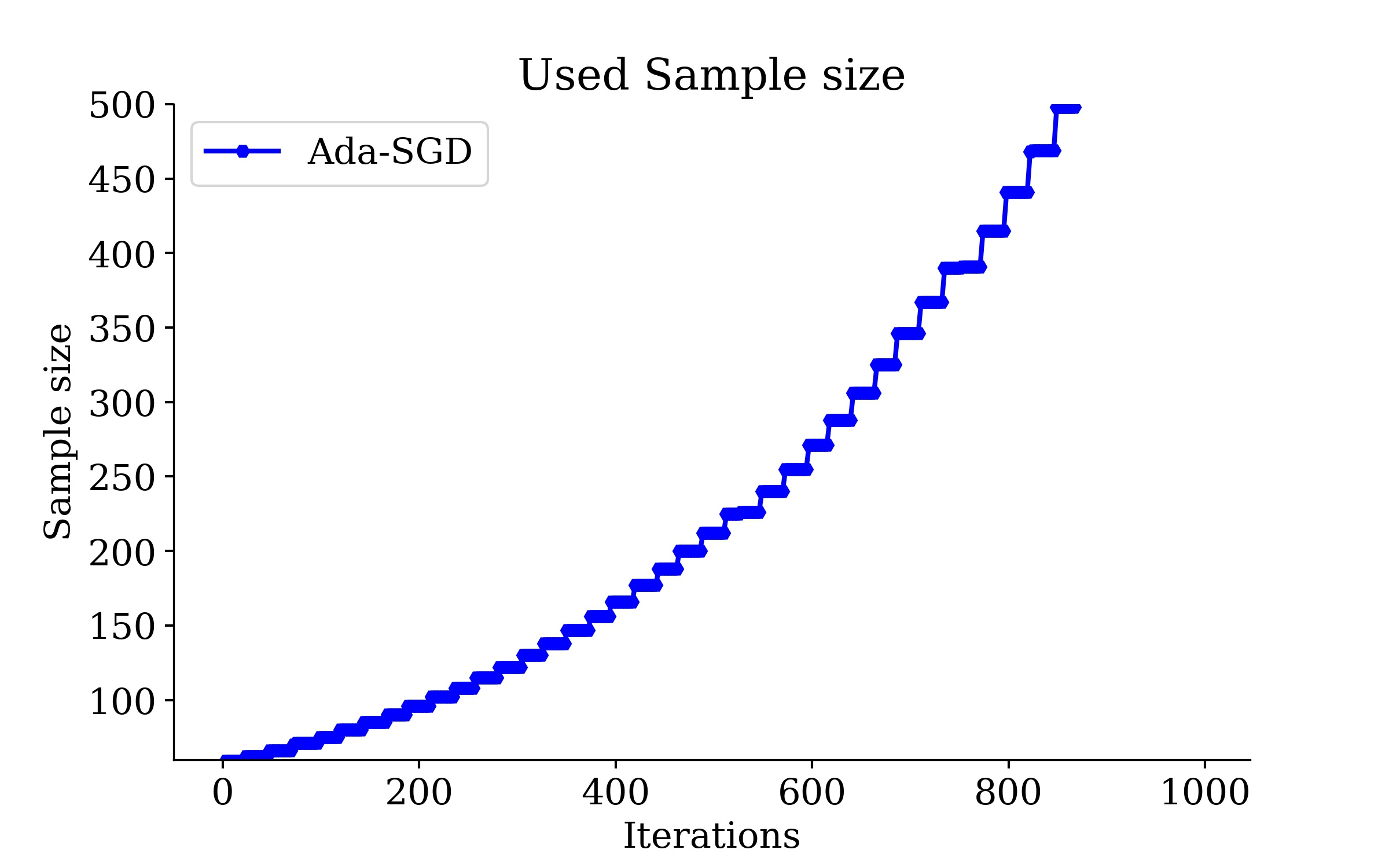} & 
 \includegraphics[width=53mm]{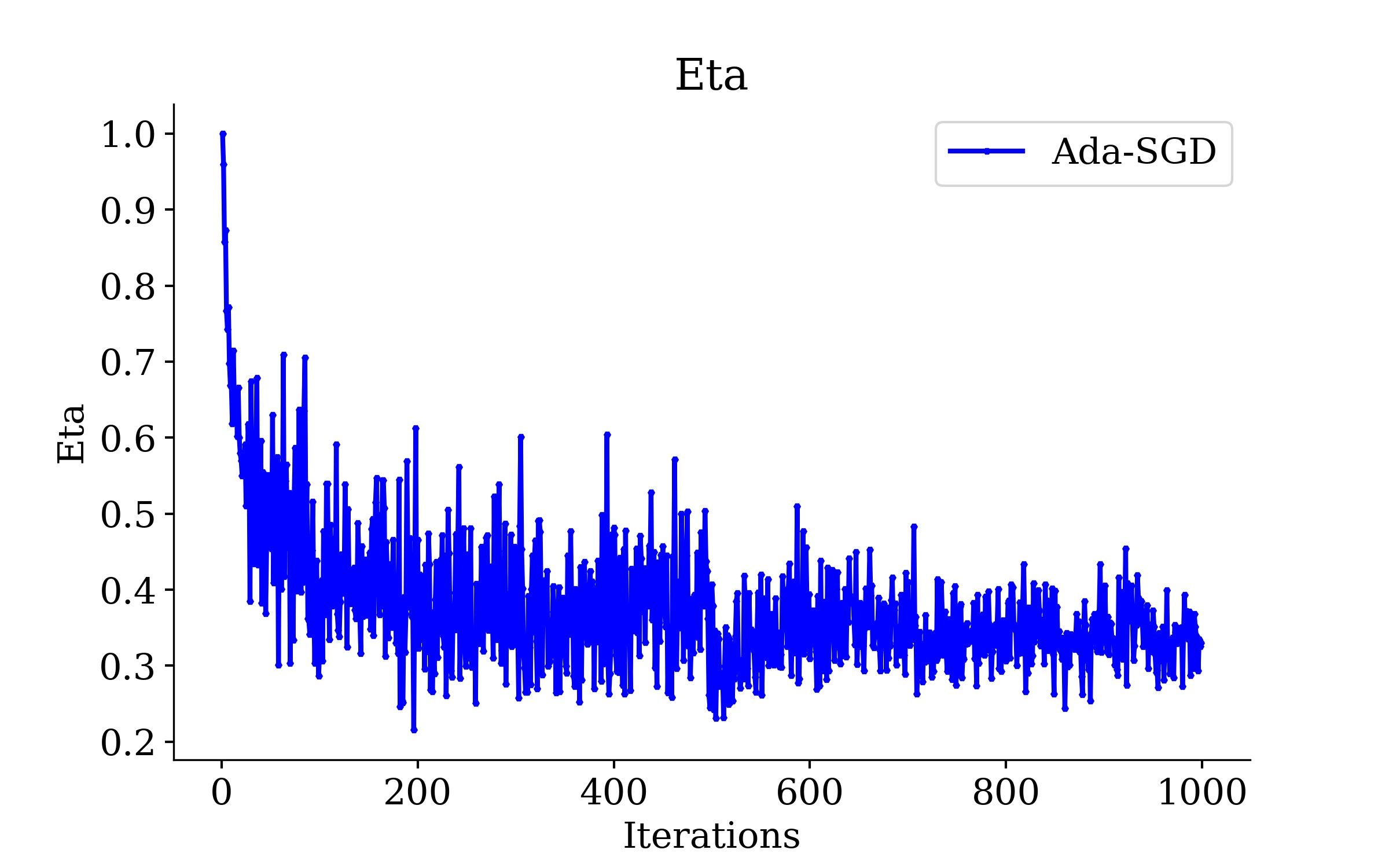} 
\end{tabular}
\caption{Numerical results logistic regression on realsim dataset}
\label{realsim}
\end{figure}
\clearpage

\subsection{Deep Neural Networks Experiments}

MNIST is a database of handwritten digits (0 to 9) that contains 60,000 training images and 10,000 testing images each of size 28x28. In MNIST CNN experiment in the main text, we used the following hyperparameters:

\begin{center}
 \begin{tabular}{||c | c||} 
 \hline
 Hyperparameter & Values  \\ [0.5ex] 
 \hline\hline
 Architecture of CNN & \{Conv2d(1,16,5)+BN+ReLU+MaxPool(2,2), \\ & Conv2d(16,32,5)+BN+ReLU+MaxPool(2,2), \\
 & Linear(1568,10)\}  \\ 
 \hline
 probability of test failure $p$ & \{0.1\}  \\ 
 \hline
 level of angle acuteness $\nu$ & \{0.1\}  \\
 \hline
 initial batch size & \{16\} \\
 \hline
 level of curvature precision $\epsilon$ & \{0.01\} \\
 \hline
\end{tabular}
\end{center}

The CIFAR10 dataset consists of 60,000 32x32 color images of 10 object categories (50,000 training and 10,000 testing ones). We use raw images without pre-processing. Data augmentation follows the standard manner: during training, we zero-pad 4 pixels along each image side, and sample a 32x32 region crop from the padded image or its horizontal flip; during testing, we use the original non-padded image and we report the top-1 accuracy.

For CIFAR10, we used architectures of VGG \cite{VGGNet}, ResNet \cite{resnetpaper} and DavidNet (custom ResNet)
on CIFAR10 to study the behaviours of adaptive learning rates compared to the established baseline found in \cite{resnetpaper} (0.9 momentum, 0.1 learning rate and the learning rate decay schedule occurs when the epoch number is \{3/9, 6/9, 8/9\} of the maximum epoch length).

\subsubsection{Results on DavidNet:}

\begin{figure}
\centering
  \includegraphics[width=102mm]{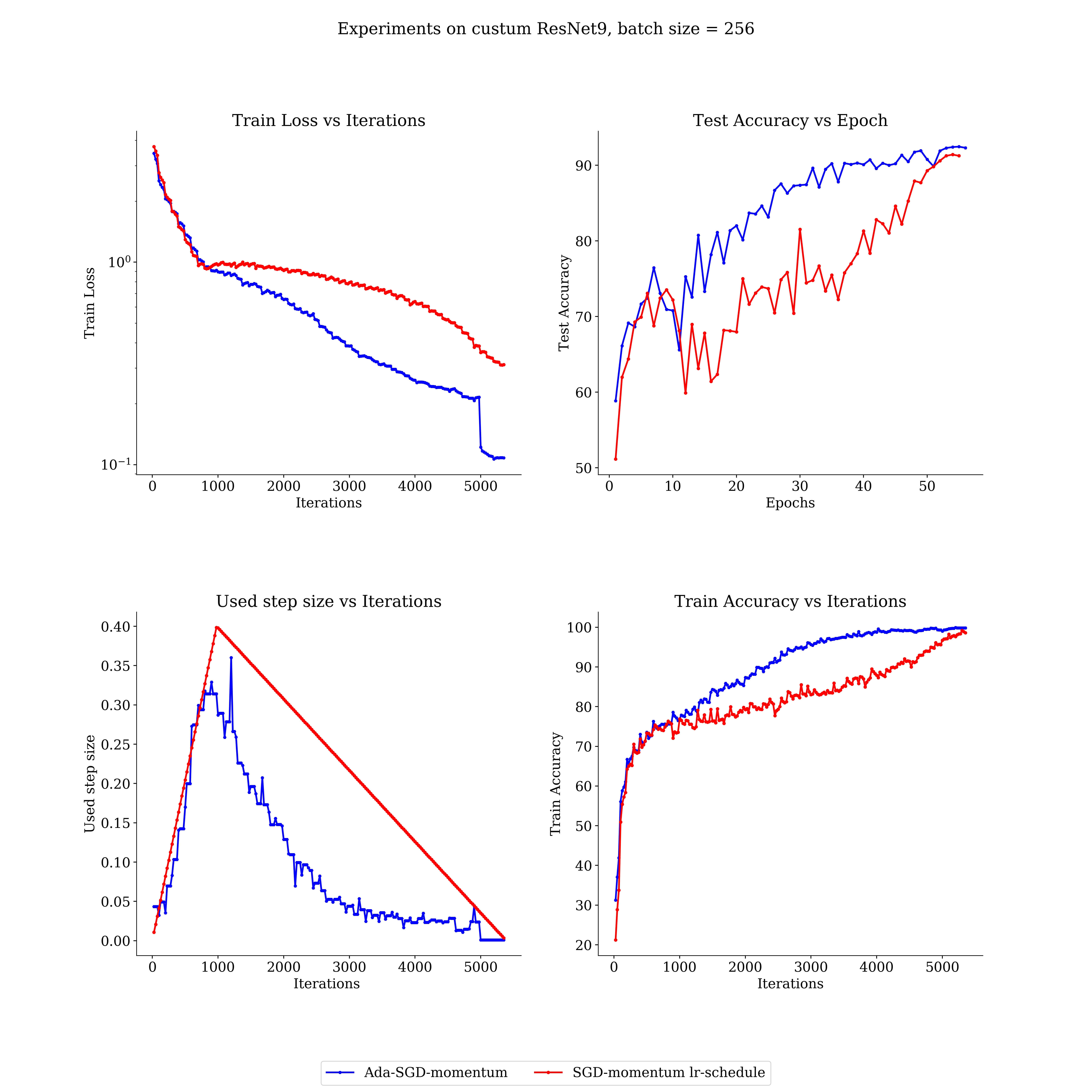}
\caption{Numerical results on CIFAR10 using DavidNet model}
\label{davidcifar}
\end{figure}

According to DAWNBench\footnote{ \url{https://dawn.cs.stanford.edu/benchmark/CIFAR10/train.html}.}, DavidNet\footnote{ \url{https://myrtle.ai/how-to-train-your-resnet-4-architecture/}.} (a custom 9-layer Residual ConvNet proposed by David
C. Page.) is the fastest model for CIFAR-10 dataset (as of April 1st, 2019). We choose to experiment with this model because the author made a considerable effort to find a near-optimal piece-wise learning-rate schedule and we wanted to see whether our method mimics this schedule. Results are presented in Figure \ref{davidcifar}.

\subsubsection{Numerical Results on Ada-ADAM:}

We also ran experiments on the scale invariant version of our adaptive framework using ADAM update direction on two neural networks settings described bellow. For both settings we fixed the batch-size and only exploited the adaptive learning rate part of our framework.

\begin{figure}
\centering
  \includegraphics[width=102mm]{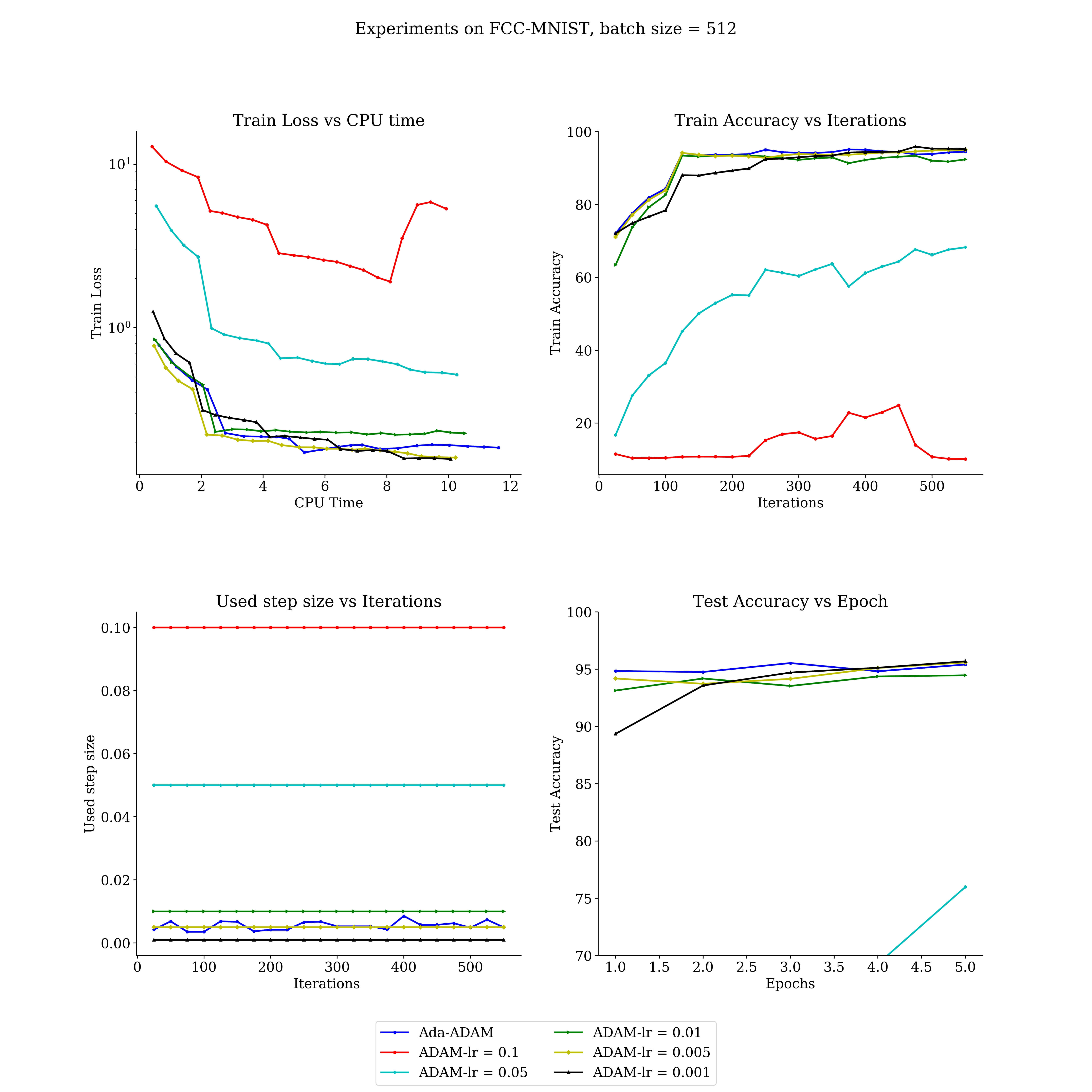}
\caption{Ada-ADAM Numerical results on MNIST}
\label{adadammnist}
\end{figure}

{\bf Fully Connected MNIST:} We trained a two-layers fully connected neural network on MNIST dataset using Ada-ADAM and ADAM with various learning rates, the hyper-parameters for this experiment are:

\begin{center}
 \begin{tabular}{||c | c||} 
 \hline
 Hyperparameter & Values  \\ [0.5ex] 
 \hline\hline
 Architecture of FCC & \{Linear(28*28, 100, bias=true) + ReLU \\ & Linear(100, 100, bias=true) + ReLU, \\
 & Linear(100,10)\}  \\ 
 \hline
  ADAM $\beta_1$ & \{0.9\}  \\ 
 \hline
  ADAM $\beta_2$ & \{0.999\}  \\
 \hline
 batch size & \{512\} \\
 \hline
 Constant learning rates comparison & \{0.1, 0.05, 0.01, 0.005, 0.001\} \\
 \hline
 level of curvature precision $\epsilon$ & \{0.01\} \\
 \hline
\end{tabular}
\end{center}
Results are presented in Figure \ref{adadammnist}.

\begin{figure}
\centering
  \includegraphics[width=102mm]{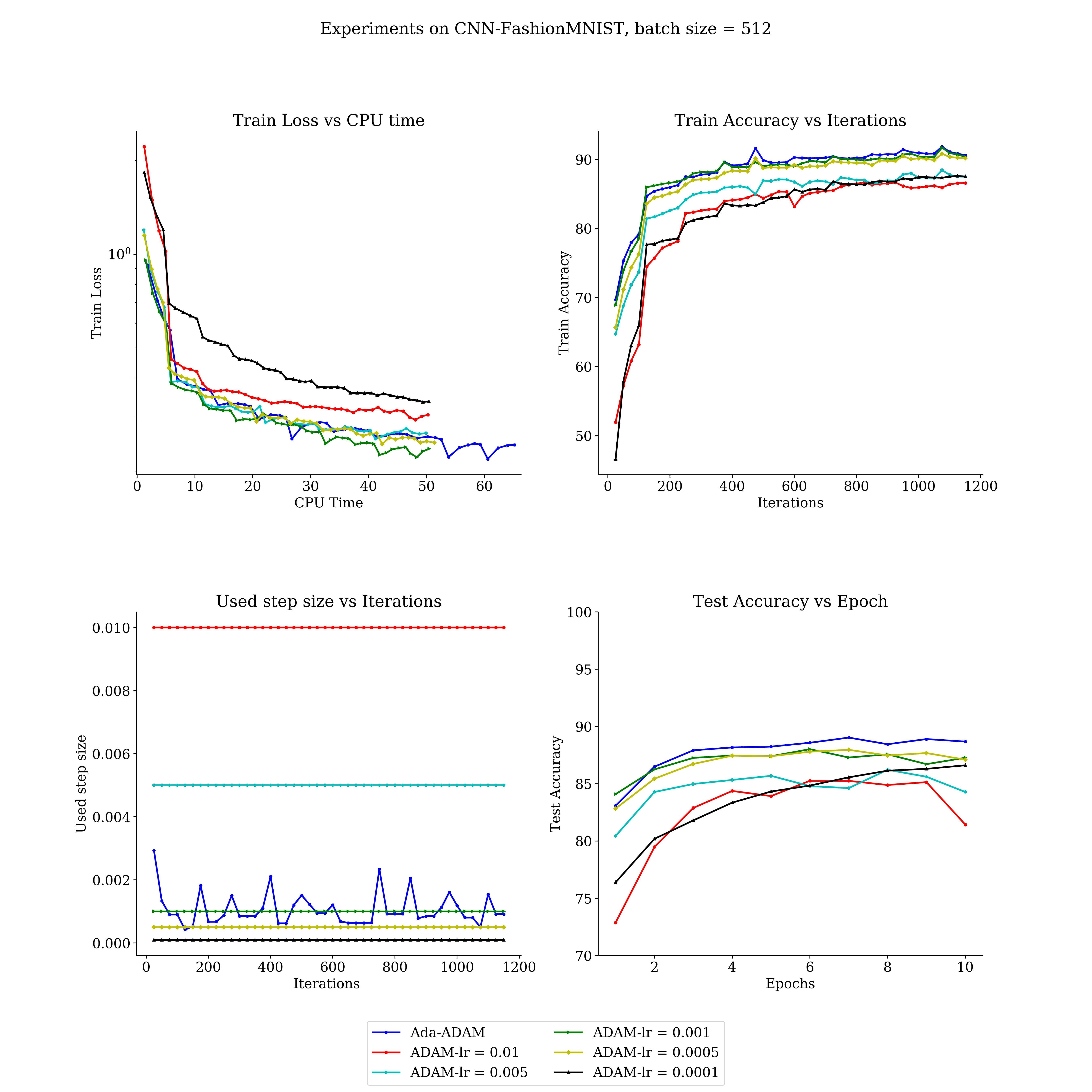}
\caption{Ada-ADAM Numerical results on Fashion-MNIST}
\label{adadamfmnist}
\end{figure}

{\bf CNN Fashion-MNIST:} We trained a CNN on Fashion-MNIST dataset using Ada-ADAM and ADAM with various learning rates. Fashion-MNIST is a dataset comprising of 28x28 grayscale images of 70,000 fashion products from 10 categories, with 7,000 images per category. The training set has 60,000 images and the test set has 10,000 images. Fashion-MNIST is intended to serve as a direct drop-in replacement for the original MNIST dataset for benchmarking machine learning algorithms, as it shares the same image size, data format and the structure of training and testing splits. The hyper-parameters for this experiment are:

\begin{center}
 \begin{tabular}{||c | c||} 
 \hline
 Hyperparameter & Values  \\ [0.5ex] 
 \hline\hline
 Architecture of CNN & \{Conv2d(1,16,5)+BN+ReLU+MaxPool(2,2), \\ & Conv2d(16,32,5)+BN+ReLU+MaxPool(2,2), \\
 & Linear(1568,10)\}  \\ 
 \hline
  ADAM $\beta_1$ & \{0.9\}  \\ 
 \hline
  ADAM $\beta_2$ & \{0.999\}  \\
 \hline
 batch size & \{512\} \\
 \hline
 Constant learning rates comparison & \{0.01, 0.005, 0.001, 0.0005, 0.0001\} \\
 \hline
 level of curvature precision $\epsilon$ & \{0.01\} \\
 \hline
\end{tabular}
\end{center}
Results are presented in Figure \ref{adadamfmnist}.

\subsubsection{Machines used for experiments:} 
- Personal computer, Processor: Intel(R) Core(TM) i7-7700HQ CPU @ 2.80GHz (8 CPUs), ~2.8GHz. Memory: 16GB RAM. GPU Card name: NVIDIA GeForce GTX 1050.

- Cloud instance, 8 vCPUs, 30Gb RAM, 1GPU Nvidia Tesla P100.

\end{document}